\documentclass[reqno]{amsart}
\usepackage{graphicx}
\usepackage{amssymb}
\usepackage{subcaption}
\usepackage{float}
\usepackage[]{algorithm2e}

\usepackage[pagewise]{lineno}
\usepackage[toc, page]{appendix}

\usepackage{algorithmic}
\usepackage{hyperref}
\hypersetup{
colorlinks=true,
linkcolor=red,
filecolor=magenta, 
urlcolor=cyan,
} 
\usepackage{color, soul}
\usepackage{xcolor}
\usepackage{bm}

\newtheorem{theorem}{Theorem}[section]

\theoremstyle{definition}
\newtheorem{definition}[theorem]{Definition}
\newtheorem{example}[theorem]{Example}

\theoremstyle{remark}

\DeclareMathOperator*{\argmin}{arg\,min}
\numberwithin{equation}{section}

\newcommand{\mbb}[1]{\mathbb{#1}}
\newcommand{\mcal}[1]{\mathcal{#1}}
\newcommand{\NN}{\mathcal{N}\mathcal{N}}

\title{Regularizing Instabilities in Image Reconstruction Arising from Learned Denoisers}
\author{Abinash Nayak}
\usepackage{fancyhdr}
\begin{document}




\date{\today}


\begin{abstract}
It's well-known that inverse problems are ill-posed and to solve them meaningfully one has to employ regularization methods. Traditionally, popular regularization methods have been the penalized Variational approaches, also known as Tikhonov-type regularization. In recent years, the classical regularized-reconstruction approaches have been outclassed by the (deep-learning based) learned reconstruction algorithms. However, unlike the traditional regularization methods, the theoretical underpinnings, such as stability and regularization convergence, have been insufficient for such learned reconstruction algorithms. Hence, the results obtained from such algorithms, though empirically outstanding, can't always be completely trusted, as they may contain certain instabilities or hallucinated features arising from the learned process. In fact, it has been shown that such learned algorithms are very susceptible to small (adversarial) noises in the data and can lead to severe instabilities in the recovered solution, which can be quite different from the inherent instabilities of the ill-posed inverse problem. Where as, the classical regularization methods can handle such noises very well and can produce stable recoveries. Here, we try to present certain regularization methods to stabilize such unstable learned reconstruction methods and recover a regularized solution, even when dealing with noisy data. For this, we need to extend the classical notion of regularization and incorporate it in the learned reconstruction algorithms. We also present some regularization techniques to regularize two of the most popular learned reconstruction algorithms, the Learned Post-Processing Reconstruction and the Learned Unrolling Reconstruction. We conclude with numerical examples validating the developed theories, i.e., showing instabilities in the learned reconstruction algorithms and providing regularization/stabilizing techniques to subdue them.
\end{abstract}
\maketitle

\section{Classical Regularization}
If the inverse problem, corresponding to the following forward equation
\begin{equation}\label{Forward equation}
Tx = y,
\end{equation}
is ill-posed, in the sense of violating any of the Hadamard's conditions for well-posedness: 1. Existence, 2. Uniqueness, and 3. Continuity, then one has to invoke certain regularization methods to recover a meaningful solution of \eqref{Forward equation}.

As stated in \cite{Engl+Hanke+Neubauer}, \textbf{``a (convergent) regularization method"} is defined as follows
\begin{definition}\label{Definition Regularization}
Let $T: \mcal{X} \rightarrow \mcal{Y}$ be a bounded linear operator between Hilbert spaces $\mcal{X}$ and $\mcal{Y}$, $\alpha_0 \in (0, +\infty]$. For every $\alpha \in (0,\alpha_0)$, let
\begin{equation}
R_\alpha : \mcal{Y} \rightarrow \mcal{X}
\end{equation}
be a continuous (not necessarily linear) operator. The family $\{R_\alpha\}$ is called a regularization or a regularization operator for $T^\dagger$ (the pseudo-inverse of $T$), if, for all $y \in \mcal{D}(T^\dagger)$, there exists a parameter choice rule $\alpha = \alpha(\delta, y_\delta)$ such that 
\begin{equation}\label{Regularization convergence}
\limsup_{\delta \rightarrow 0}\; \{ ||R_{\alpha(\delta,y_\delta)}y_\delta - T^\dagger y||_{\mcal{X}} \; : \; y_\delta \in \mcal{Y}, ||y_\delta - y||_{\mcal{Y}} \leq \delta \} = 0
\end{equation}
holds. In addition,
\begin{equation}
\alpha : \mbb{R}^+ \times \mcal{Y} \rightarrow (0,\alpha_0)
\end{equation}
is such that
\begin{equation}\label{Parameter choice convergence}
\limsup_{\delta \rightarrow 0} \; \{ \alpha(\delta,y_\delta) \; : \; y_\delta \in \mcal{Y}, ||y_\delta - y||_{\mcal{Y}} \leq \delta \} = 0.
\end{equation}
For a specific $y \in \mcal{D}(T^\dagger)$, a pair $(R_\alpha, \alpha)$ is called a (convergent) regularization method for solving $Tx = y$ if \eqref{Regularization convergence} and \eqref{Parameter choice convergence} hold.
\end{definition}

The most popular way to formulate such a regularization method is known as 
\subsection{Variational/Penalized/Tikhonov- Regularization:}
\mbox{ }

Here, one can define the regularization operator $R_\alpha$ as follows:
\begin{equation}\label{Tikhonov regularization}
R_\alpha(y_\delta) := \argmin_{x} \; D(Tx,y_\delta) \; + \; \alpha \mcal{R}(x),
\end{equation}
where $D(Tx,y_\delta)$ is a discrepancy term (such as $||Tx - y_\delta||_\mcal{Y}$), $\mcal{R}(x)$ is a regularization term (like $||Lx - y_0||_\mcal{Y}$, for some smoothing operator $L: \mcal{X} \rightarrow \mcal{Y}$, or something else), and $\alpha \geq 0$ (the regularization parameter) balances between the data consistency term and the regularization term. Hence, a regularized solution, in this case, is given by $R_{\alpha(\delta,y_\delta)}(y_\delta)$, where $\alpha(\delta,y_\delta)$ is determined via a parameter choice rule, which satisfies \eqref{Parameter choice convergence}. The formulation \eqref{Tikhonov regularization} also has a Bayesian interpretation, where the minimization of $R_\alpha(y_\delta)$ corresponds to the maximum-a-posteriori (MAP) estimate of $x$ given $y_\delta$, where the likelihood of $y_\delta$ is proportional to $exp(-D(Tx,y_\delta))$ and the prior distribution of $x$ is proportional to $exp(-\mcal{R}(x))$. {Instances of the above framework include total variational (TV) regularization} \cite{Rudin_Osher_Fatemi, Rudin_Osher_1994}, {sparsity in some basis or compressed-sensing based regularization} \cite{Donoho_Elad_Bruckstein_2009, Daubechies_Defrise_DeMol_2004, Candes_Romberg_Tao_2006, Tibshirani_2006}\nocite{Abinash1}, etc.

\section{Learned Reconstruction Methods and Regularization}
\subsection{Deep-Learning Based Reconstruction Methods:}\mbox{ }

With the advent of deep learning, the focus of solving an inverse problem has shifted from the traditional fashion towards the data-driven methods. That is, one can train a (deep) neural network $\mcal{N}\mcal{N}_\theta$, for a given set of training examples, so as to formulate a \textit{``reconstruction method"} that enables to recover a solution of \eqref{Forward equation}, for a given new data $y_\delta$, which is outside of the training example set. In other words, given $\theta \in \mbb{R}^d$, for some high dimension $d$, and neural-network structures $\mcal{N}\mcal{N}_\theta$ (depending on $\theta$), one has \textit{``a family of reconstruction methods"}:
\begin{equation}\label{R(NN,theta)}
R(\mcal{N}\mcal{N}, \theta) := \mcal{N}\mcal{N}_\theta : \mcal{Y} \rightarrow \mcal{X}.
\end{equation}
Now, fixing the structure of the neural network, one still gets \textit{``a family of reconstruction methods"}, for $\theta \in \mbb{R}^d$, for some $d >>1$,
\begin{equation}\label{R(theta)}
R_{\theta} := \mcal{N}\mcal{N}_\theta : \mcal{Y} \rightarrow \mcal{X}.
\end{equation}
However, after training the network, one only has \textit{``a fixed reconstruction method"}
\begin{equation}\label{R(theta0)}
R_{\theta_0} := \mcal{N}\mcal{N}_{\theta_0} : \mcal{Y} \rightarrow \mcal{X}.
\end{equation}
{In recent years, considerable amount of reconstruction networks have been introduced in the literature to solve inverse problems, see }\cite{Zhu_Liu_Cauley_Bruce_Matthew_2018, Gregor_LeCun_2010, Yang_Sun_Li_Xu_2016, Oktem_Adler_2017, Unser_Gupta_Jin_Nguyen_McCann_2018, Jin_McCann_Froustey_Unser_2017, Knoll_Pock_Hammernik_Klatzer_Kobler_Recht_Sodickson_2018, Bora_Dimakis_Jalal_Price_2017, Wang_Chen_Yi_Weihua_Liao_Li_Zhou_2017, Genzel_Macdonald_Marz_arXiv2020} { and survey papers }\cite{Carola_Oktem_Maass_arridge_2019,Willet_Dimakis_Ongie_Jalal_Metzler_Baraniuk_2020}. {Usually, these data-driven methods significantly outperform classical methods in terms of reconstruction accuracy and speed. However, they lack the theoretical underpinnings, such as robustness, stability, regularization convergence, etc., which are essentials when solving an inverse problem. Although, empirical results in many papers reflect the resilience of these methods against generic noise, see} \cite{Zhu_Liu_Cauley_Bruce_Matthew_2018, Oktem_Andreas_Jonas_Arridge_2020}, {numerous evidences also indicate the inherent instabilities arising in such deep-learning based reconstruction methods, see} \cite{Antun_Renna_Poon_Adcock_Hansen, Antun_Hansen_Adcock_Gottschling_arXiv2020, Huang_Maier_Tobias_Katharina_Ling_Gunther_2018}. {In this paper, we also present certain instabilities and hallucinations arising in reconstruction algorithms involving deep-learning based components, such as pre-trained networks or denoisers, and regularization techniques to stabilize them.}

\subsection{Regularizing Learned Reconstruction Methods:}\mbox{ }

{As aforementioned, while the learned reconstruction methods empirically yield much better results than the traditional regularization approaches, they are not known to be a convergent regularization method, as defined in Definition} \ref{Definition Regularization}. {In} \cite{Schwab_Antholzer_Haltmeier_2019}, {the authors proposed a deep-learning based reconstruction algorithm which is a convergent regularization method of certain type, which they define as a $\Phi$-regularization method. The authors proposed a special kind of neural network (\textit{null space network}) and introduced the concept of $\Phi$-regularization. The focus is to construct a neural network structure (say \textbf{L}) that supports data-consistency, i.e., $TLx = Tx$, for all $x \in \mcal{X}$. Hence, the goal is to split $Lx = x + z$ such that $z \in \mbox{ker}(T)$. They proposed a network of the following form, which is termed as the null space network,}
\begin{equation}\label{L nullspace network}
L = Id_X + (Id_X - T^* T)N \hspace{0.5cm} \mbox{ for Lipschitz continuous } N:\mcal{X} \rightarrow \mcal{X}.
\end{equation}
{Note that, $(Id_X - T^* T)$ is a projector onto the null space of T, i.e.,\\ $z = (Id_X - T^* T)N(x) \in \mbox{ker}(T)$, where $Id_X(x) = x$. In addition, if the network $L$ is trained on a set of training examples $\{(x_i, T^*Tx_i)\}_{i=1}^N$, then, for any classical regularization method $(B_\alpha)_{\alpha > 0}$, the two-stage reconstruction algorithm}
\begin{equation} \label{RalphaL}
R_\alpha^L := L \circ B_\alpha, \;\;\; \alpha > 0,
\end{equation}
{yields a $\Phi$-regularization method with $\mcal{M} := L(\mbox{ran}(T^*))$, where $\Phi$-regularization is defined as follows}
\begin{definition}\textbf{($\Phi$-regularization method)}\\
{Let $(R_\alpha)_{\alpha>0}$ be a family of continuous (not necessarily linear) mappings $R_\alpha : \mcal{Y} \rightarrow \mcal{X}$ and let $\alpha^*:(0,\infty)\times \mcal{Y} \rightarrow (0,\infty)$. The pair $((R_\alpha)_{\alpha>0},\alpha^*)$ is called a $\Phi$-regularization method for the equation $Tx = y$ if the following hold:}
\begin{itemize}
\item $\forall y \in \mcal{D}(T^*): \lim_{\delta \rightarrow 0} \sup\{ \alpha^*(\delta,y^\delta) \; | \; y^\delta \in \mcal{D}(T^*) \wedge ||y^\delta - y|| \leq \delta \} = 0$

\item $\forall y \in \mcal{D}(T^*): \lim_{\delta \rightarrow 0}\sup \{ ||T^\Phi y - R_{\alpha^*(\delta,y^\delta)}y^\delta|| \; | \; y^\delta \in \mcal{D}(T^*) \wedge ||y^\delta - y|| \leq \delta \} = 0$,
\end{itemize}
{where the function $T^\Phi$ is defined below.}
\end{definition}
{In the case that $((R_\alpha)_{\alpha > 0}, \alpha^*)$ is a $\Phi$-regularization method for $Tx = y$, then the family $(R_\alpha)_{\alpha > 0}$ is called a regularization of $T^\Phi$ and $\alpha^*$ an admissible parameter choice.}

{For a given linear and bounded $T: \mcal{X} \rightarrow \mcal{Y}$, let $\Phi : \mcal{X} \rightarrow \mbox{ker}(T) \subseteq \mcal{X}$ be Lipschitz continuous and $\mcal{M} := (Id_X + \Phi)\mbox{ran}(T^*)$}

\begin{definition}\textbf{($\Phi$-generalized inverse)}\\
{A map $T^\Phi : \mcal{D}(T^*) \subseteq \mcal{Y} \rightarrow \mcal{X}$ is called the $\Phi$-regularized inverse of $T$ if}
\begin{equation}
\forall y \in \mcal{D}(T^*) : T^\Phi y = (Id_X + \Phi)(T^* y).
\end{equation}
\end{definition}
{Note that, the $\Phi$-generalized inverse ($T^\Phi$) coincides with the Moore-Penrose generalized inverse ($T^\dagger$) if and only if $\Phi \equiv 0 $, in which case $\mcal{M} = \mbox{ran}(T^*) = \mbox{ker}(T)^\perp$.}

{In short, the essence of a $\Phi$-regularization method is as follows: }
\begin{itemize}
\item {There exist a tuple $((R_\alpha)_{\alpha > 0}, \alpha^*)$ such that}
	\begin{enumerate}
	\item {$R_\alpha : \mcal{Y} \rightarrow \mcal{X}$ are continuous mappings}
	\item {$\alpha^* = \alpha^*(\delta,y^\delta)$ is a suitable parameter choice rule, and}
	\end{enumerate}
\item {$\mcal{M} \subseteq \mcal{X}$ is a set of admissible elements, defined by the function $\Phi$, such that}
\item {For any $x \in \mcal{M}$, such that $Tx = y$, one has $R_{\alpha^*(\delta,y^\delta)}(y^\delta) \rightarrow x$, as $\delta \rightarrow 0$.}
\end{itemize}

{Therefore, for operators defined in} \eqref{RalphaL}, {one would obtain a $\Phi$-regularization method with $\mcal{M} := L(\mbox{ran}(T^*))$. Note that, as described in} \cite{Schwab_Antholzer_Haltmeier_2019}, {the key to $\Phi$-regularization is constructing and training a \textit{null space network} \textbf{L} (as defined in} \eqref{L nullspace network}) {appropriately, such that $TLx  = Tx$, which can be nontrivial.}

{In contrast, here we consider a different approach to regularize a deep-learning based reconstruction algorithm, where no strict constraints on the neural networks are needed. For simplicity, we stuck with the traditional regularization method $(R_\alpha,\alpha)$ and the classical Moore-Penrose generalized inverse solution ($T^\dagger y_\delta$), instead of the $\Phi$-genralized inverse and regularization. One of the most significant advantage of our regularization method over the null-space network regularization is that, one does not need to impose further restrictions on the neural networks to regularize it, rather, one regularizes the reconstruction algorithm through additional parameters. Hence, in addition to not worrying about any special training requirements, there is also no curbing for the expressivity of the networks.}

Note that, to connect the reconstruction algorithm \eqref{R(theta0)}, depending on $\theta \in \mbb{R}^d$ for $d >> 1$, with the classical regularization methods, depending $\alpha \in \mbb{R}$, one needs to first generalize the definition of a convergent regularization method. From Definition \ref{Definition Regularization}, one would like to approximate (in some sense) the discontinuous operator $T^\dagger$ via a family of continuous (not necessarily linear) operators $\{ R_\alpha \}_{\alpha \geq 0}$ such that, for all $y \in \mcal{D}(T^\dagger)$ and $y_\delta \in \mcal{Y}$, with $||y_\delta - y||_{\mcal{Y}} \leq \delta$, we have
\begin{gather}
\limsup_{\delta \rightarrow 0} \; ||R_{\alpha(\delta,y_\delta)}y_\delta - T^\dagger y||_{\mcal{X}}  = 0\\
\limsup_{\delta \rightarrow 0} \; \alpha(\delta,y_\delta) = 0. \label{alpha convergence}
\end{gather}
Now, by extending the condition \eqref{alpha convergence}, one can generalize the idea of the single-parameter regularization methods $\{R_\alpha\}_\alpha$, $\alpha \geq 0$, to multi-parameters reconstruction methods $\{ {R}_\theta \}_\theta$, for ${\theta \in \mbb{R}^d}$. Here, we can define a convergent regularization method as a collection of ``\textit{continuous operator (not necessarily linear)}" such that, for a parameter choice rule $\theta(\delta,y_\delta): \mbb{R}_+\times \mcal{Y} \mapsto \mbb{R}^d$, we have 
\begin{gather}
\limsup_{\delta \rightarrow 0} \; ||R_{\theta(\delta,y_\delta)}y_\delta - T^\dagger y||_{\mcal{X}}  = 0 \label{Rtheta convergence}\\
\limsup_{\delta \rightarrow 0} \; ||\theta(\delta,y_\delta)||_{\mbb{R}^d} = 0. \label{theta convergence}
\end{gather}
However, the above extended definition \eqref{Rtheta convergence}+\eqref{theta convergence} would not always work for a reconstruction method based on a neural network $\mcal{N}\mcal{N}_\theta$, especially for those $\NN_\theta$ which collapses when $\theta$ satisfies \eqref{theta convergence}. Hence, one can try to further generalize the concept of \textit{a convergent regularization method} as, when $\delta \rightarrow 0$, we need to have (in some sense of convergence)
\begin{equation}\label{Rthetadelta convergence}
{R}_{\theta(\delta)} \rightarrow T^\dagger,
\end{equation}
where ${R}_{\theta(\delta)}$ is a reconstruction method based on a training example set $\mcal{E}_\delta := \{ (x^i,y^i,y_\delta^{ij}) \}_{i,j=1}^{n,m}$, such that $Tx^i = y^i \in \mcal{D}(T^\dagger)$ and $y_\delta^{ij} \in \mcal{Y}$, such that $||y_\delta^{ij} - y^i||_{\mcal{Y}} \leq \delta$, and $R_{\theta(\delta)}y_\delta^{ij} = x^i$, for all $1 \leq i \leq n$ and $1 \leq j \leq m$. Note that, the noisy data $y_\delta^{ij}$ doesn't necessarily mean only adding noise to the noiseless data, i.e., $y_\delta^{ij} = y^i + \epsilon_\delta^j$. Here, $y_\delta^{ij}$ can also be  some form of distortion/perturbation of the exact data, for example, $y_\delta^{ij}$ can be undersampled data (in case of compressed sensing problems) together with certain stochastic noise in it.

The problem with this formulation is that, the reconstruction method $R_{\theta(\delta)}$ may not be continuous, i.e., if in the set of examples $\mcal{E}_\delta$, there exists two points $x^{i_1}$ and $x^{i_2}$, such that $||x^{i_1} - x^{i_2}||_\mcal{X} >> \delta$ but $||y^{i_1} - y^{i_2}||_{\mcal{Y}} \leq \delta$, then the reconstruction method $R_{\theta(\delta)}$ is discontinuous, as $||R_{\theta(\delta)}y^{i_1} - R_{\theta(\delta)}y^{i_2}||_{\mcal{Y}} >> \delta$. Consequently, although for the training examples, we have $R_{\theta(\delta)}y^{i_1j}_\delta = x^{i_1}$ and $R_{\theta(\delta)}y^{i_2j}_\delta = x^{i_2}$, for all $1\leq j \leq m$, but, for a new $y_\delta = y^{i_1} + \epsilon_\delta \in \mcal{Y}$ (outside of the training examples set) such that $||y^{i_1} - y_\delta||_{\mcal{Y}} \leq \delta$, it is possible to have $||R_{\theta(\delta)}y_\delta - x^{i_1}|| >> \delta$ and $||R_{\theta(\delta)}y_\delta - x^{i_2}|| \leq C(\delta)$, i.e., closer to $x^{i_2}$ and farther from $x^{i_1}$ (true solution). Now, the same is also true even when the training set $\mcal{E}_\delta$ does not have such pairs of $x^i$'s, but the operator behaves in that manner, i.e., if for a $x^{i_1} \in \mcal{E}_\delta$, there exist a $x^{i_2} \in \mcal{X}\backslash \mcal{E}_\delta$ such that $||x^{i_1} - x^{i_2}||_{\mcal{X}}>> \delta$ but $||Tx^{i_1} - Tx^{i_2}||_{\mcal{Y}} \leq \delta$, then for a $y_\delta = Tx^{i_2} + \epsilon_\delta$, such that $||Tx^{i_1} - y_\delta||_{\mcal{Y}} \leq \delta$, we can have $||R_{\theta(\delta)}y_\delta - x^{i_2}||_{\mcal{X}} >> \delta$ and $||R_{\theta(\delta)}y_\delta - x^{i_1}||_{\mcal{X}} \leq C(\delta)$. But, for an injective $T$ one can circumvent the above issue, as for any two $x^{i_1}, x^{i_2} \in \mcal{X}$, there exist a $\delta_0 > 0$ such that $||Tx^{i_1} - Tx^{i_2}||_{\mcal{Y}} > 2\delta_0$, and hence, $R_{\theta(\delta)}$ is continuous for all $\delta \leq \delta_0$, at least on the training example set $\mcal{E}_\delta$. Thus, in such scenarios, if \eqref{Rthetadelta convergence} holds and $R_{\theta(\delta)}$ generalize properly (i.e., $R_{\theta(\delta)}$ is also continuous on $\mcal{X}\backslash \mcal{E}_\delta$), then $(R_{\theta}, \theta(\delta))$ can be considered as a convergent regularization method. However, for a non-injective $T$ (i.e., non empty null space, $\mcal{N}(T) \neq \phi$), this is not possible (following the above procedure), as there does not exist any $\delta_0 > 0$ for which $\forall \; x^{i_1}, x^{i_2} \in \mcal{X}$, $||Tx^{i_1} - Tx^{x_2}||_{\mcal{Y}} > 2\delta_0$. The instabilities arising in such scenarios are illustrated in \cite{Antun_Renna_Poon_Adcock_Hansen}.

Moreover, one doesn't construct a family of trained neural networks $\{ R_{\theta(\delta)} := \NN_{\theta(\delta)} \}_{\delta \geq 0}$, based on a family of training examples sets $\{\mcal{E}_\delta\}_{\delta \geq 0}$, depending on different noise levels $\delta$, rather, one simply trains a neural network $R_{\theta_0} := \NN_{\theta_0}$ for various noise levels and hopes to obtain a regularized method for solving the inverse problem \eqref{Forward equation}, for all noise levels. Therefore, this defies the very essence of a (classical) convergent regularization method, which depends on a family of continuous operators designed to converge (in some sense) to the pesudo-inverse, when the noise level in the data goes to zero.

\section{Generating regularization methods from $R_{\theta_0}$}
Here, we suggest certain techniques to formulate a convergent regularization method, for solving the inverse problem \eqref{Forward equation}, when given a pre-trained/learned reconstruction method $R_{\theta_0} : \mcal{Y} \mapsto \mcal{X}$. First, note that, the keys to a convergent regularization method are the following three conditions:
\begin{enumerate}
\item \textbf{Continuity:} The regularization method $R_\alpha$, $\alpha \geq 0$, (or $R_\theta$, $\theta \in \mbb{R}^d$) needs to be continuous, as a function of $y$.
\item \textbf{Parameter Choice Rule:} There needs to be a parameter choice rule $\alpha(\delta,y_\delta)$ (or $\theta(\delta,y_\delta)$) such that for $y \in \mcal{D}(T^\dagger)$ and $y_\delta \in \mcal{Y}$, with $||y_\delta - y||_{\mcal{Y}} \leq \delta$, 
\begin{equation}
\limsup_{\delta \rightarrow 0} \; \alpha(\delta,y_\delta) = 0 \;\;\; \mbox{(or,} \;\; \limsup_{\delta \rightarrow 0} \; ||\theta(\delta,y_\delta)||_{\mbb{R}^d} = 0)
\end{equation}
\item \textbf{Convergence:} Again, for $y \in \mcal{D}(T^\dagger)$ and $y_\delta \in \mcal{Y}$, such that $||y_\delta - y||_{\mcal{Y}} \leq \delta$, we must have
\begin{gather}
\limsup_{\delta \rightarrow 0} \; ||R_{\alpha(\delta,y_\delta)}y_\delta - T^\dagger y||_{\mcal{X}}  = 0 \\
\left( \mbox{or,}\;\; \limsup_{\delta \rightarrow 0} \; ||R_{\theta(\delta,y_\delta)}y_\delta - T^\dagger y||_{\mcal{X}}  = 0 \right)
\end{gather}
\end{enumerate}

Hence, for a given learned $R_{\theta_0}$, one can attempt to define a family of regularization operator as follows, for $\alpha \geq 0$,
\begin{equation}\label{Rtheta0 regularized 1}
R_\alpha = T^\dagger + \alpha R_{\theta_0},
\end{equation}
where $T^\dagger$ is the pesudo-inverse of the forward operator $T$. However, the problem with this formulation is that, the operator $R_\alpha$ (as defined in \eqref{Rtheta0 regularized 1}) might not be continuous, since $T^\dagger$ is usually discontinuous for ill-posed inverse problems, and thus, $R_\alpha$ is also discontinuous even when $R_{\theta_0}$ is continuous. 

Therefore, one can circumvent the discontinuity issue by making use of a convergent family of regularized operators $\{R_\alpha\}$, such as, for all $y \in \mcal{Y}$, define 
\begin{equation}\label{Rtheta0 regularized 2}
R^{\theta_0}_{\alpha}(y) = R_\alpha(y) + \beta(\alpha,y) R_{\theta_0}(y),
\end{equation}
where $\beta : [0,\infty) \times \mcal{Y} \mapsto [0,\infty]$, with $\beta(\alpha,y) = \infty$ implying $R_\alpha^{\theta_0}(y) = R_{\theta_0}(y)$, is such that, for all $\alpha \geq 0$,
\begin{equation}\label{betaRtheta0 continuous}
\bullet \hspace{0.5cm} \mbox{the product $\beta(\alpha,.)R_{\theta_0}(.)$ is continuous,}
\end{equation}
and $\beta(\alpha,y)$ also satisfies, for all $y \in \mcal{Y}$,
\begin{equation}\label{betaRtheta0 convergence to 0}
\limsup_{\alpha \rightarrow 0} \; \beta(\alpha,y)||R_{\theta_0}(y)||_{\mcal{X}} = 0.
\end{equation}
Hence, the new family of operator $\{R_\alpha^{\theta_0} \}$ is also a convergent regularization method {(Proof in Appendix} \ref{Proof 1}).

Typically, one can alter $\beta(\alpha,y)$ in \eqref{Rtheta0 regularized 2} to a weighted average, such as,
\begin{equation}\label{Classical-Learned regularization}
R^{\theta_0}_{\alpha}(y) = (1 - \beta(\alpha,y))R_\alpha(y) + \beta(\alpha,y)R_{\theta_0}(y),
\end{equation}
where (now) $\beta : [0,\infty) \times \mcal{Y} \mapsto [0, 1]$. That is, for  a given $y_\delta \in \mcal{Y}$, such that $||y_\delta - y||_{\mcal{Y}} \leq \delta$, a regularized solution of the new regularization method, corresponding to a parameter choice $\alpha(\delta,y_\delta)$, is given by
\begin{equation}\label{Classical-Learned regularized solution}
R^{\theta_0}_{\alpha(\delta,y_\delta)}(y_\delta) = (1 - \beta(\alpha(\delta,y_\delta),y_\delta))R_{\alpha(\delta,y_\delta)}(y_\delta) + \beta(\alpha(\delta,y_\delta),y_\delta)R_{\theta_0}(y_\delta),
\end{equation}
Hence, even when the learned reconstruction $R_{\theta_0}(y_\delta)$ is unstable, by choosing an appropriate $\beta(\alpha(\delta,y_\delta),y_\delta)$, one can still recover a regularized (stable) solution $R^{\theta_0}_{\alpha(\delta,y_\delta)}(y_\delta)$, for the inverse problem \eqref{Forward equation}.

Now, one can further generalize the regularization operator defined in \eqref{Rtheta0 regularized 2} or \eqref{Classical-Learned regularization}, when having a family of classical regularization methods $\{ R_{\alpha_i} \}_{i=1}^N$ and a collection of learned reconstruction algorithms $\{ R_{\theta_i} \}_{i=1}^N$, in the following manner
\begin{equation}\label{Classical-Learned regularization extension}
R(y;\{\alpha_i\}, \{ \theta_i \}) = \sum_{i=1}^N \beta_i^1(\alpha_i,y) R_{\alpha_i}(y) + \sum_{i=1}^N \beta_i^2(\alpha_i,y) R_{\theta_i}(y),
\end{equation}
such that, for all $y \in \mcal{Y}$,
\begin{equation}
\sum_{j=1}^2 \sum_{i=1}^N \beta_i^j(\alpha_i,y) = 1
\end{equation}
and $\beta_i^j(\alpha_i,y)$ satisfies conditions \eqref{betaRtheta0 continuous} and \eqref{betaRtheta0 convergence to 0}, for all $1 \leq i \leq N$ and $j=1,2$. This extended definition will be helpful when regularizing a learned unrolled reconstruction algorithm, which is defined in the next section.

\subsection{Error Estimates:}\mbox{ }

{Note that, performing error estimates for any arbitrary deep-learning reconstruction method $R_{\theta_0}(y_\delta)$ is nontrivial, as we are not imposing any restrictions on the neural network. Moreover, $R_{\theta_0}$ also implicitly depends on the structure of the neural networks, the training set of examples and the training strategies. However, upon imposing certain restrictions, one can connect the overall reconstruction algorithm to (semi-) iterative regularization methods, and hence, error estimates can be performed to an extent. For example, given a noisy $y_\delta$, with  $\delta > 0$, and a differentiable discrepancy term, $\nabla_x D(Tx, y_\delta)$ exists, if $R_{\theta_0}(y_\delta)$ satisfies the following criterion}
\begin{equation}\label{*}
\Big( R_{\theta_0}(y_\delta), -\nabla_x D(TR_{\alpha(\delta,y_\delta)}(y_\delta), y_\delta) \Big)_{\mcal{X}} > 0 \;, 
\end{equation}
{where $\Big( ., . \Big)_{\mcal{X}}$ denotes the inner-product in the Hilbert space $\mcal{X}$ and $\nabla_x D(TR_{\alpha(\delta,y_\delta)}(y_\delta),y_\delta)$ is the gradient of $D$ at $x = R_{\alpha(\delta,y_\delta)}(y_\delta)$, then the learned solution $R_{\alpha(\delta,y_\delta)}^{\theta_0}(y_\delta)$ is a better fit for the given data ($y_\delta$) than the classical solution $R_{\alpha(\delta,y_\delta)}(y_\delta)$, i.e.,}
\begin{equation}\label{**}
||TR_{\alpha(\delta,y_\delta)}^{\theta_0}(y_\delta) - y_\delta||_{\mcal{Y}} \leq ||TR_{\alpha(\delta,y_\delta)}(y_\delta) - y_\delta||_{\mcal{Y}}. 
\end{equation}
{However,} \eqref{**} {doesn't imply that}
\begin{equation}
||R_{\alpha(\delta,y_\delta)}^{\theta_0}(y_\delta) - T^\dagger y||_{\mcal{X}} \leq ||R_{\alpha(\delta,y_\delta)}(y_\delta) - T^\dagger y||_{\mcal{Y}},
\end{equation}
{since $R_{\alpha(\delta,y_\delta)}^{\theta_0}(y_\delta)$ may either overfit to the noisy data ($y_\delta$), leading to a noisy recovery, or $R_{\alpha(\delta,y_\delta)}^{\theta_0}(y_\delta)$ may contain certain false/hallucinated features, arising from the training set distribution. Equation} \eqref{**} {also doesn't imply that}
\begin{equation}\label{Ralphatheta0 > Ralpha}
||R_{\alpha(\delta,y_\delta)}^{\theta_0}(y_\delta) - T^\dagger y||_{\mcal{X}} \geq ||R_{\alpha(\delta,y_\delta)}(y_\delta) - T^\dagger y||_{\mcal{Y}},
\end{equation}
{since, for a given parameter choice rule $\alpha(\delta,y_\delta)$, the classical solution $R_{\alpha(\delta,y_\delta)}(y_\delta)$ might have underfit the data $y_\delta$ and condition} \eqref{*} {leads to a solution $R_{\alpha(\delta,y_\delta)}^{\theta_0}(y_\delta)$ which fits $y_\delta$ \textit{appropriately}, and hence, a better approximation. In other words, a convergent regularization method $(R_\alpha,\alpha)$, for solving the inverse problem $Tx = y$, only describes the necessary asymptotic behavior of the pair $(R_\alpha,\alpha)$, i.e., satisfying conditions} \eqref{Regularization convergence} {and} \eqref{Parameter choice convergence}. {For a non asymptotic case ($\delta > 0$), the accuracy of the recovered solution $R_\alpha(\delta,y_\delta)y_\delta$ is dependent not only on the regularization operator $R_\alpha$, but also on the parameter choice rule $\alpha(\delta,y_\delta)$. Hence, for $\delta > 0$, the ``classical+learned regularized" solution $R_{\alpha(\delta,y_\delta)}^{\theta_0}(y_\delta)$ will be a better ``approximation" to the true solution than the ``classical regularized" solution $R_{\alpha(\delta,y_\delta)}(y_\delta)$ depending on the parameter choice rule $\alpha(\delta,y_\delta)$ and $R_{\theta_0}$, which instead depends on the training set of examples, training procedures, etc. }

{Note that, the importance of $\beta(\alpha,y)$ is prominent when $R_{\alpha(\delta,y_\delta)}^{\theta_0}(y_\delta)$ is a poorer recovery than $R_{\alpha(\delta,y_\delta)}(y_\delta)$, i.e.,} \eqref{Ralphatheta0 > Ralpha} {holds. In such scenarios one can construct another sequence of solutions $\{ R_{\alpha(\delta,y_\delta)}^{\theta_0}(y_\delta; \gamma) := R_{\alpha(\delta,y_\delta)}y_\delta + \gamma R_{\theta_0}(y_\delta) \}_\gamma$, where \\$0 \leq \gamma \leq \beta(\alpha(\delta,y_\delta),y_\delta)$, and an appropriate regularized solution $R_{\alpha(\delta,y_\delta)}^{\theta}(y_\delta; \gamma(\delta,y_\delta))$ can be obtained based on a parameter choice rule $\gamma(\delta,y_\delta)$. Another instance, where $\beta(\alpha,y)$ can be utilized to produce a better data-consistent result is when $R_{\theta_0}(y_\delta)$ fails to satisfy} \eqref{*}. {In this case, one can improve the data-consistency of the recovered solution $R_{\alpha(\delta,y_\delta)}^{\theta_0}(y_\delta)$ by modifying} \eqref{Rtheta0 regularized 2} {to} \eqref{A1} {as follows}
\begin{equation}\label{A1}
R_\alpha^{\theta_0}(y;\tau) = R_\alpha(y) + \beta(\alpha,y) R_{\theta_0} (y) - \tau\nabla_x D(TR_\alpha(y),y). 
\end{equation}
{The advantage of the above modification is that, when $R_{\theta_0}(y_\delta)$ doesn't satisfy} \eqref{*}, {then by having a}
\begin{equation}
\beta(\delta,y_\delta,\tau) \leq \min\{ \beta(\alpha(\delta,y_\delta),y_\delta) \; , \; \beta_0(\alpha(\delta,y_\delta), \tau) \},
\end{equation}
{where}
\begin{equation}
\beta_0(\alpha(\delta,y_\delta), \tau) < \frac{|| \tau \nabla_x D(TR_{\alpha(\delta,y_\delta)}(y_\delta), y_\delta) ||_{\mcal{X}}}{||R_{\theta_0}(y_\delta)||_{\mcal{X}}},
\end{equation}
{the recovered solution}
\begin{equation}
R_{\alpha(\delta,y_\delta)}^{\theta_0}(y_\delta;\tau) = R_{\alpha(\delta,y_\delta)}(y_\delta) + \beta(\delta,y_\delta,\tau) R_{\theta_0} (y_\delta) - \tau\nabla_x D(TR_{\alpha(\delta,y_\delta)}(y_\delta),y_\delta). \notag
\end{equation}
{is not only regularized, but also more data consistent, i.e., $R_{\alpha(\delta,y_\delta)}^{\theta_0}(y_\delta;\tau)$ satisfy} \eqref{**}. {Now, one can further extend} \eqref{A1} {by replacing the single gradient term $- \tau\nabla_x D(TR_\alpha(y),y)$ with a more general data-consistent term and repeating the process, which leads to the Unrolled scheme, discussed in \S} \ref{Regularized Unrolled Scheme Sec.}.

\section{Practical Applications}
Here, we provide two instances where one can use the above formulation to regularize two popular deep-learning based reconstruction algorithms:

\subsection{Regularizing Learned Post-Processing Reconstruction Methods:}
\mbox{ }

Here, for a given $y_\delta$, one recovers a solution of \eqref{Forward equation} via the following two steps
\begin{itemize}
\item Step 1: A classical recovery method, as defined in \eqref{Tikhonov regularization}, i.e.,  $R_{\alpha(\delta,y_\delta)}(y_\delta) := \argmin_{x} \; D(Tx,y_\delta) \; + \; \alpha(\delta,y_\delta) \mcal{R}(x)$, typically, $\mcal{R}(x) = ||x||_{\mcal{X}}$.

\item Step 2: A learned recovery method ($R_{\theta_0}$) applied to $R_{\alpha(\delta,y_\delta)}(y_\delta)$, usually, $R_{\theta_0} = \NN_{\theta_0}$, where $\NN_{\theta_0}$ is a pre-trained neural network.
\end{itemize}
That is, the final recovered solution is given by a composition of the above two methods:
\begin{equation}\label{Post-Processing method}
R^{\theta_0}_{\alpha(\delta,y_\delta)}(y_\delta) := R_{\theta_0}(R_{\alpha(\delta,y_\delta)}(y_\delta)).
\end{equation}
Note that, the recovery method (as defined in \eqref{Post-Processing method}) may not be continuous for a discontinuous $R_{\theta_0}$, and hence, the recovered solution might not be a ``regularized" solution for the inverse problem \eqref{Forward equation}. Now, as defined in \eqref{Classical-Learned regularized solution}, one can recover a regularized solution, for an appropriate $\beta(\alpha(\delta,y_\delta),y_\delta)$, as follows
\begin{equation}\label{Post-Processing Regularization 1}
R^{\theta_0}_{\alpha(\delta,y_\delta)}(y_\delta) = (1 - \beta(\alpha(\delta,y_\delta),y_\delta))R_{\alpha(\delta,y_\delta)}(y_\delta) + \beta(\alpha(\delta,y_\delta),y_\delta)R^{\theta_0}_{\alpha(\delta,y_\delta)}(y_\delta).
\end{equation}
In other words, starting from the classical regularized solution $R_{\alpha(\delta,y_\delta)}(y_\delta)$, one moves along the direction of the learned reconstruction,
\begin{equation}\label{Post-Processing direction}
d_{\alpha,\theta_0}^\delta(y_\delta) := R^{\theta_0}_{\alpha(\delta,y_\delta)}(y_\delta) - R_{\alpha(\delta,y_\delta)}(y_\delta),
\end{equation}
upto a certain extent, i.e.,
\begin{equation}\label{Post-Processing Regularization 2}
R^{\theta_0}_{\alpha(\delta,y_\delta)}(y_\delta) = R_{\alpha(\delta,y_\delta)}(y_\delta) + \beta(\alpha(\delta,y_\delta),y_\delta) \; d_{\alpha,\theta_0}^\delta(y_\delta).
\end{equation}
Hence, for a continuous $R_\alpha(.)$, the expression in \eqref{Post-Processing Regularization 2} is continuous when $\beta(\alpha,.)d_{\alpha,\theta_0}(.)$ is continuous.

\subsection{Regularizing Learned Unrolled Reconstruction Methods:}\label{Regularized Unrolled Scheme Sec.}
\mbox{ }

In this case, for a given $y_\delta$, instead of a single classical recovery $( R_{\alpha(\delta,y_\delta)}(y_\delta))$ which is then followed by a single learned recovery $(R_{\alpha(\delta,y_\delta)}^{\theta_0}(y_\delta))$, there is a series or sequence of classical recovery methods $\{ R_{\alpha_i(\delta,y_\delta)} \}_{i=0}^{N}$ and learned recovery methods $\{ R_{\theta_i} \}_{i=0}^{N}$ which are stacked alternatively (in certain fashion) to produce a final solution of the inverse problem \eqref{Forward equation}, i.e., for $1 \leq i \leq N$, we have either \textit{the composition structure}
\begin{equation}\label{Unrolling 1}
f_i(R_{\alpha_i}, R_{\theta_i}; y_\delta) = R_{\theta_i}(R_{\alpha_i}(f_{i-1}(R_{\alpha_{i-1}},R_{\theta_{i-1}};y_\delta)))
\end{equation}
or \textit{the addition structure}
\begin{equation}\label{Unrolling 2}
f_i(R_{\alpha_i}, R_{\theta_i}; y_\delta) = R_{\alpha_i}(f_{i-1}(R_{\alpha_{i-1}},R_{\theta_{i-1}};y_\delta)) + R_{\theta_i}(f_{i-1}(R_{\alpha_{i-1}},R_{\theta_{i-1}};y_\delta)),
\end{equation}
and the final recovered solution is given by
\begin{equation}\label{Unrolled Solution}
R({y_\delta;\{\alpha_i\},\{ \theta_i \}}) = f_N(R_{\alpha_N},R_{\theta_N};y_\delta).
\end{equation}
Here, $R_{\alpha_i}(.)$ are also referred as the \textit{data-consistency steps}, that generate solutions (in a regularized manner) to fit the noisy data $y_\delta$, and $R_{\theta_i}(.)$ as the \textit{data-denoising steps}, which denoise (further smoothing) the generated $R_{\alpha_i}$-solutions. The inspiration behind such algorithms are derived from optimization architectures such as Alternating Direction of Methods of Multipliers (ADMM) or Proximal-Gradient Methods (PGM). Again, the  above recovery method may not end up being a continuous process, and hence, the final recovered solution might not be ``regularized". Now, similar to the previous regularizing technique, one can regularize such unrolled methods, by defining a regularized solution as
\begin{equation}\label{Regularized Unrolled Solution 1}
R({y_\delta;\{\alpha_i\},\{ \theta_i \}}, \beta(\{ \alpha_i \},y_\delta)) = R_{\alpha_0}(y_\delta) + \beta(\{ \alpha_i \},y_\delta) \; d^\delta(y_\delta; \{\alpha_i\}, \{\theta_i\}),
\end{equation}
where the direction $d^\delta(y_\delta; \{\alpha_i\}, \{\theta_i\})$ can be defined as follows
\begin{equation}\label{Unrolling direction}
d^\delta(y_\delta; \{\alpha_i\}, \{\theta_i\}) := f_N(R_{\alpha_N},R_{\theta_N};y_\delta) - R_{\alpha_0}(y_\delta).
\end{equation}
However, in this case, instead of the starting point $R_{\alpha_0}(y_\delta)$, one can choose any intermediate iterate (of course, assuming the recovery upto that point is regularized) as the initial point for the direction $d^\delta(.)$, as defined in \eqref{Unrolling direction}, and can (even) recover a sequence of regularized solutions. In other words, similar to \eqref{Classical-Learned regularization extension}, one can generalize the definition \eqref{Regularized Unrolled Solution 1}, of a regularized solution, to a weighted regularized-solution, defined as follows
\begin{align}\label{Regularized Unrolled Solution 2}
R({y_\delta;\{\alpha_i\},\{ \theta_i \}}, \{\beta_i^j( \alpha_i ,y_\delta)\}) &= \sum_{i=1}^N \beta_i^1(\alpha_i,y_\delta)R_{\alpha_i}(y_\delta)+ \sum_{i=1}^N \beta_i^2(\alpha_i,y_\delta)R_{\theta_i}(y_\delta) \notag\\
& + \sum_{i=1}^N \beta_i^3( \alpha_i ,y_\delta) \; f_i(R_{\alpha_i}, R_{\theta_i}; y_\delta),
\end{align}
such that
\begin{equation}
\sum_{j=1}^3 \sum_{i=1}^N \beta_i^j( \alpha_i,y_\delta) = 1.
\end{equation}
Note that, with the definition \eqref{Regularized Unrolled Solution 2}, the regularization parameter $(\beta(\{ \alpha_i \},y_\delta) \geq 0)$ is not a single dimensional anymore, instead $\Theta = (\beta_i^j(\alpha_i, y_\delta)) \in \mbb{R}^{3N}$, and can be cumbersome to estimate an appropriate value for it. 

Another approach to regularize such unrolled scheme is to introduce a regularization parameter at every unrolled steps, i.e.,
\begin{align}\label{Regularized Unrolling 1}
f_i(\beta_i,R_{\alpha_i}, R_{\theta_i}; y_\delta) &= (1- \beta_i) \; R_{\alpha_i}(f_{i-1}(R_{\alpha_{i-1}},R_{\theta_{i-1}};y_\delta)) \\
& + \beta_i \; R_{\theta_i}(R_{\alpha_i}(f_{i-1}(R_{\alpha_{i-1}},R_{\theta_{i-1}};y_\delta))) \notag
\end{align}
or
\begin{align}\label{Regularized Unrolling 2}
f_i(\beta_i, R_{\alpha_i}, R_{\theta_i}; y_\delta) &= (1 - \beta_i) \; R_{\alpha_i}(f_{i-1}(R_{\alpha_{i-1}},R_{\theta_{i-1}};y_\delta)) \\
& + \beta_i \; R_{\theta_i}(f_{i-1}(R_{\alpha_{i-1}},R_{\theta_{i-1}};y_\delta)), \notag
\end{align}
to encounter instabilities arising (if any) at each and every steps.

In the next section we present examples illustrating the instabilities that can arise in such learned reconstruction algorithms, if not regularized properly. We also show that, by using the techniques presented above, one can recover a regularized/stable solution. This is explained in details in \cite{nayak2021PnPInstabilities}, where it's showed that the Plug-and-Play algorithms (which is a special case of the unrolled scheme) can also suffer instabilities, arising from a learned denoiser ($R_{\theta_0}$) in its denoising step and, by using regularization techniques, such as \eqref{Regularized Unrolling 1} or \eqref{Regularized Unrolling 2}, one can stabilize the recovery process, producing excellent stable solutions.

\section{Numerical Results}
In this section, we consider $\mcal{X} = \mbb{R}^n$, $\mcal{Y}=\mbb{R}^m$ and $T \in \mbb{R}^{m\times n}$, and solve the following linear (matrix) equation
\begin{equation}\label{Forward Equation Matrix}
T \hat{x} = y_\delta,
\end{equation}
for a given noisy $y_\delta \in \mbb{R}^m$, such that $||y_\delta - y||_2 \leq \delta$, and known $T \in \mbb{R}^{m \times n}$. In the following example, the forward equation \eqref{Forward Equation Matrix} corresponds to a discretization of the radon transformation, which is associated with a parallel-beam X-ray computed tomography (CT) image reconstruction problem. Here, we generate the matrix $T$ from the MATLAB codes presented in \cite{Hansen_IRtools}. The dimension $n = n_1 n_2$ corresponds to the size of a $n_1 \times n_2$ phantom image, and the dimension $m = m_1m_2$ corresponds to a $m_1 \times m_2$ sinogram image, where $m_1$ implies the number of rays per angle and $m_2$ implies the number of angles. We consider a $128\times 128$ chest CT-image ($\hat{x}$) (obtained from MATLAB's \textit{``chestVolume"} dataset) and generate a noiseless sparse view data ($y = A\hat{x}$), with only 90 views and 181 rays/angle, i.e., a $181\times 90$ sinogram image. To avoid inverse-crime we add  Poisson noise of intensity $I_0=10^6$, i.e., first normalize the intensities of $y$ to $y^{01} := \frac{y - \min(y)}{\max(y)-\min(y)} \in [0,1]$ and add Poisson noise
\begin{equation}
y^{01}_\delta = -\log \left( \frac{\mbox{poissrnd}(I_0\exp(-y^{01}))}{I_0} \right),
\end{equation}
where $\log()$, $\mbox{poissrnd}()$ and $\exp()$ are MATLAB's inbuilt routines. Finally, $y^{01}_\delta$ is scaled back up via $y_\delta := y_\delta^{01}(\max(y) - \min(y)) + \min(y)$.

{Note that, the experiments presented here is to illustrate the instabilities arising in a reconstruction algorithm, which is based on learned components, and regularization techniques to subdue these instabilities. It is empirically shown in many recent papers, such as} \cite{Zhu_Liu_Cauley_Bruce_Matthew_2018, Gregor_LeCun_2010, Oktem_Adler_2017, Jin_McCann_Froustey_Unser_2017, Knoll_Pock_Hammernik_Klatzer_Kobler_Recht_Sodickson_2018, Bora_Dimakis_Jalal_Price_2017, Wang_Chen_Yi_Weihua_Liao_Li_Zhou_2017}, {that such deep-learning based reconstruction algorithms can significantly outperform any traditional regularization methods, such as total variation (TV) regularization or sparsity based regularization, in terms of the accuracy and speed of the reconstruction. However, the performance of these learned methods suffer critically under any distribution shifts or adversarial noises. Hence, the goal of this paper is not to propose another new $\NN_\theta$ structure and claiming it to outperform some existing structures. Instead, we use a pre-trained $\NN_{\theta_0}$ structure (which can also be a deep-learning denoiser) in our learned modules and show that, if used naively, it can produce strange instabilities, such as hallucinated features/artifacts or complete breakdown of the recovery process, which are quite different from the inherent instabilities of an ill-posed inverse problem. Furthermore, to show that the $\NN_{\theta_0}$ is not completely unfeasible, we use the same $\NN_{\theta_0}$ in a regularized fashion to recover excellent stable/regularized solutions. Therefore, validating that the instabilities not only arise from the adversarial noises, but also the manner in which the reconstruction algorithms are implemented, i.e., the instabilities can be inherent to the reconstruction structure and with proper regularization techniques, one can subdue these instabilities and recover stable solutions.}

\begin{figure}[h!]
    \centering
    \begin{subfigure}{0.495\textwidth}
        \includegraphics[width=\textwidth]{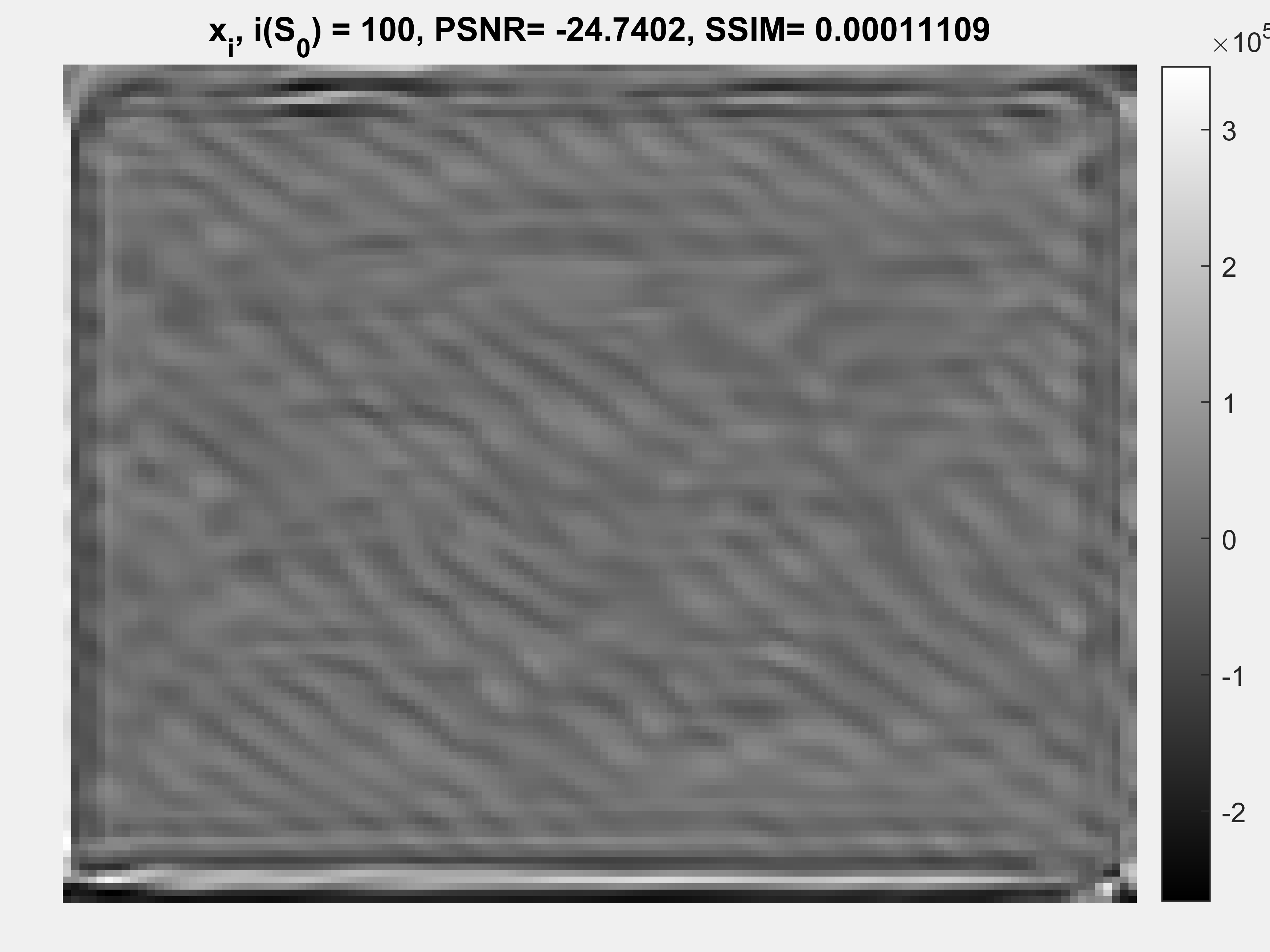}
        \caption{$x_N^{N_0}(\beta_i)$; $N_0=1, \beta_i = 1$}
    \end{subfigure}       
    \begin{subfigure}{0.495\textwidth}
        \includegraphics[width=\textwidth]{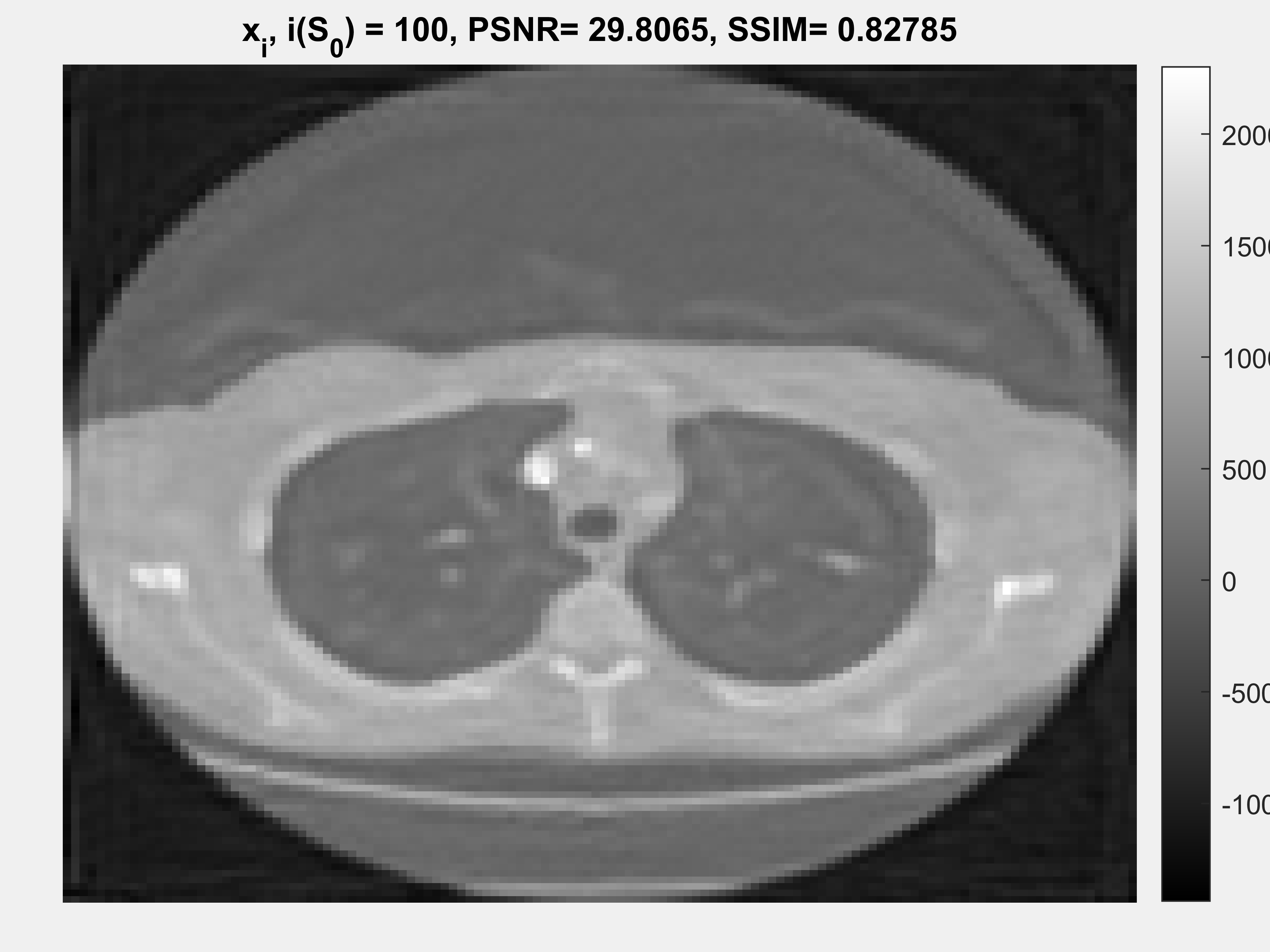}
        \caption{$x_N^{N_0}(\beta_i)$; $N_0=1, \beta_i = \beta_i(\mcal{S}_0)$}
    \end{subfigure}       
    \begin{subfigure}{0.495\textwidth}
        \includegraphics[width=\textwidth]{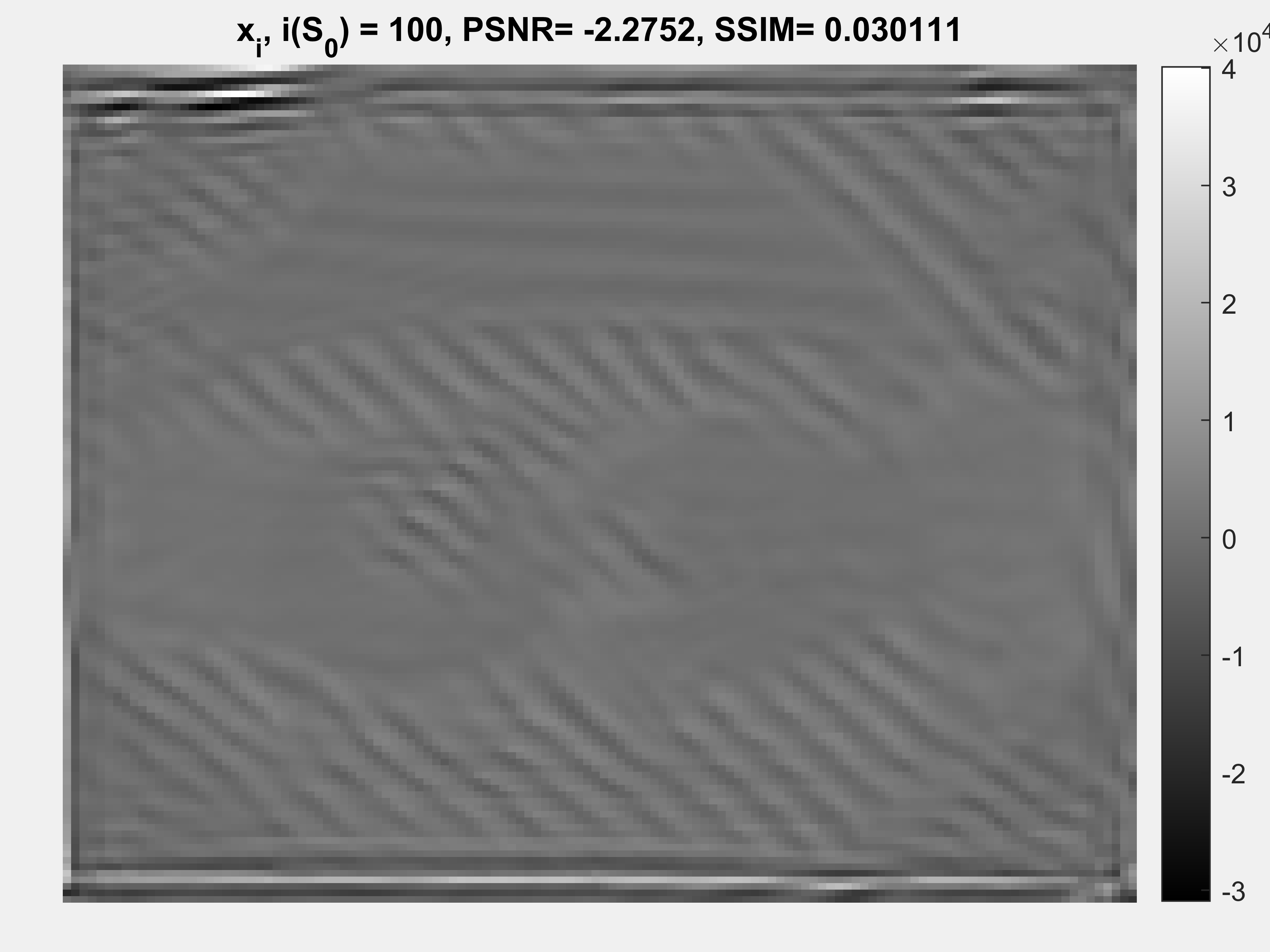}
        \caption{$x_N^{N_0}(\beta_i)$; $N_0=10, \beta_i = 1$}
    \end{subfigure} 
    \begin{subfigure}{0.495\textwidth}
        \includegraphics[width=\textwidth]{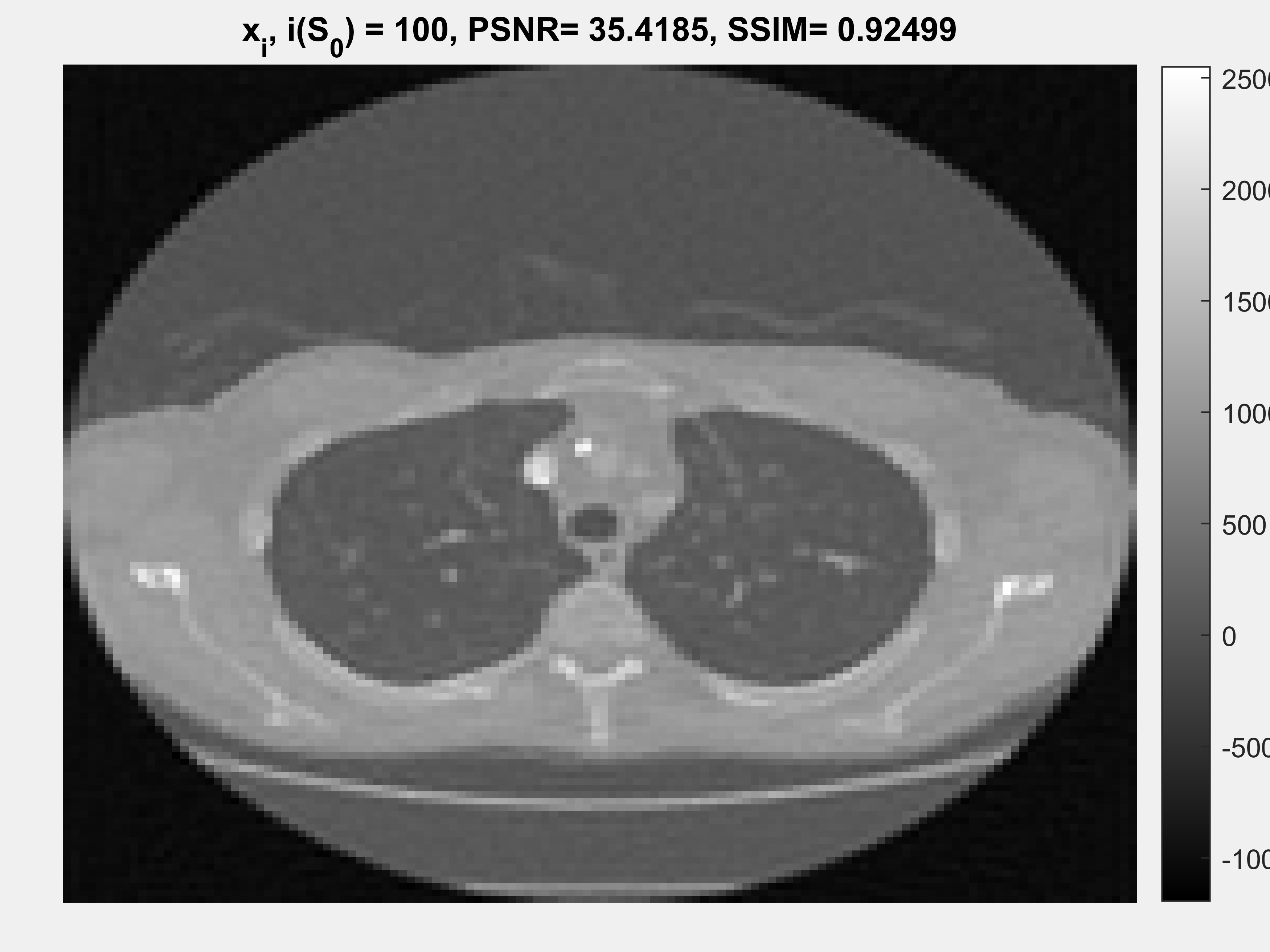}
        \caption{$x_N^{N_0}(\beta_i)$; $N_0=10, \beta_i = \beta_i(\mcal{S}_0)$}
    \end{subfigure}    
    \begin{subfigure}{0.495\textwidth}
        \includegraphics[width=\textwidth]{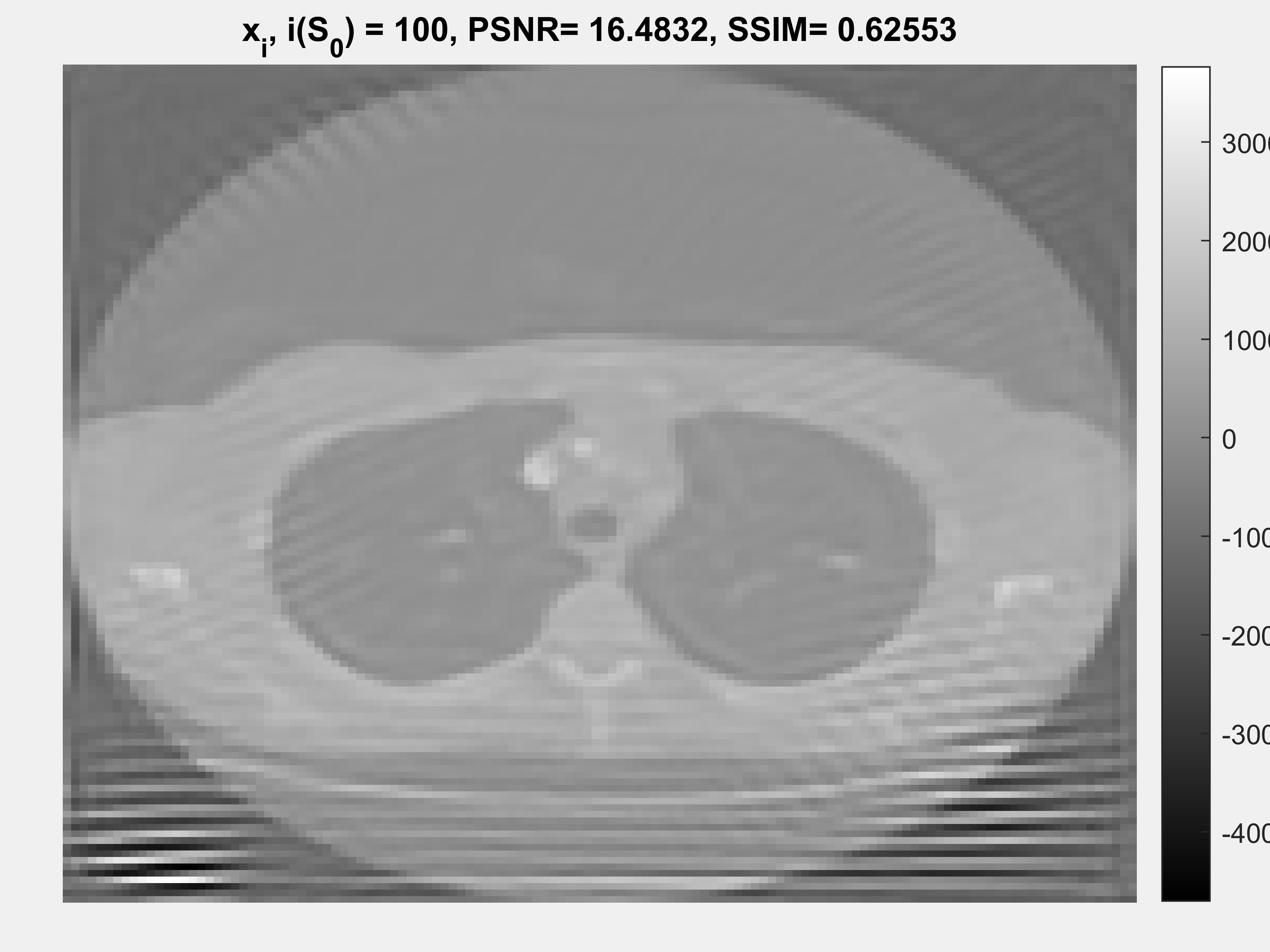}
        \caption{$x_N^{N_0}(\beta_i)$; $N_0=50, \beta_i = 1$}
    \end{subfigure}
    \begin{subfigure}{0.495\textwidth}
        \includegraphics[width=\textwidth]{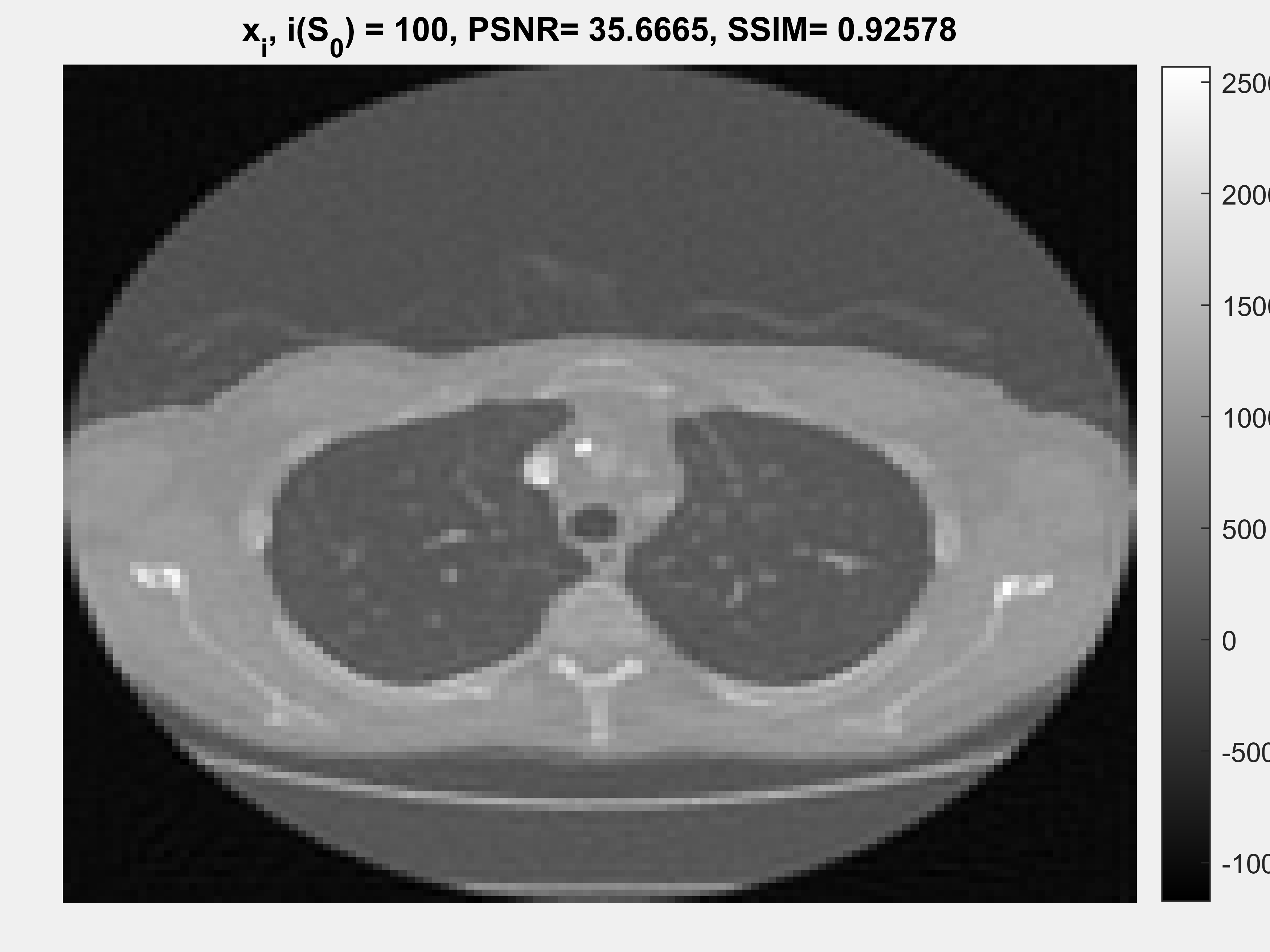}
        \caption{$x_N^{N_0}(\beta_i)$; $N_0=50, \beta_i = \beta_i(\mcal{S}_0)$}
    \end{subfigure}
    \begin{subfigure}{0.495\textwidth}
        \includegraphics[width=\textwidth]{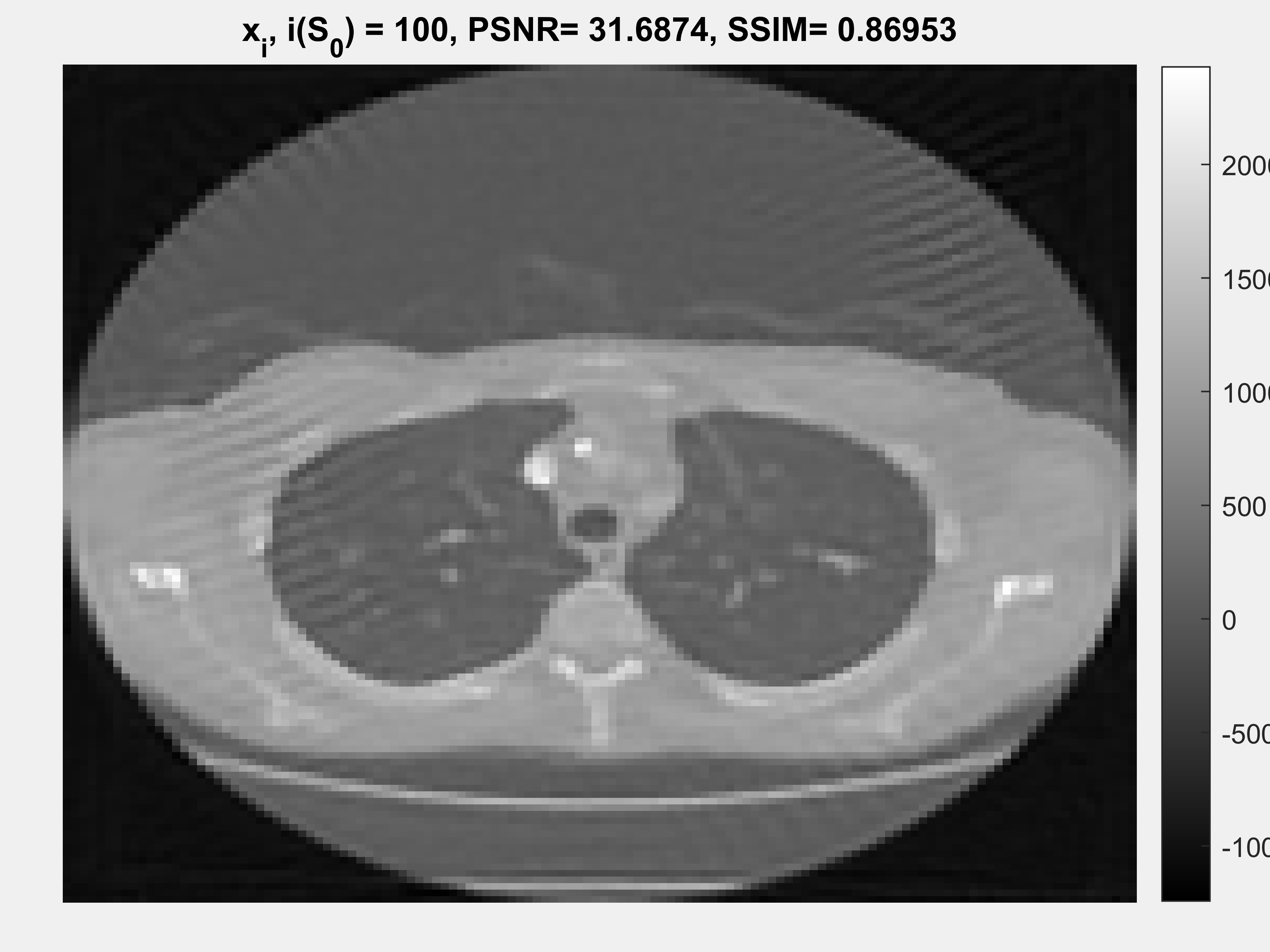}
        \caption{$x_N^{N_0}(\beta_i)$; $N_0=100, \beta_i = 1$}
    \end{subfigure}
    \begin{subfigure}{0.495\textwidth}
        \includegraphics[width=\textwidth]{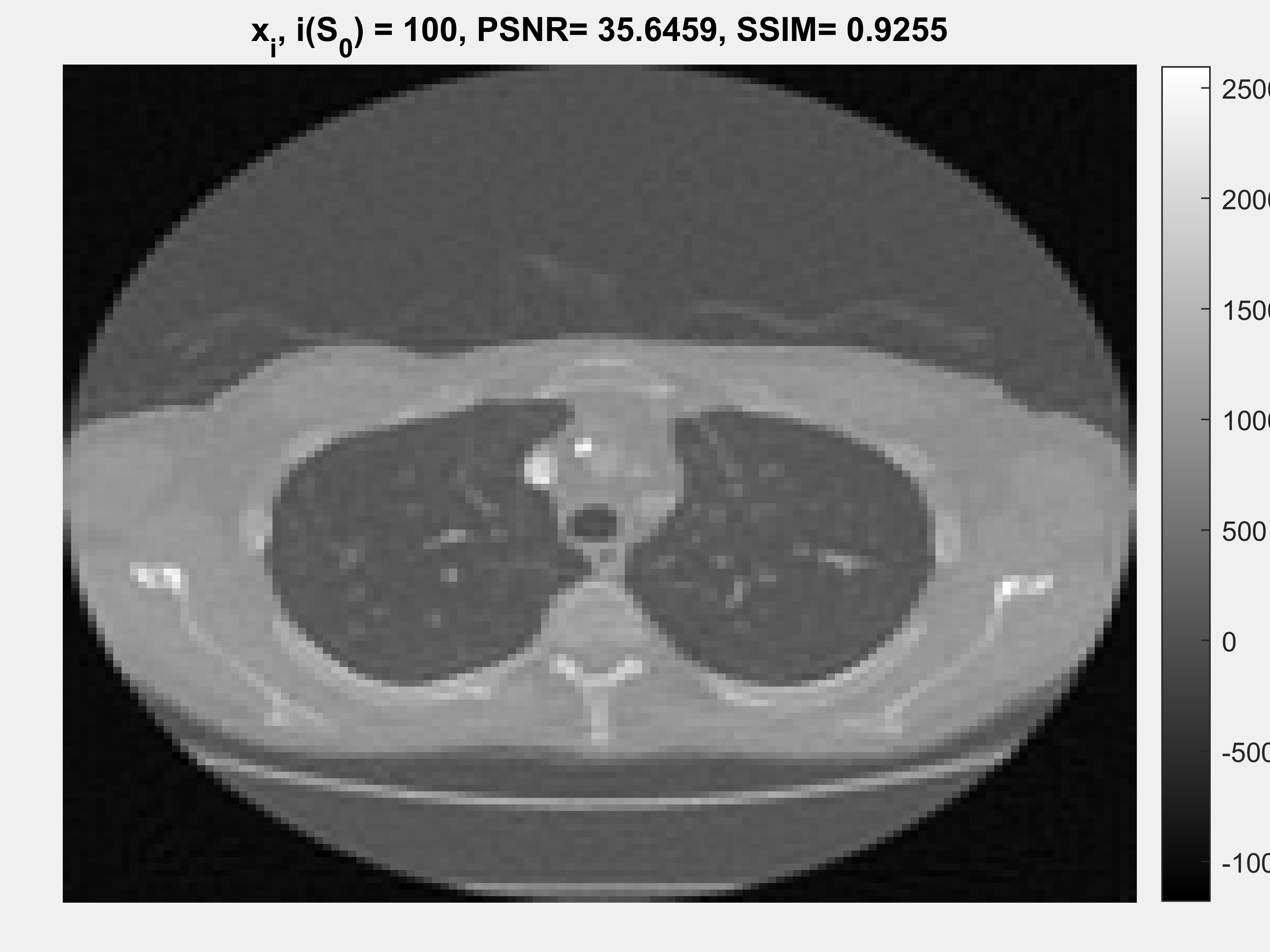}
        \caption{$x_N^{N_0}(\beta_i)$; $N_0=100, \beta_i = \beta_i(\mcal{S}_0)$}
    \end{subfigure}    
    \caption{Final solution of an unrolled reconstruction algorithm for various $R_{\alpha_i}s$' strength, with and without regularizing $R_{\theta_i}s$.} 
    \label{Fig. Final solution recovery}
\end{figure}

\begin{figure}[h!]
    \centering
    \begin{subfigure}{0.495\textwidth}
        \includegraphics[width=\textwidth]{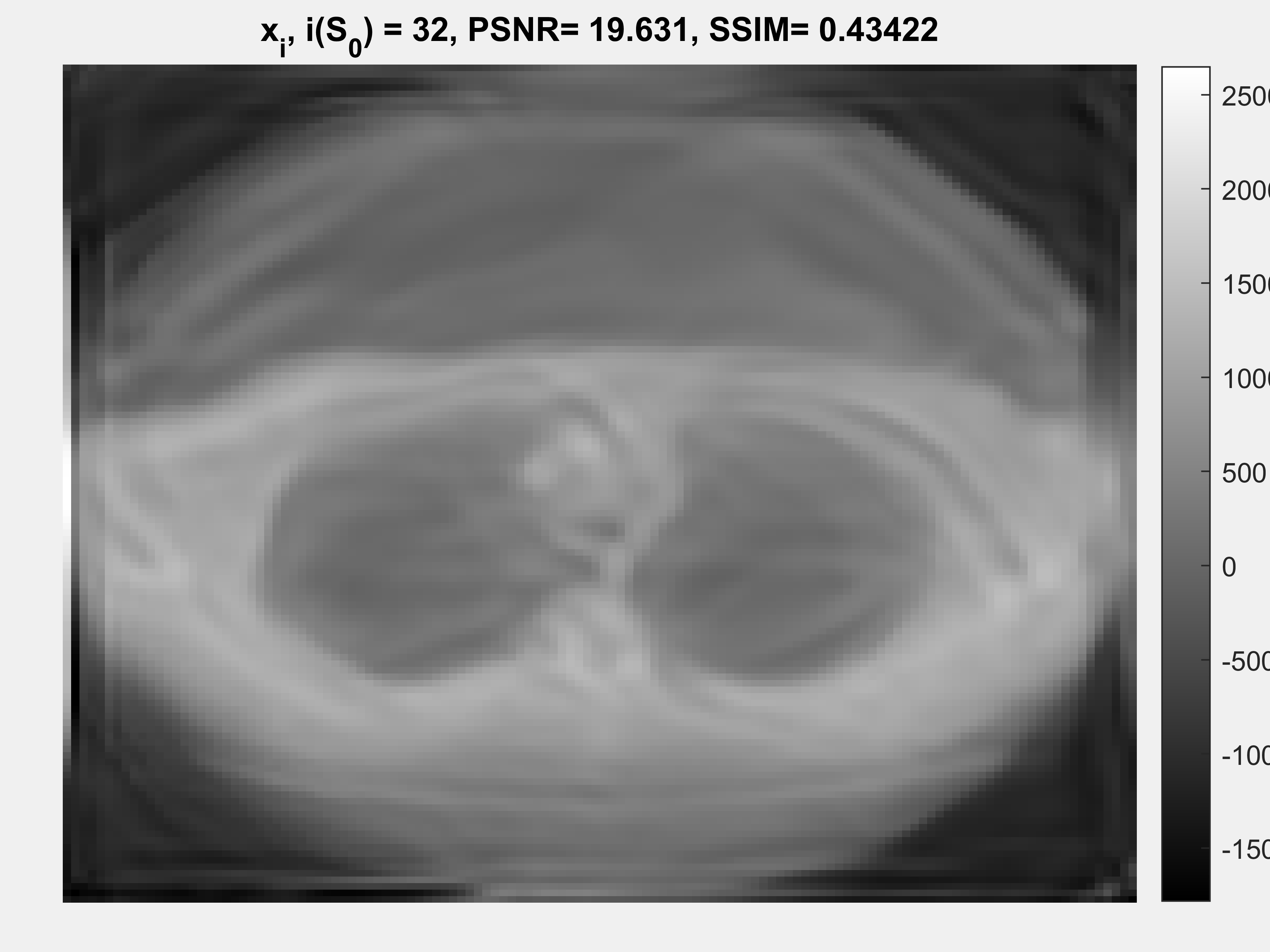}
        \caption{$x_{i(\mcal{S}_0)}^{N_0}(\beta_i)$; $N_0=1, \beta_i = 1$}
    \end{subfigure}       
    \begin{subfigure}{0.495\textwidth}
        \includegraphics[width=\textwidth]{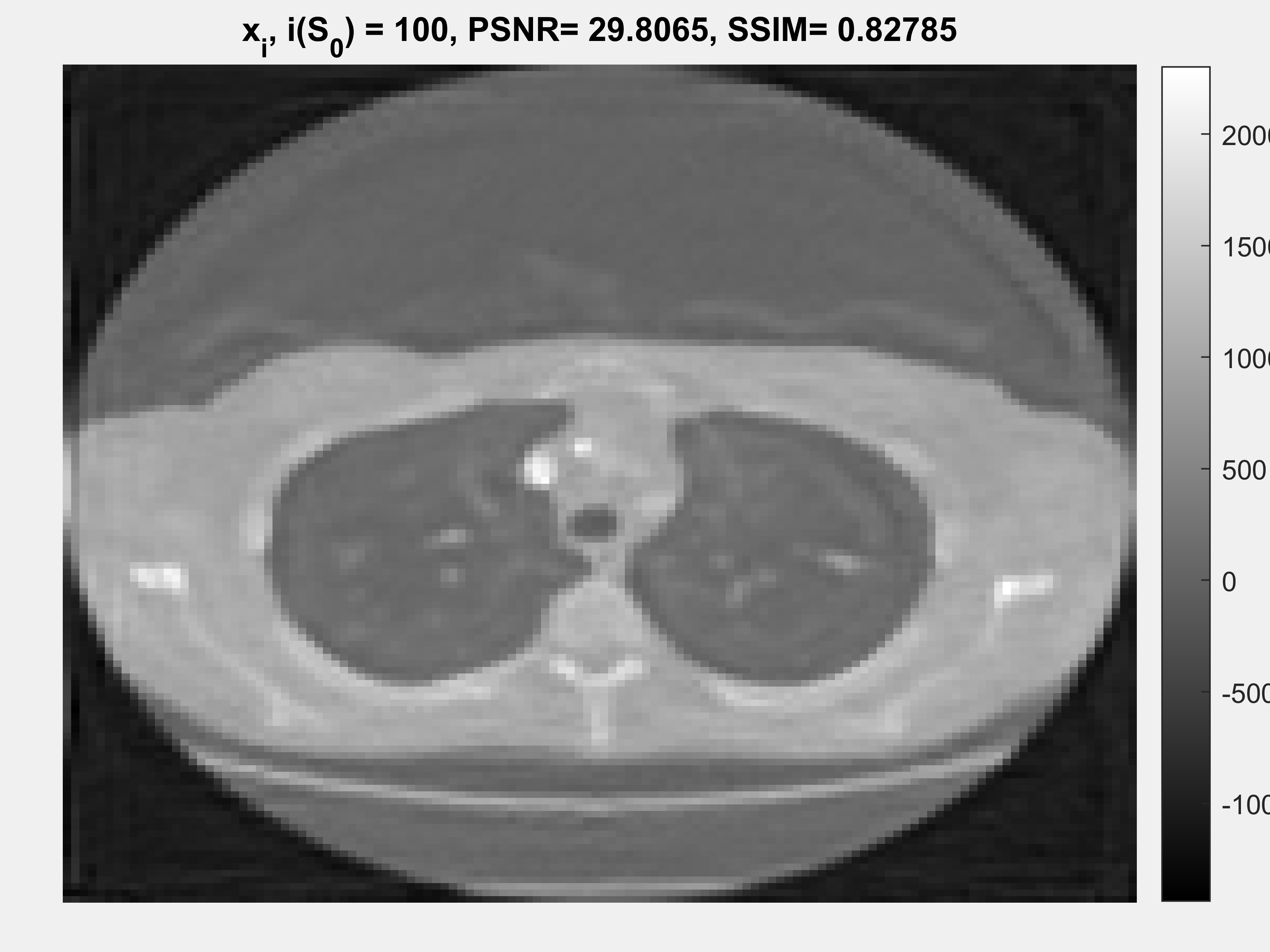}
        \caption{$x_{i(\mcal{S}_0)}^{N_0}(\beta_i)$; $N_0=1, \beta_i = \beta_i(\mcal{S}_0)$}
    \end{subfigure}       
    \begin{subfigure}{0.495\textwidth}
        \includegraphics[width=\textwidth]{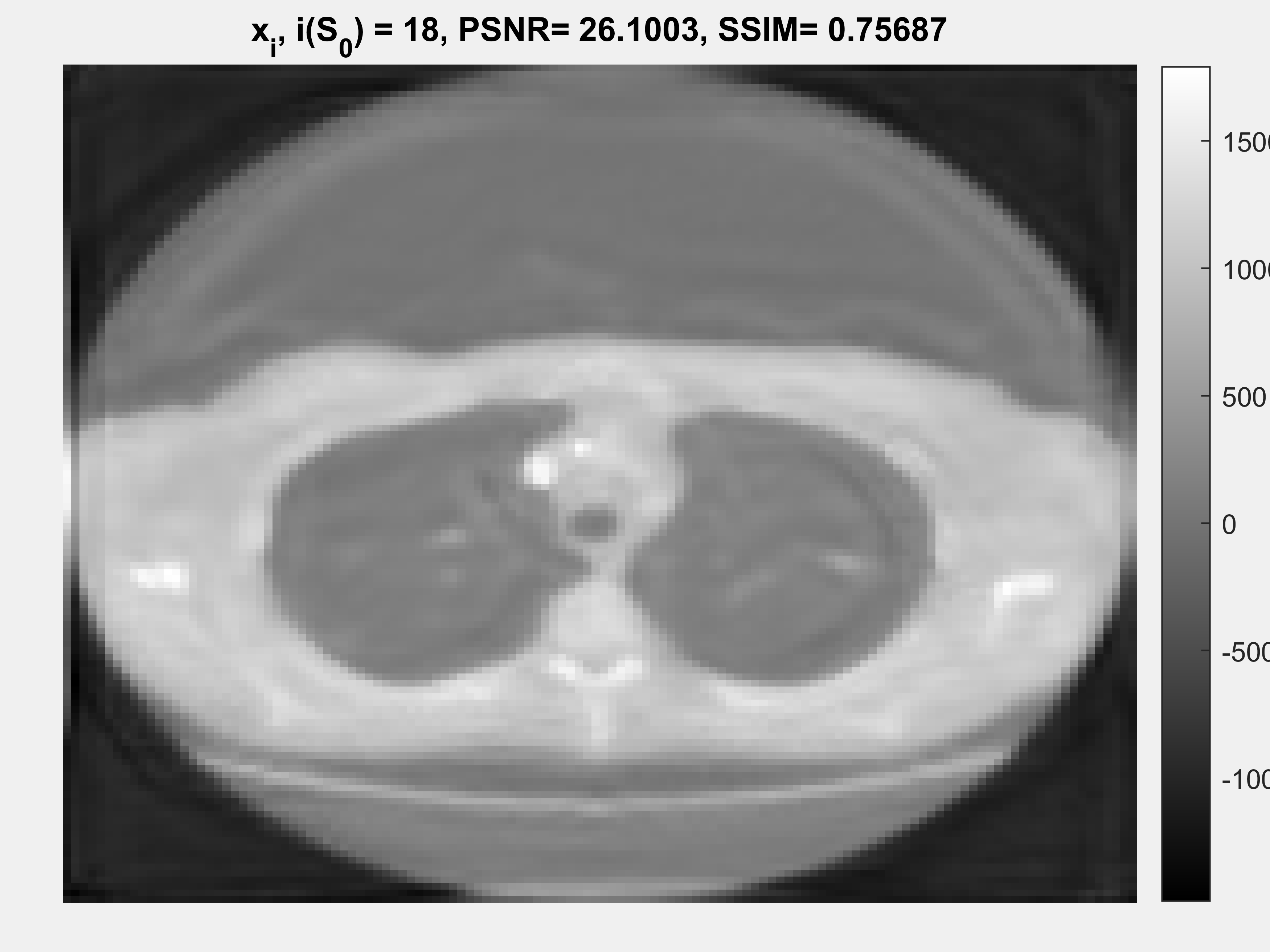}
        \caption{$x_{i(\mcal{S}_0)}^{N_0}(\beta_i)$; $N_0=10, \beta_i = 1$}
    \end{subfigure} 
    \begin{subfigure}{0.495\textwidth}
        \includegraphics[width=\textwidth]{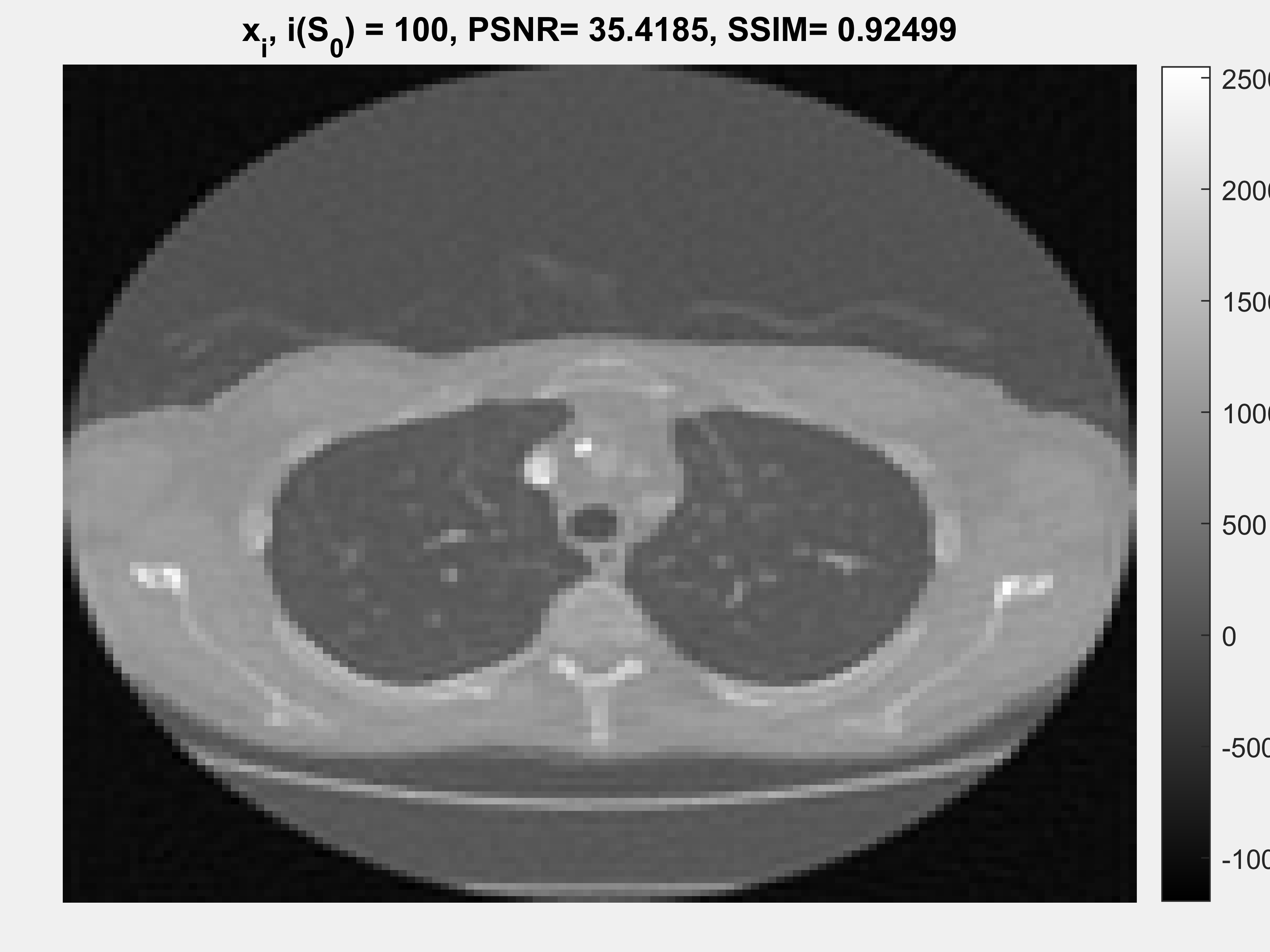}
        \caption{$x_{i(\mcal{S}_0)}^{N_0}(\beta_i)$; $N_0=10, \beta_i = \beta_i(\mcal{S}_0)$}
    \end{subfigure}    
    \begin{subfigure}{0.495\textwidth}
        \includegraphics[width=\textwidth]{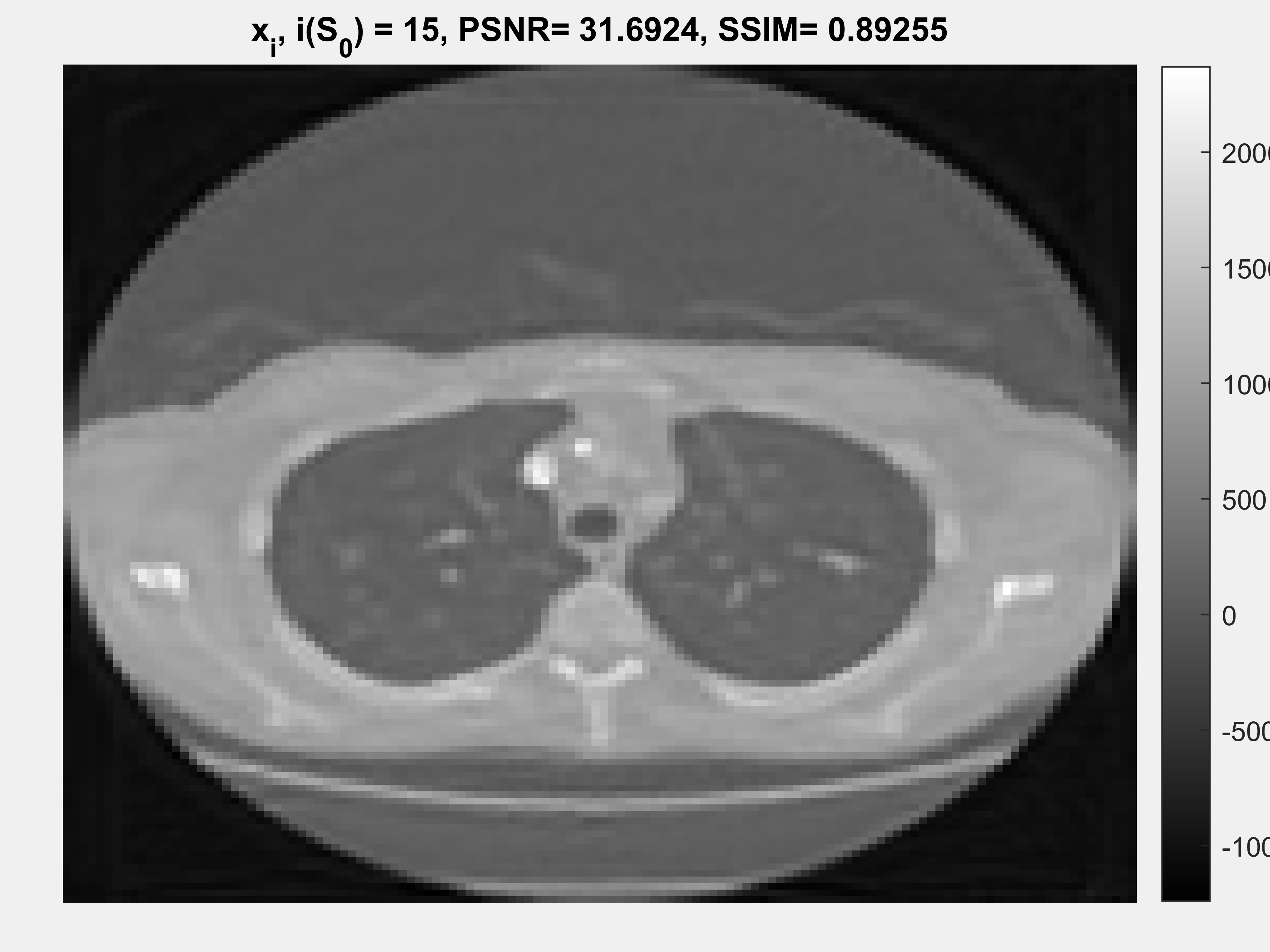}
        \caption{$x_{i(\mcal{S}_0)}^{N_0}(\beta_i)$; $N_0=50, \beta_i = 1$}
    \end{subfigure}
    \begin{subfigure}{0.495\textwidth}
        \includegraphics[width=\textwidth]{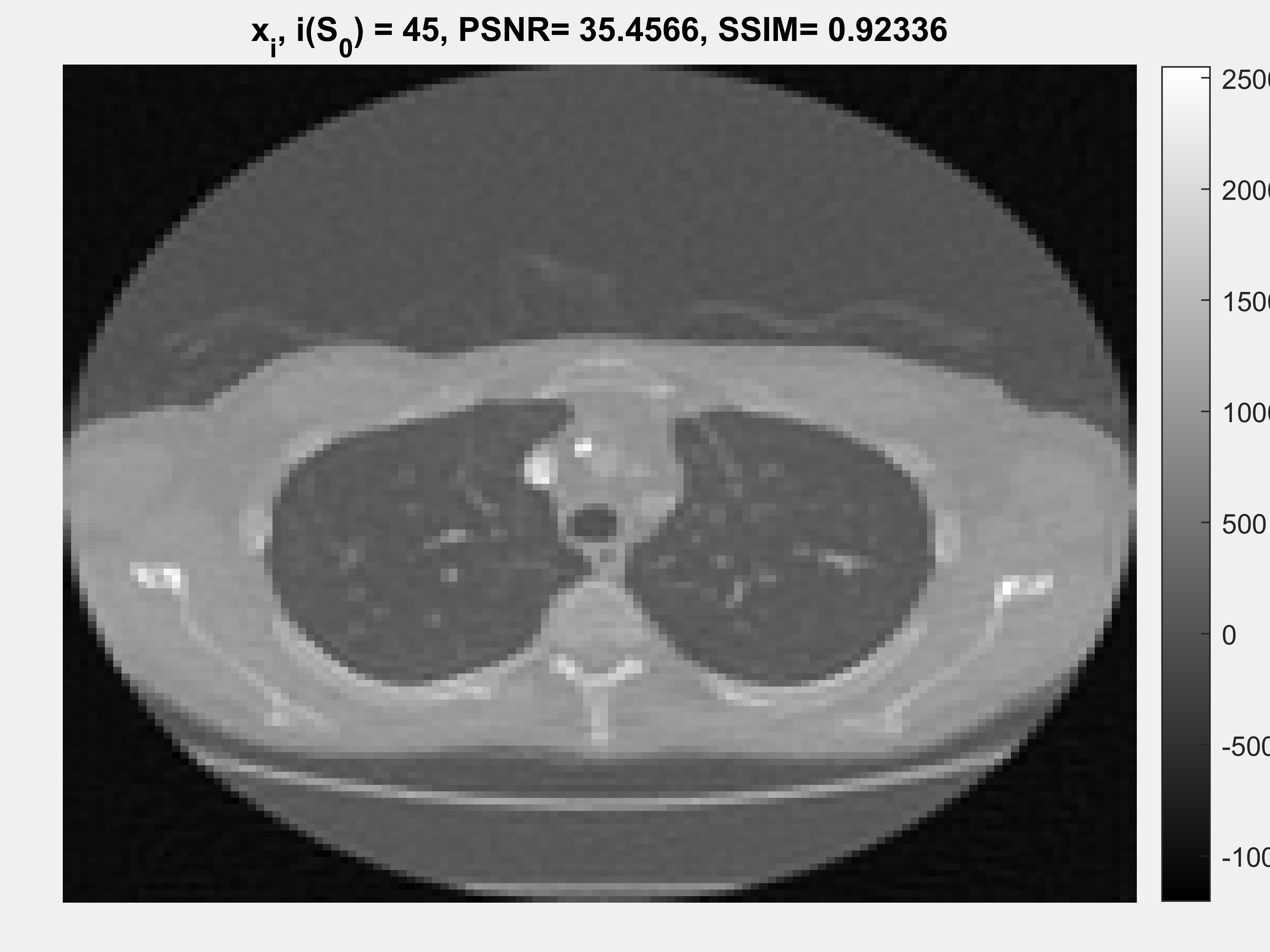}
        \caption{$x_{i(\mcal{S}_0)}^{N_0}(\beta_i)$; $N_0=50, \beta_i = \beta_i(\mcal{S}_0)$}
    \end{subfigure}
    \begin{subfigure}{0.495\textwidth}
        \includegraphics[width=\textwidth]{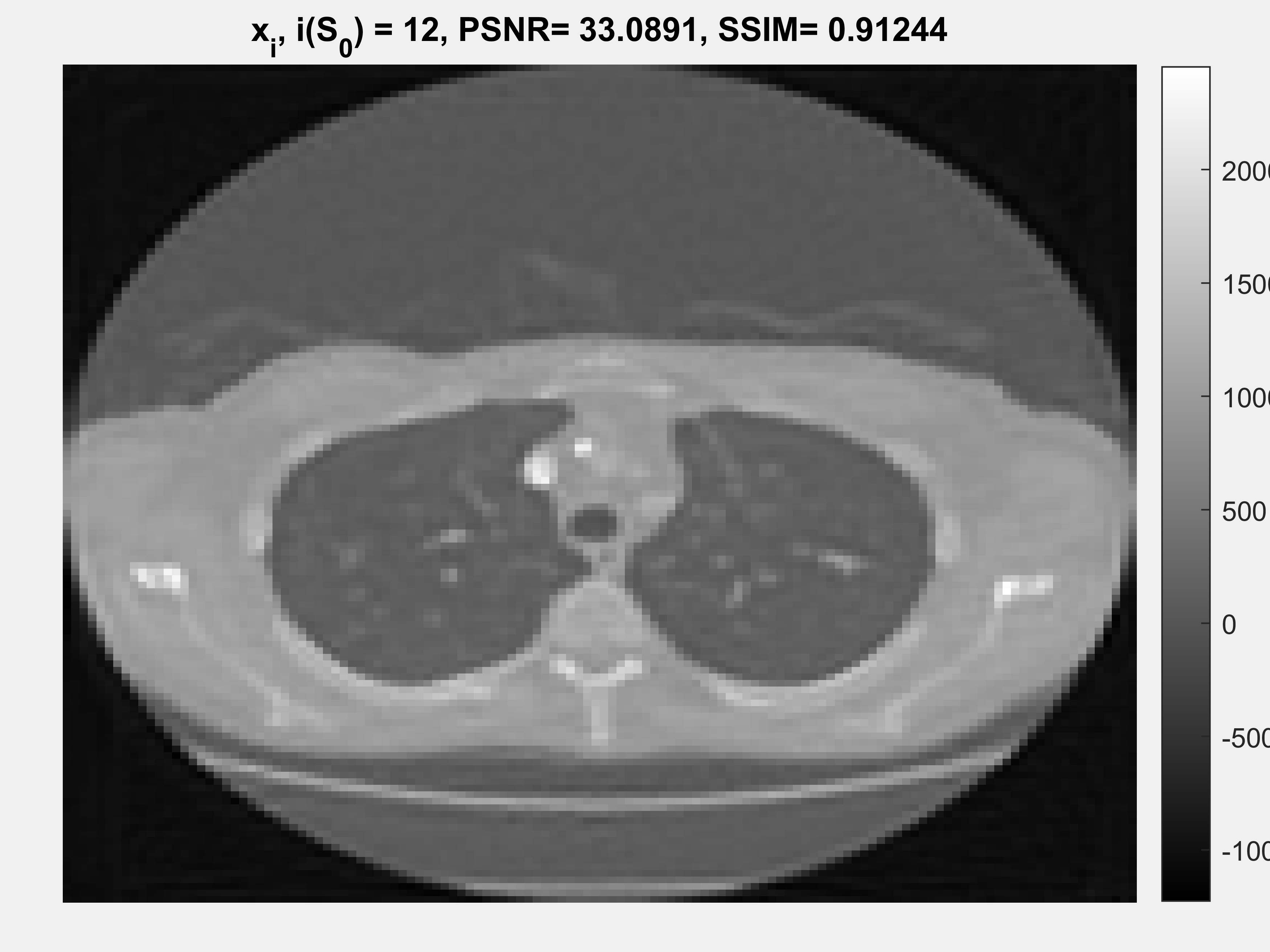}
        \caption{$x_{i(\mcal{S}_0)}^{N_0}(\beta_i)$; $N_0=100, \beta_i = 1$}
    \end{subfigure}
    \begin{subfigure}{0.495\textwidth}
        \includegraphics[width=\textwidth]{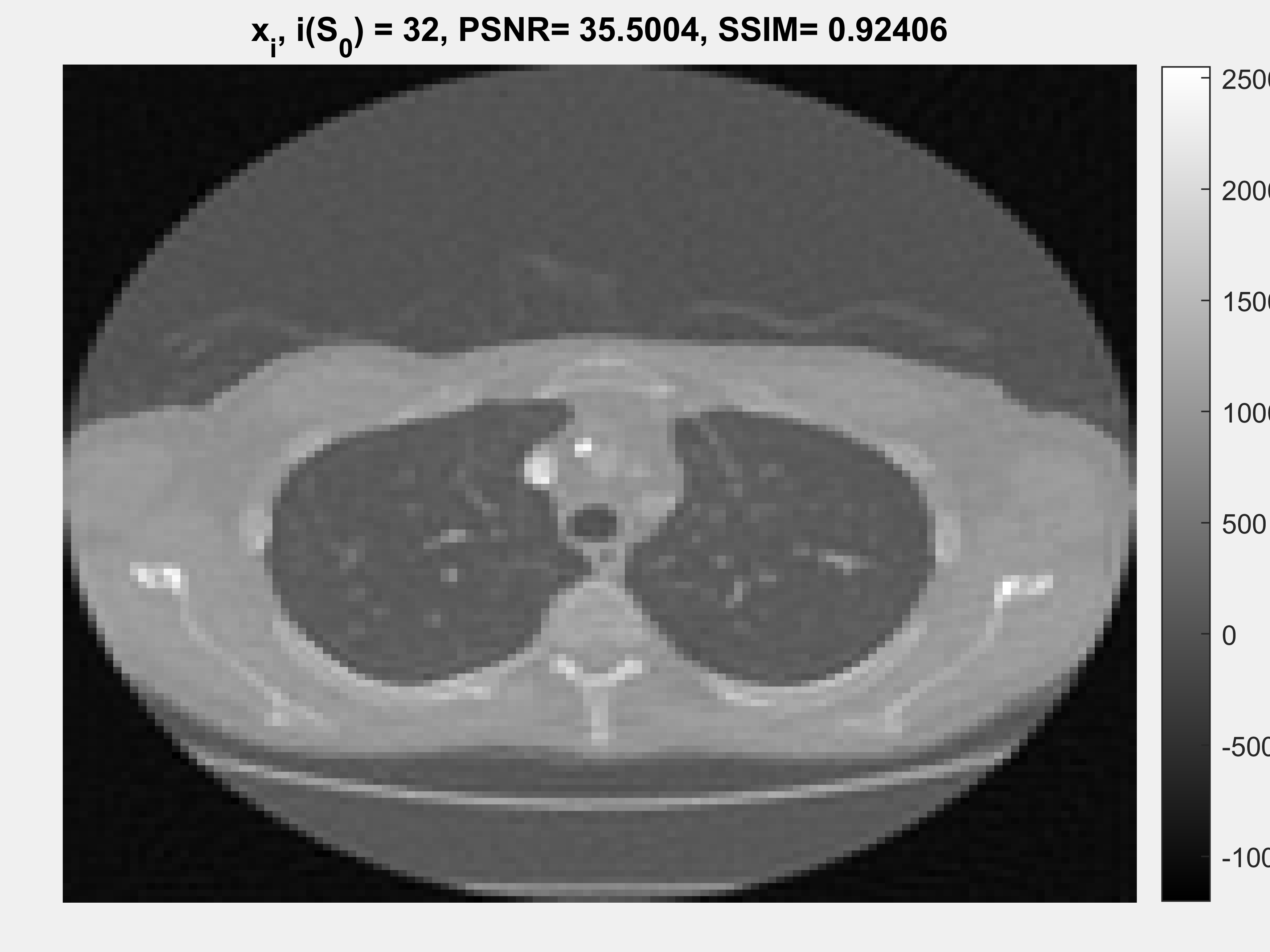}
        \caption{$x_{i(\mcal{S}_0)}^{N_0}(\beta_i)$; $N_0=100, \beta_i = \beta_i(\mcal{S}_0)$}
    \end{subfigure}    
    \caption{Selection criterion solution of an unrolled scheme for various $R_{\alpha_i}s$' strength, with and without regularizing $R_{\theta_i}s$.}
    \label{Fig. Selection criterion recovery}
\end{figure}

\begin{example}\label{Example Instabilities}
\textbf{[Breakdown of the recovery process]}

The unrolled scheme followed here is as follows, for $1 \leq i \leq N$,
\begin{equation}\label{Example Ralphai}
R_{\alpha_i}(y_\delta) := x_{N_i}^\delta(x_0^i),
\end{equation}
where, starting from an initial point $x_0^\delta = x_0^i$, for $1 \leq k \leq N_i$,
\begin{equation}\label{GD steps}
x_k^\delta = x_{k-1}^\delta - \tau_k^i T^*(Tx_{k-1}^\delta - y_\delta),
\end{equation}
for appropriate (step-sizes) $\tau_k^i \geq 0$. For simplicity, we choose a fixed number of inner iterations ($N_i = N_0$) and a constant step-size $\tau_k^i = \tau$ for all $1\leq i \leq N$ and $1 \leq k \leq N_i$, where the values of these parameters are stated below. However, the starting points ($x_0^i$s) for each inner iterative process are layer dependent (i.e., $x_0^\delta = x_0^\delta(i)$ in \eqref{GD steps}), and the nature of dependence is also defined below. Also, note that, the regularization parameter for each of these classical regularization components (data-consistency steps) is the iteration index, $\alpha_i(\delta,y_\delta) = N_i(\delta,y_\delta)$, i.e., here $R_{\alpha_i}$ is the traditional (semi-) iterative regularization method.

As for the learned components (data-denoising steps), we use MATLAB's pre-trained DnCNN denoising network ($R_{\theta_0} = \NN_{\theta_0}$), which is a substantially effective denoiser. That is, for all $1 \leq i \leq N$,
\begin{equation}
R_{\theta_i}(y_\delta) := R_{\theta_0}(R_{\alpha_i}(y_\delta))
\end{equation}

Now, the initial point ($x_0^i$) for each data-consistency step is given by, starting from a user defined $x_0^0 = x_0$ (typically, $x_0 \equiv 0$), for $1 \leq i \leq N$,
\begin{align}
& x_0^{i} = R_{\theta_i}(R_{\alpha_i}(y_\delta)), \\
& x_0^i \mapsto x_0^i = x_0^{i} + \gamma_i (x_0^i - x_0^{i-1}), \label{Momentum step}
\end{align}
where the parameters in the momentum step \eqref{Momentum step} are defined as $\gamma_i = \frac{t_{k-1} - 1}{t_k}$, $t_k = \frac{1+\sqrt{1+4t_{k-1}^2}}{2}$ and $t_0=1$. Note that, for $N_0=1$ and $N \rightarrow \infty$ we have the FISTA-PnP algorithm and for some additional tweaks in \eqref{GD steps} (or to \eqref{Example Ralphai}) one can get the ADMM-PnP algorithm. However, with the above formulation, it's neither of them, rather, it's a superposition of both these algorithms. Nevertheless, one can still have a meaningful interpretation of the recovery process and the recovered solution, according to some class (family) of regularized solutions, as defined in \cite{nayak2021PnP}. {One can also connect the above reconstruction algorithm to the operator formulation in} \eqref{Rtheta0 regularized 2}, {see Appendix} \ref{PnP Regularization Connection} {for details.}

For all the simulations, we consider (the total unrolling steps) $N = 100$, the constant step-size $\tau = 10^{-5}$ and repeated the experiment for $N_0 = \{1,10,50,100\}$, the total number of inner iterations. Here, we choose the parameter choice criterion ($\mcal{S}_0$) as the Cross-Validation errors, for a fixed leave out set $y_{\delta,s} \subset y_\delta$, where $s \subset \{1,2,\cdots,m\}$ and $|s| = 0.01m$. That is, when performing regularization at each unrolled step, the values of $\beta_i$ in \eqref{Regularized Unrolling 1} is determined by the selection criterion $\mcal{S}_0$, i.e., $\beta_i = \beta_i(\mcal{S}_0)$, such that
\begin{equation}\label{betaS0}
\beta_i(\mcal{S}_0) := \argmin_\beta \; ||Tf_i(\beta,R_{\alpha_i},R_{\theta_i};y_\delta) - y_{\delta,s}||_2,
\end{equation}
where $f_i(.)$ is as defined in \eqref{Regularized Unrolling 1}. Note that, the minimization problem \eqref{betaS0} may not be convex, and hence, can be difficult to find the global minimum. Still, even for a local minimizer, the regularized iterate $f_i(\beta_i(\mcal{S}_0),R_{\alpha_i},R_{\theta_i};y_\delta)$ can be much better (in the sense of satisfying the selection criterion $\mcal{S}_0$) than the unregularized iterate $R_{\theta_i}(y_\delta)$, in addition to regularizing $R_{\theta_i}s$, when $R_{\theta_i}(y_\delta)$ suffers from instabilities. This can be seen in the presented examples. Furthermore, \eqref{betaS0} is a single variable optimization problem, and hence, not much computationally expensive to solve.

 Now, although one can also perform further regularization in the data-consistency steps by defining a regularized solution as
\begin{equation}
R_{\alpha_i}(y_\delta) := x^\delta_{k(\mcal{S}_0)}(x_0^i),
\end{equation}
instead of $x^\delta_{N_i}(x_0^i)$, we opted out of it, as it does not make much difference here.

Figure \ref{Fig. Final solution recovery} shows the final recovered solutions ($x^{N_0}_N$, where $N=100$) for various values of $N_0$ and Figure \ref{Fig. Selection criterion recovery} shows the iterates ($x^{N_0}_{i(\mcal{S}_0)}$) which best satisfies the selection criterion $\mcal{S}_0$ during the unrolled iterative process, i.e.,
\begin{equation}\label{Unrolled S0 solution}
x^{N_0}_{i(\mcal{S}_0)} := \{ x_{i_0}^{N_0}: x_i^{N_0} := R_{\theta_i}(y_\delta) \mbox{ and } i_0 = \argmin_{1 \leq i \leq N} \mcal{S}_0(x_i^{N_0})\}
\end{equation}
The performance metrics for each of these recoveries are stated at the top of the figures, where PSNR stands for the peak signal-to-noise ratio, SSIM stands for the structure similarity index measure and $i(\mcal{S}_0)$ indicates the index for which $x_i^{N_0}$ best satisfies $\mcal{S}_0$ or the last iterate when $i(S_0) = N$. 

Figure \ref{Fig. Final solution recovery} reflects that, the final reconstructed image ($x_N^{N_0}$) in an unrolled scheme may not always be stable, without regularizing $R_{\theta_i}$ (i.e., $\beta_i=1$ in \eqref{Regularized Unrolling 2}), however, it can improve when the classical (regularized) components ($R_{\alpha_i}s$) are improved ($N_0$ increases). In other words, in this case, the learned components ($R_{\theta_i}s$) are very strong, and hence, dominate the reconstruction algorithm for weaker classical components, leading to an unstable recovery process. Note that, the instability in the solution (for weaker $R_{\alpha_i}s$ and $\beta_i=1$) is quite different than the instabilities of the ill-posed inverse problem, which are noisy features and streaky artifacts, as can be seen when $R_{\alpha_i}s$ get stronger. Where as, for the same classical regularization components $R_{\alpha_i}s$, the final solution corresponding to regularized $R_{\theta_i}s$ is quite stable and well denoised, where the quality improves when $R_{\alpha_i}s$ improve.

Figure \ref{Fig. Selection criterion recovery} presents the iterates obtained during the recovery process which best satisfies the selection criterion $\mcal{S}_0$, i.e., $x^{N_0}_{i(\mcal{S}_0)}$ as defined in \eqref{Unrolled S0 solution}, for various values of $N_0$. Here, one can observe that  even without regularizing $R_{\theta_i}s$, one can still recover fine solutions (much better than its final counterparts $x_N^{N_0}$), of course, the quality improves upon regularizing $R_{\theta_i}s$. Note that, there is not much difference in $x_{i(\mcal{S}_0)}^{N_0}$ over $x_N^{N_0}$, when using regularized $R_{\theta_i}s$, in fact, the later is even slightly better than the former, in terms of the evaluation metrics. This suggests that one can consider $x_N^{N_0}$ as the recovered solution, over $x_{i(\mcal{S}_0)}^{N_0}$, when using regularized $R_{\theta_i}s$ and the converse, otherwise.

\begin{figure}[h!]
    \centering
    \begin{subfigure}{0.495\textwidth}
        \includegraphics[width=\textwidth]{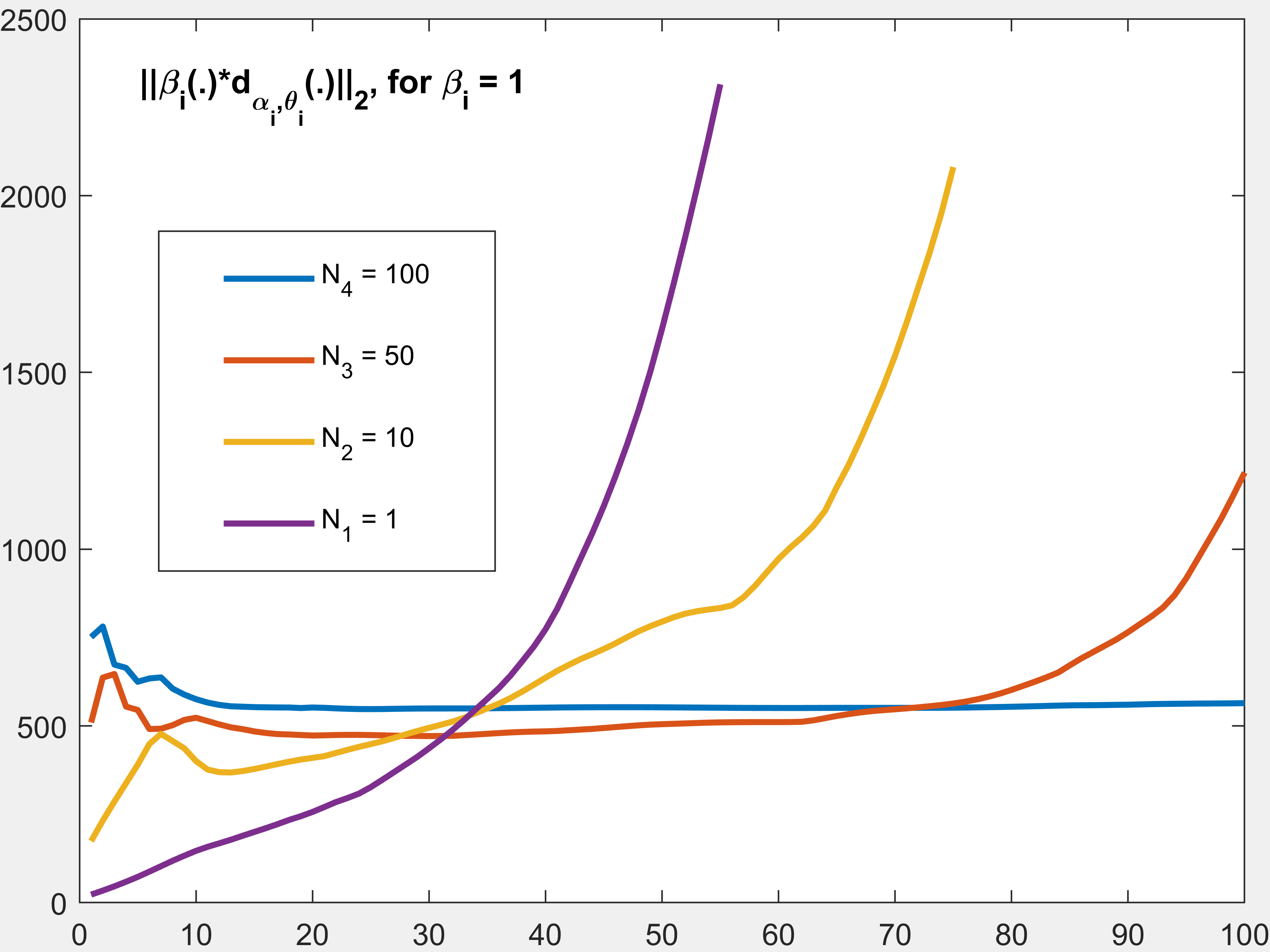}
        \caption{$||d_{\alpha_i,\theta_i}^\delta(y_\delta)||_2$, No regularization}
        \label{e_1_NR}
    \end{subfigure}       
    \begin{subfigure}{0.495\textwidth}
        \includegraphics[width=\textwidth]{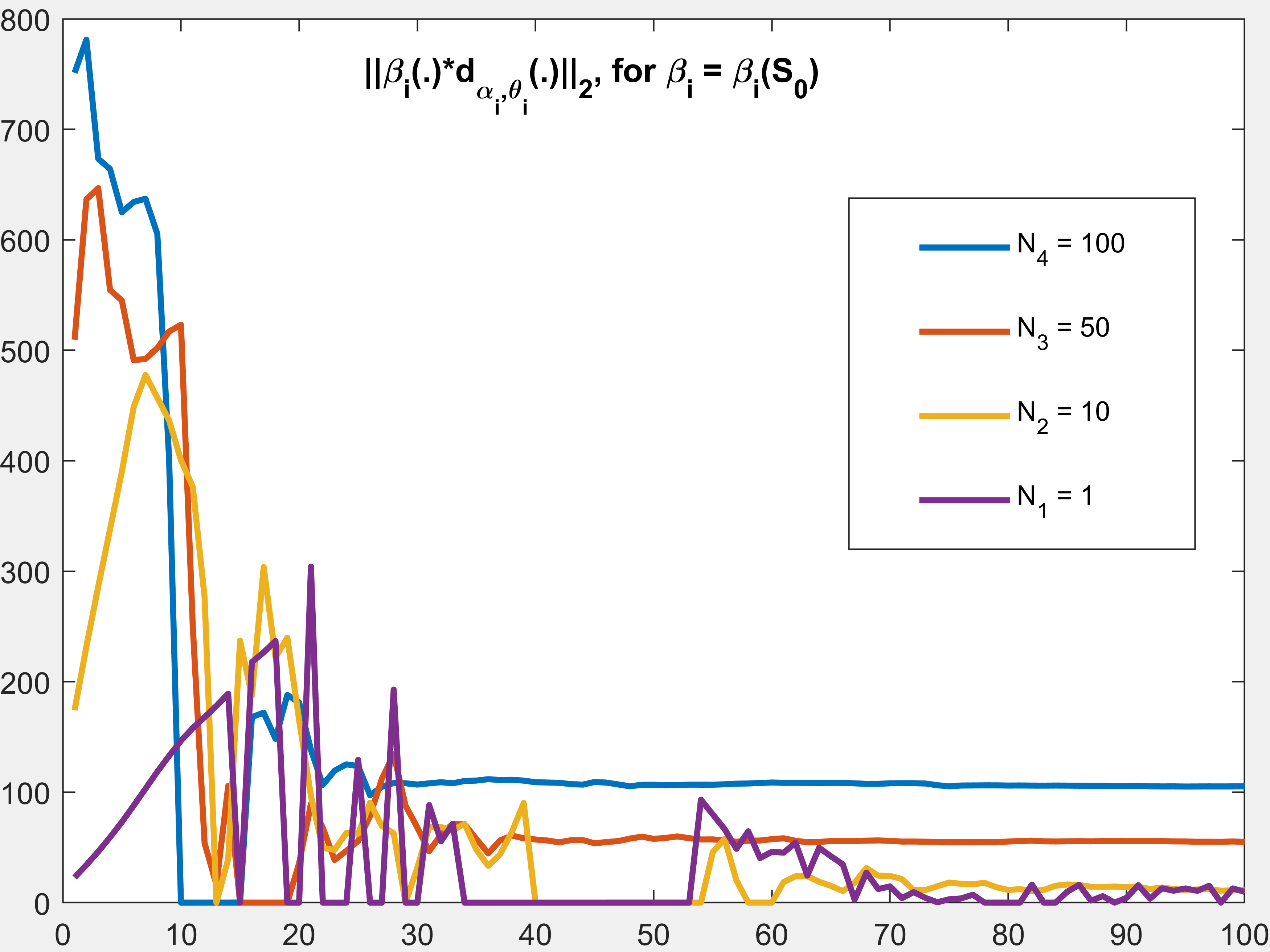}
        \caption{$||\beta_i(\mcal{S}_0,y_\delta) d_{\alpha_i,\theta_i}^\delta(y_\delta)||_2$, Regularization}
        \label{e_1_R}
    \end{subfigure} 
    \begin{subfigure}{0.495\textwidth}
        \includegraphics[width=\textwidth]{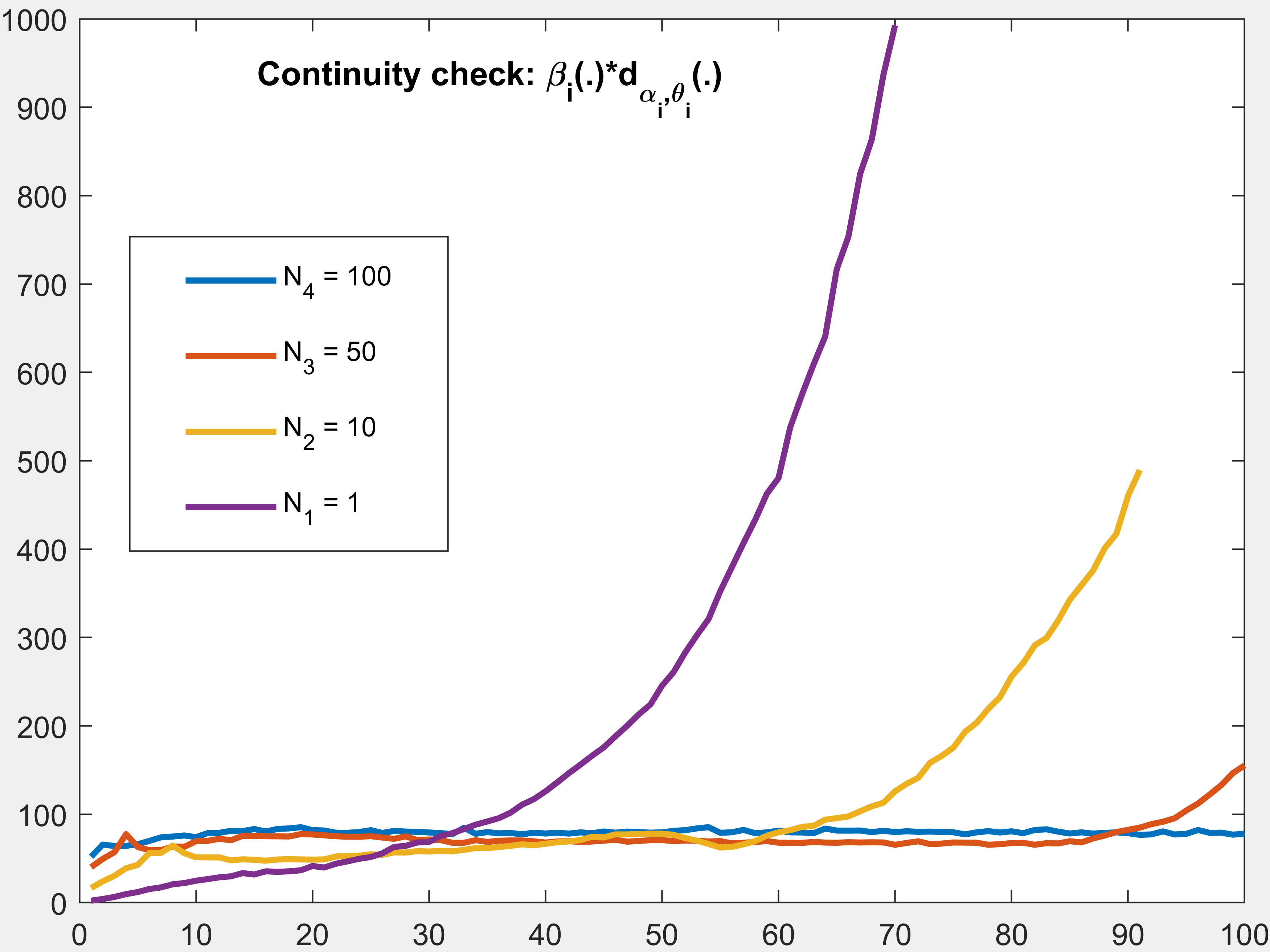}
        \caption{Discontinuity, without regularization}
        \label{ContiCheck_NR}
    \end{subfigure}       
    \begin{subfigure}{0.495\textwidth}
        \includegraphics[width=\textwidth]{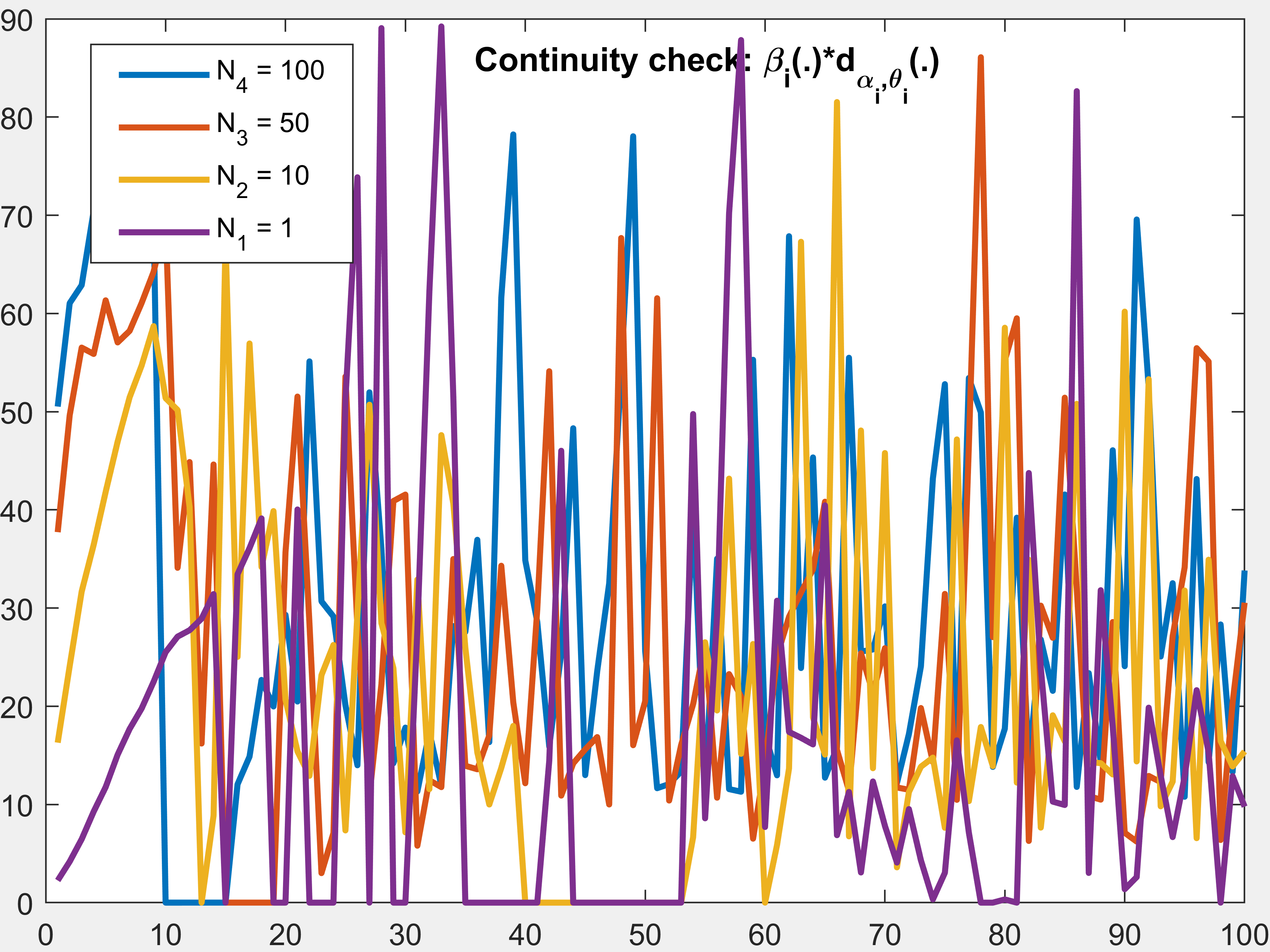}
        \caption{Continuity, with regularization}
        \label{ContiCheck_R}
    \end{subfigure}     
    \begin{subfigure}{0.495\textwidth}
        \includegraphics[width=\textwidth]{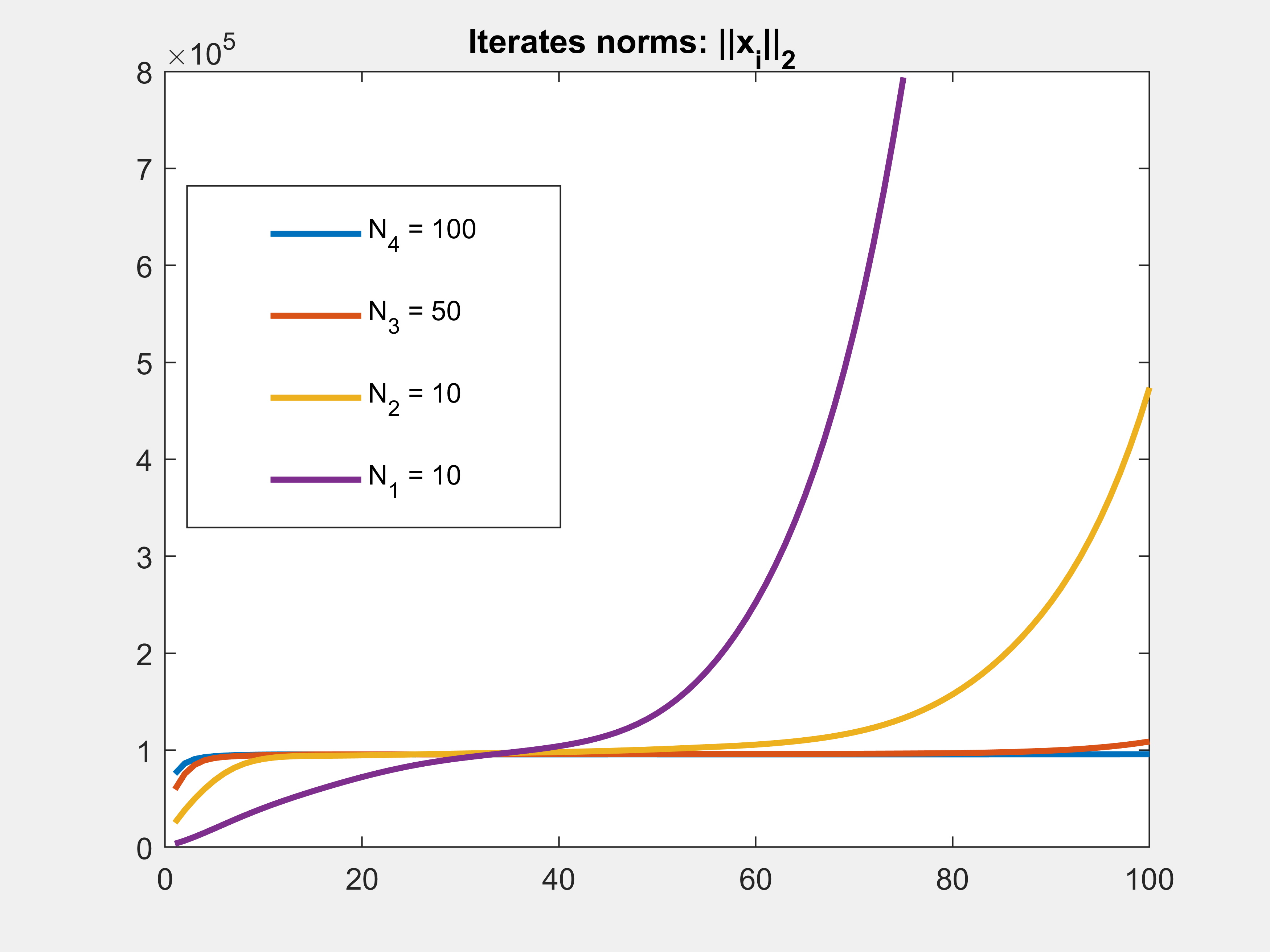}
        \caption{Iterates norm $||x_i^{N_j}||_2$}
        \label{Xnorm2}
    \end{subfigure}       
    \begin{subfigure}{0.495\textwidth}
        \includegraphics[width=\textwidth]{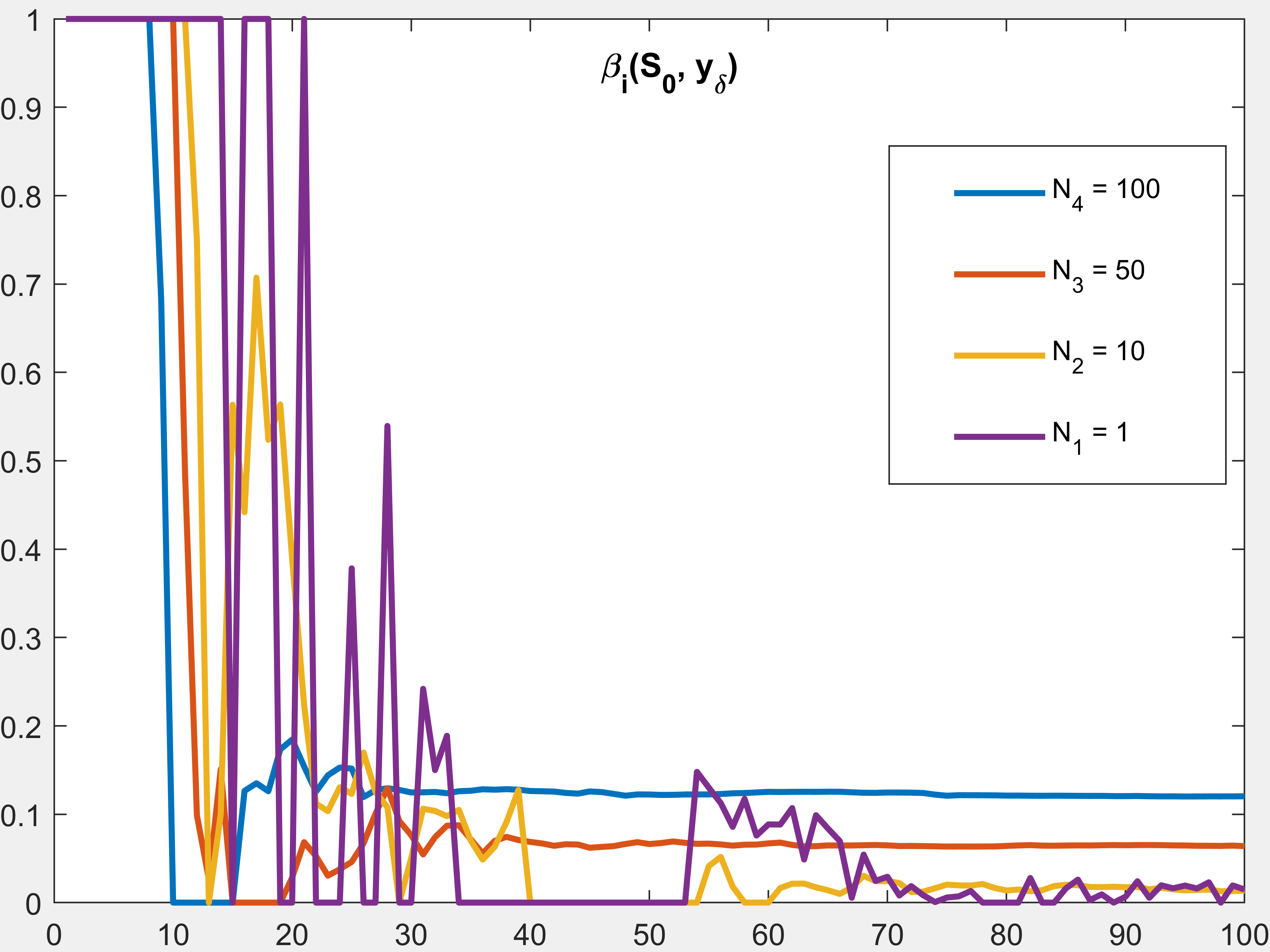}
        \caption{$R_{\theta_i}$ regularization parameter $\beta_i(\mcal{S}_0,y_\delta)$}
        \label{gamma_i}
    \end{subfigure}     
    \begin{subfigure}{0.495\textwidth}
        \includegraphics[width=\textwidth]{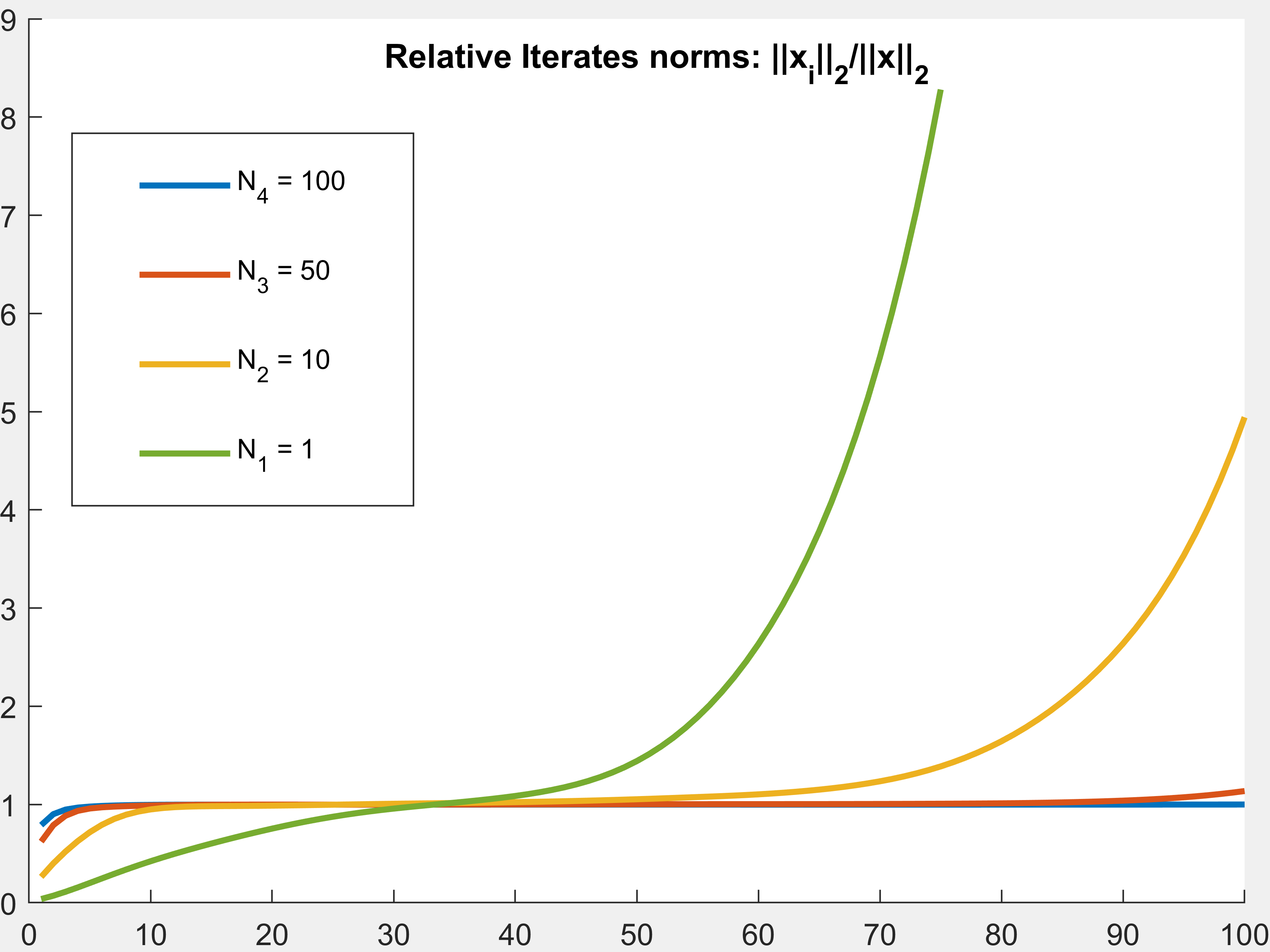}
        \caption{Relative norms $\frac{||x_i^{N_j}||_2}{||\hat{x}||_2}$, Unregularized}
        \label{RelXnorm2}
    \end{subfigure}       
    \begin{subfigure}{0.495\textwidth}
        \includegraphics[width=\textwidth]{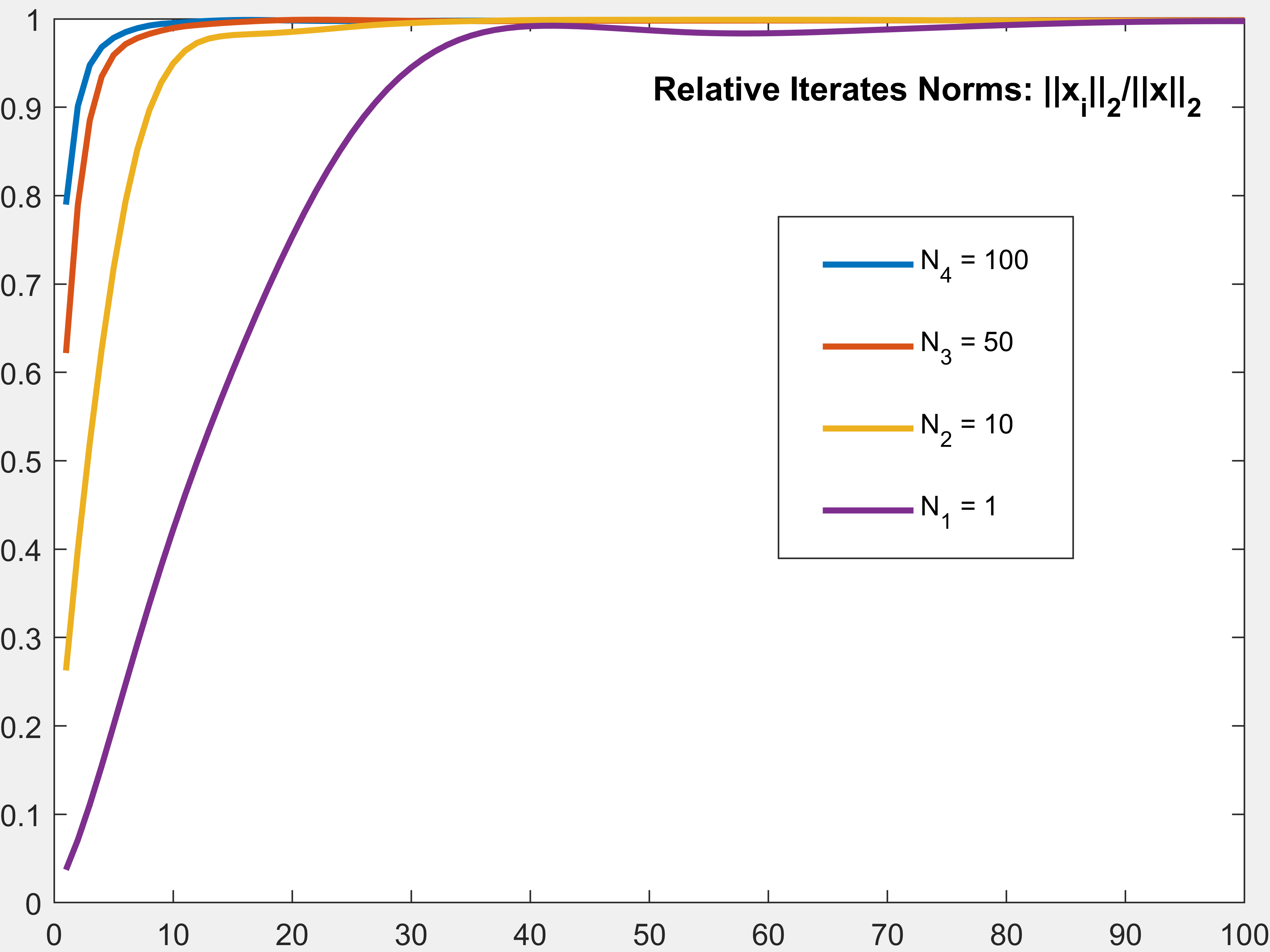}
        \caption{Relative norms $\frac{||x_i^{N_j}||_2}{||\hat{x}||_2}$, Regularized}
        \label{RelXnorm2_R}
    \end{subfigure}    
\end{figure}

The natural question that one can ask is, when does the recovery fail? The answer to this question is not straightforward, otherwise, the ill-posed problem won't be ill-posed anymore. However, one can have certain measuring/monitoring criteria to (at least) indicate when an algorithm might fail. Here, we present few of such criteria to indicate when an algorithm can break down. Note that, from \eqref{Post-Processing Regularization 2}, the recovery process $R_{\alpha_i}^{\theta_i}$ (at each unrolled step) is continuous if $\beta_i(\alpha_i(\delta,y_\delta),.)d_{\alpha_i,\theta_i}^\delta(.)$ is continuous, as a function of $y_\delta$. Again, this is not easy to verify for $\beta_i(.,y_\delta) = \beta_i(\mcal{S}_0,y_\delta)$ and for a pre-trained $R_{\theta_i}$, which can be non-linear. Now, although we don't have a theoretical proof, we empirically try to show that $\beta_i(\mcal{S}_0,.)d_{\alpha_i,\theta_i}^\delta(.)$ is ``continuous" here. First, we show that (in Figure \ref{e_1_NR}) for $\beta_i = 1$ (unregularized $R_{\theta_i}s$), the function $g(i) = ||\beta_i(.,y_\delta)d_{\alpha_i,\theta_i}^\delta(y_\delta)||_2$ tends to be unbounded ($g(i)$ increases exponentially) as $i$ increases, especially for smaller values of $N_0$ (weaker $R_{\alpha_i}$), indicating an unstable recovery process. Where as, the function $g(i)$ does not blow up for $\beta_i = \beta_i(\mcal{S}_0,y_\delta)$ (regularized $R_{\theta_i}s$), see Figure \ref{e_1_R}. However, this might not guarantee continuity or discontinuity, for $\beta_i=\beta_i(\mcal{S}_0,y_\delta)$ and $\beta_i=1$, respectively, as $\beta_i(.,y)d_{\alpha_i,\theta_i}^\delta(y)$ may be non-linear. Hence, in our second attempt we try to verify the continuity of $\beta_i(.,y)d_{\alpha_i,\theta_i}^\delta(y)$ directly by computing 
\begin{equation}
g(i) = ||\beta_i(.,y_\delta)d_{\alpha_i,\theta_i}^\delta(y_\delta) - \beta_i(.,y_\delta')d_{\alpha_i,\theta_i}^\delta(y_\delta')||_2,
\end{equation}
where $y_{\delta,j}' = y_{\delta,j} + \epsilon_j'$, $\epsilon_j'$ is a Gaussian noise with mean zero and std. dev. equaling $\max(|y_\delta|)/1000$, and $d_{\alpha_i,\theta_i}^\delta(y_\delta')$ is computed according to the definition in \eqref{Post-Processing direction}. Again, from Figure \ref{ContiCheck_NR}, one can observer that, the values of $g(i)$ are (relatively) high for unstable recovery processes (when $N_0$ is small and unregularized $R_{\theta_i}s$) and (relatively) low for stable recovery process, in Figure \ref{ContiCheck_R}, especially when $R_{\theta_i}$s' are regularized (irrespective of $R_{\alpha_i}$s' strength), indicating that $\beta_i(.,y)d_{\alpha_i,\theta_i}^\delta(y)$ might be continuous for $\beta_i(.,y) = \beta_i(\mcal{S}_0,y)$. 

Another indicator, for an unstable recovery process, can be the iterates' norms during the iterative process, i.e., if $||x_i||_2$ blows up exponentially, then the recovery algorithm has become unstable. This can be seen in Figure \ref{Xnorm2} or in Figure \ref{RelXnorm2}, which shows the relative norms $\left( r_i := \frac{||x_i^{N_0}||_2}{||\hat{x}||_2} \right)$ of the iterates $x_i^{N_0}$ for various values $N_0$, i.e., for different $R_{\alpha_i}$s' strength. As expected, the ratio $r_i$ saturates to one for stable recovery processes, Figure \ref{RelXnorm2_R}, and explodes beyond one for unstable processes, Figure \ref{RelXnorm2}. In addition, to understand how an unstable process becomes stable, upon regularizing it via the parameter $\beta_i(\mcal{S}_0,y_\delta)$, we plot the values of $\beta_i(\mcal{S}_0,y_\delta)$ over the iterations, see Figure \ref{gamma_i}. Here, one should also expect that, for large values of $r_is$ (indicating instabilities in those iterates), the corresponding values of $\beta_is$ should be small, to compensate it. This counter balancing effect, small $\beta_i$ values for large $r_i$ values, can be seen by comparing Figure \ref{Xnorm2} ($r_i$ values) with Figure \ref{gamma_i} ($\beta_i$ values)

\begin{figure}[h!]
    \centering
    \begin{subfigure}{0.495\textwidth}
        \includegraphics[width=\textwidth]{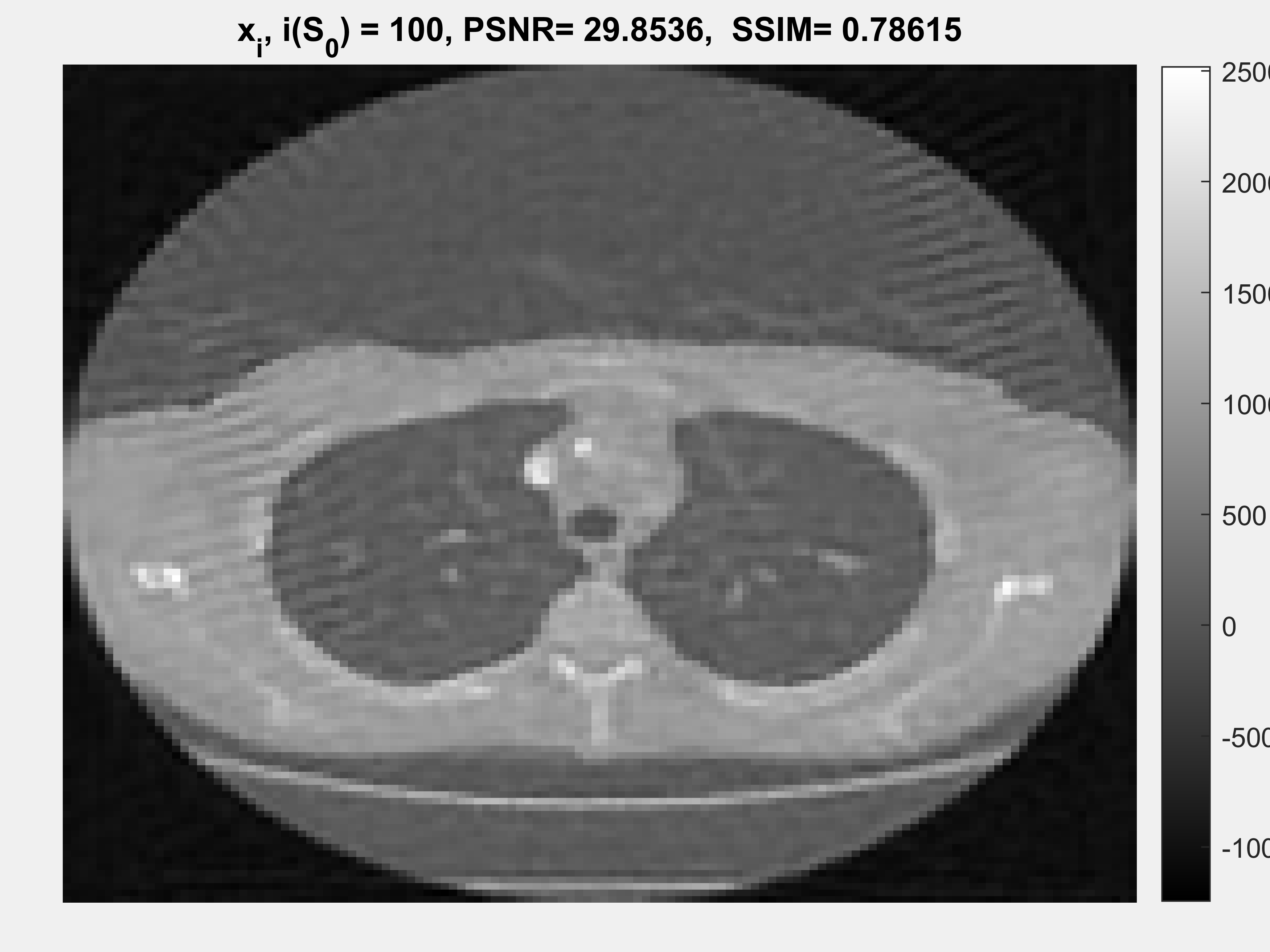}
        \caption{$x_{N}^{N_0}(\beta_i)$; $N=N_0=100, \beta_i = 1$}
        \label{Fig. high noise NR}
    \end{subfigure}       
    \begin{subfigure}{0.495\textwidth}
        \includegraphics[width=\textwidth]{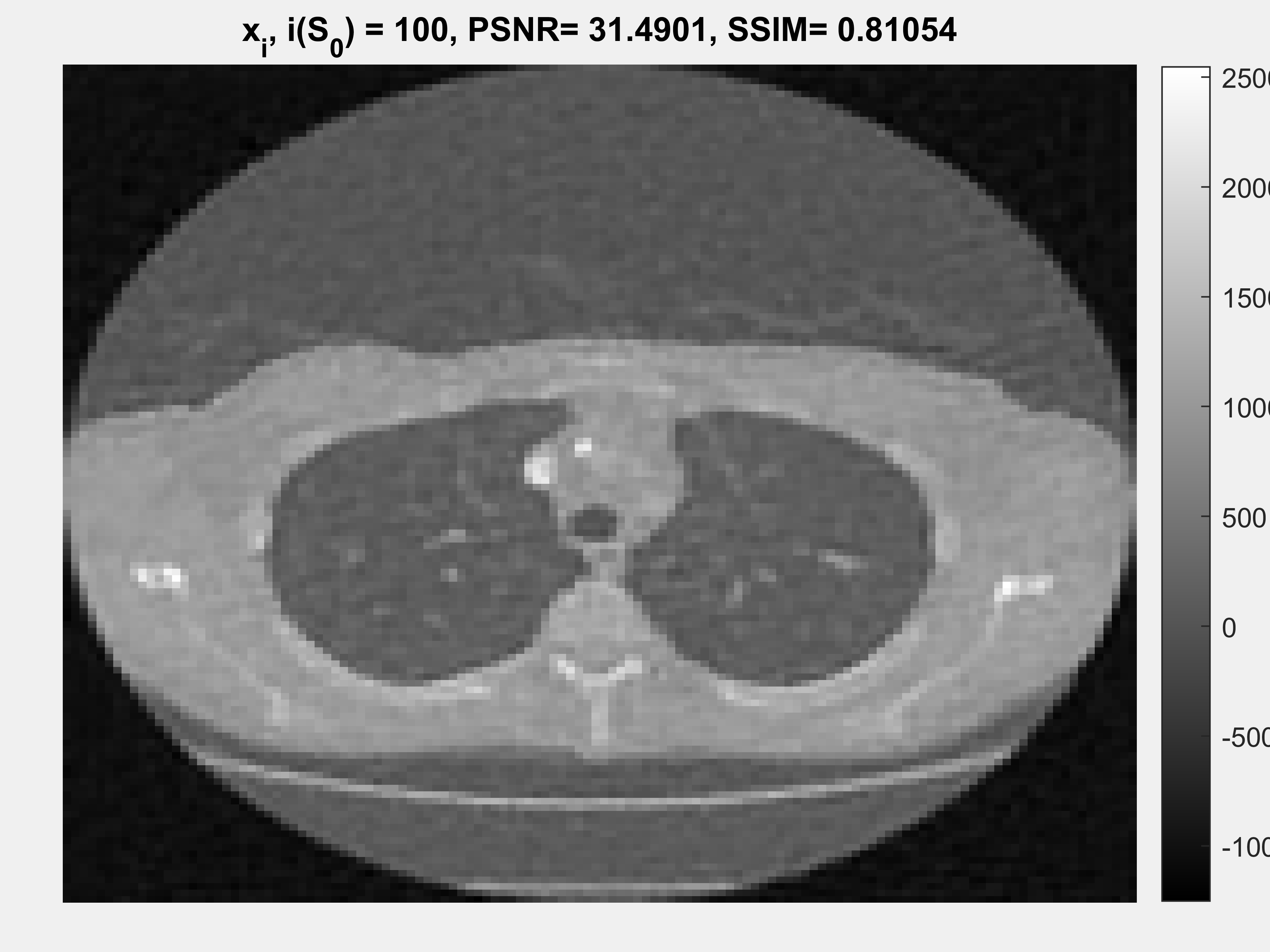}
        \caption{$x_{N}^{N_0}(\beta_i)$; $N=N_0=100, \beta_i = \beta_i(\mcal{S}_0,y_\delta)$}
        \label{Fig. high noise R}
    \end{subfigure}       
    \begin{subfigure}{0.495\textwidth}
        \includegraphics[width=\textwidth]{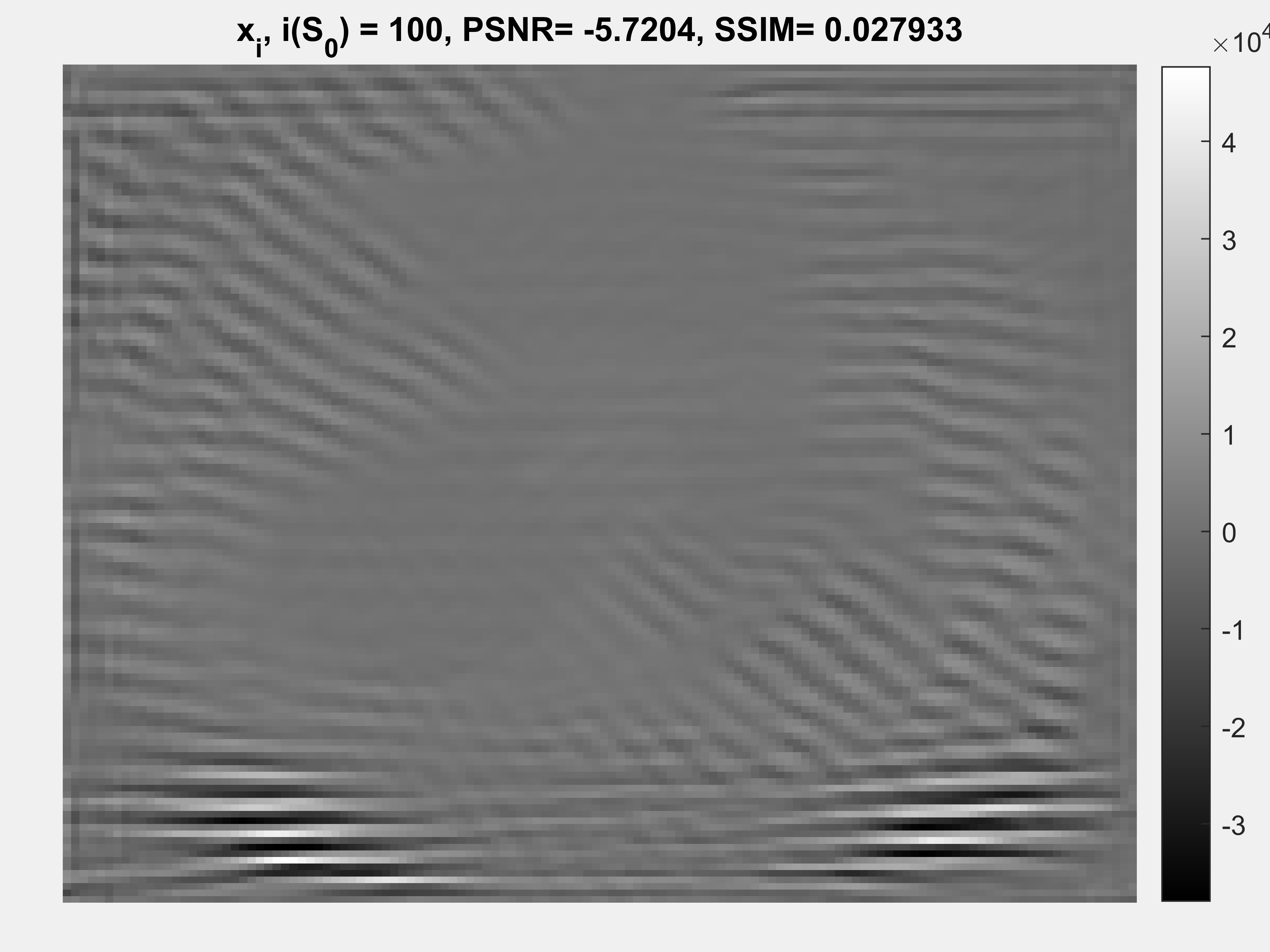}
        \caption{$x_{N}^{N_0}(\beta_i)$; $N=N_0=100, \beta_i = 1$}
        \label{Fig. sparser views NR}
    \end{subfigure} 
    \begin{subfigure}{0.495\textwidth}
        \includegraphics[width=\textwidth]{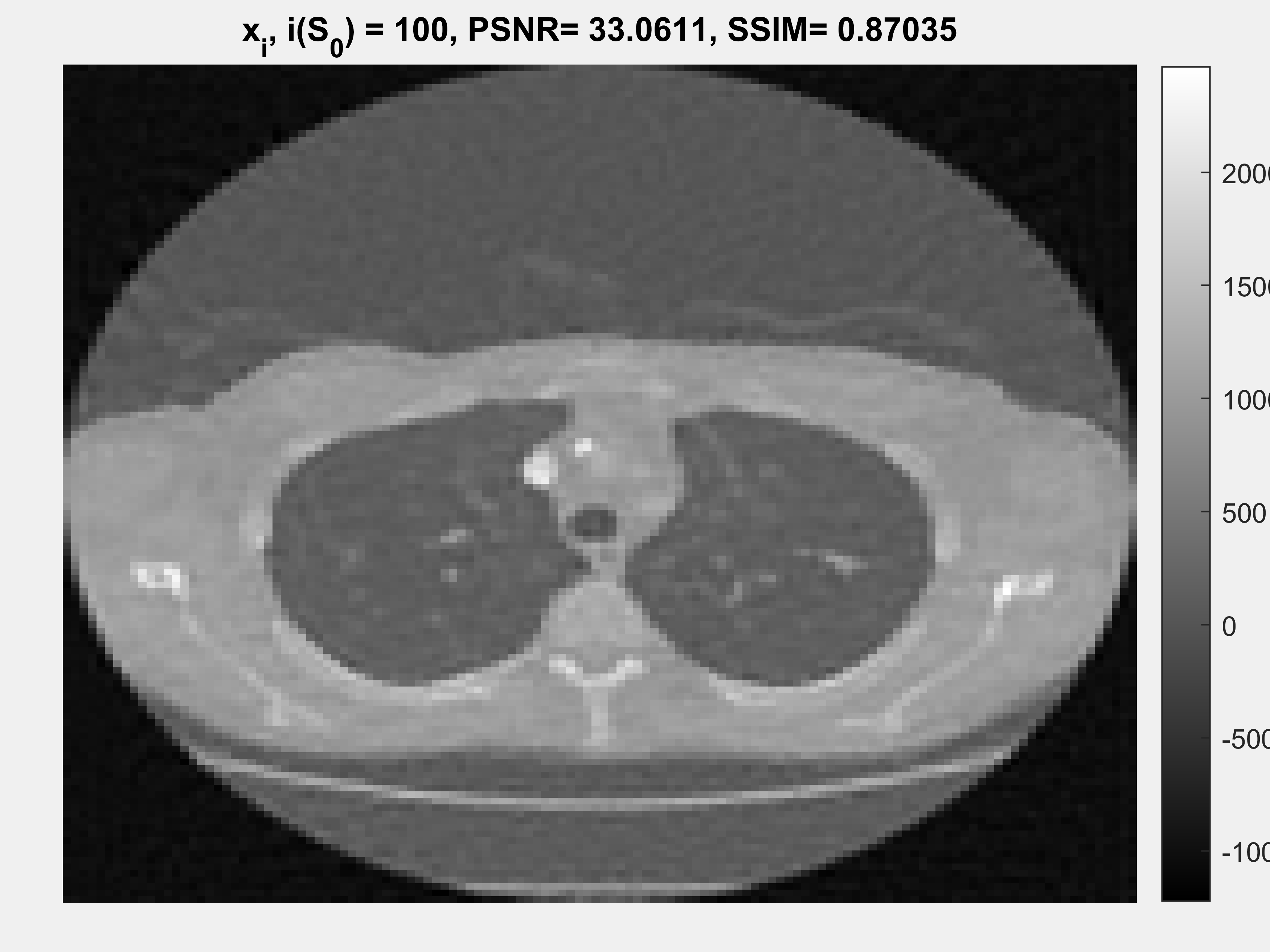}
        \caption{$x_{N}^{N_0}(\beta_i)$; $N=N_0=100, \beta_i = \beta_i(\mcal{S}_0,y_\delta)$}
        \label{Fig. sparser views R}
    \end{subfigure}    
    \begin{subfigure}{0.495\textwidth}
        \includegraphics[width=\textwidth]{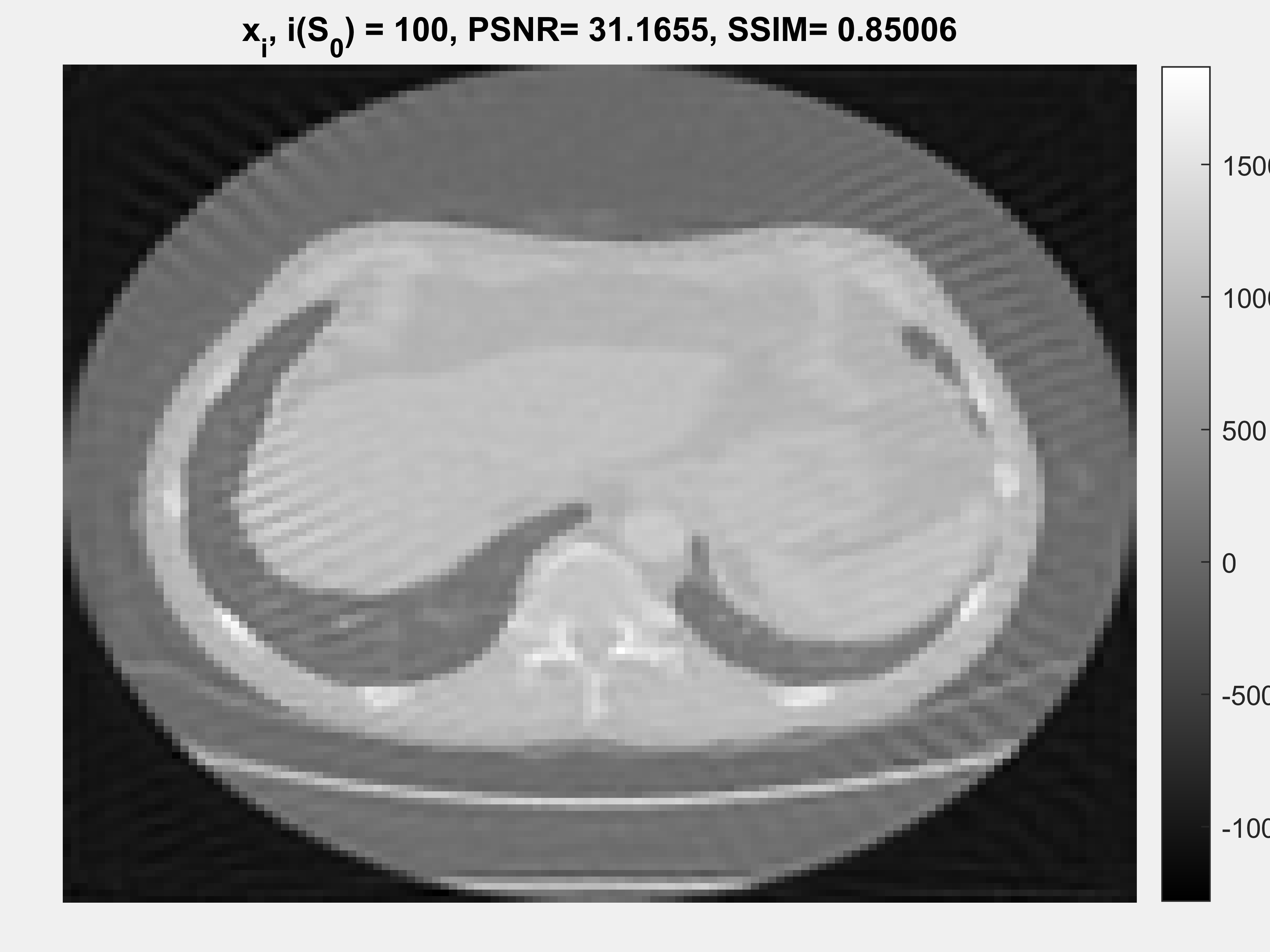}
        \caption{$x_{N}^{N_0}(\beta_i)$; $N=N_0=100, \beta_i = 1$}
        \label{Fig. Phantom 2 NR}
    \end{subfigure}
    \begin{subfigure}{0.495\textwidth}
        \includegraphics[width=\textwidth]{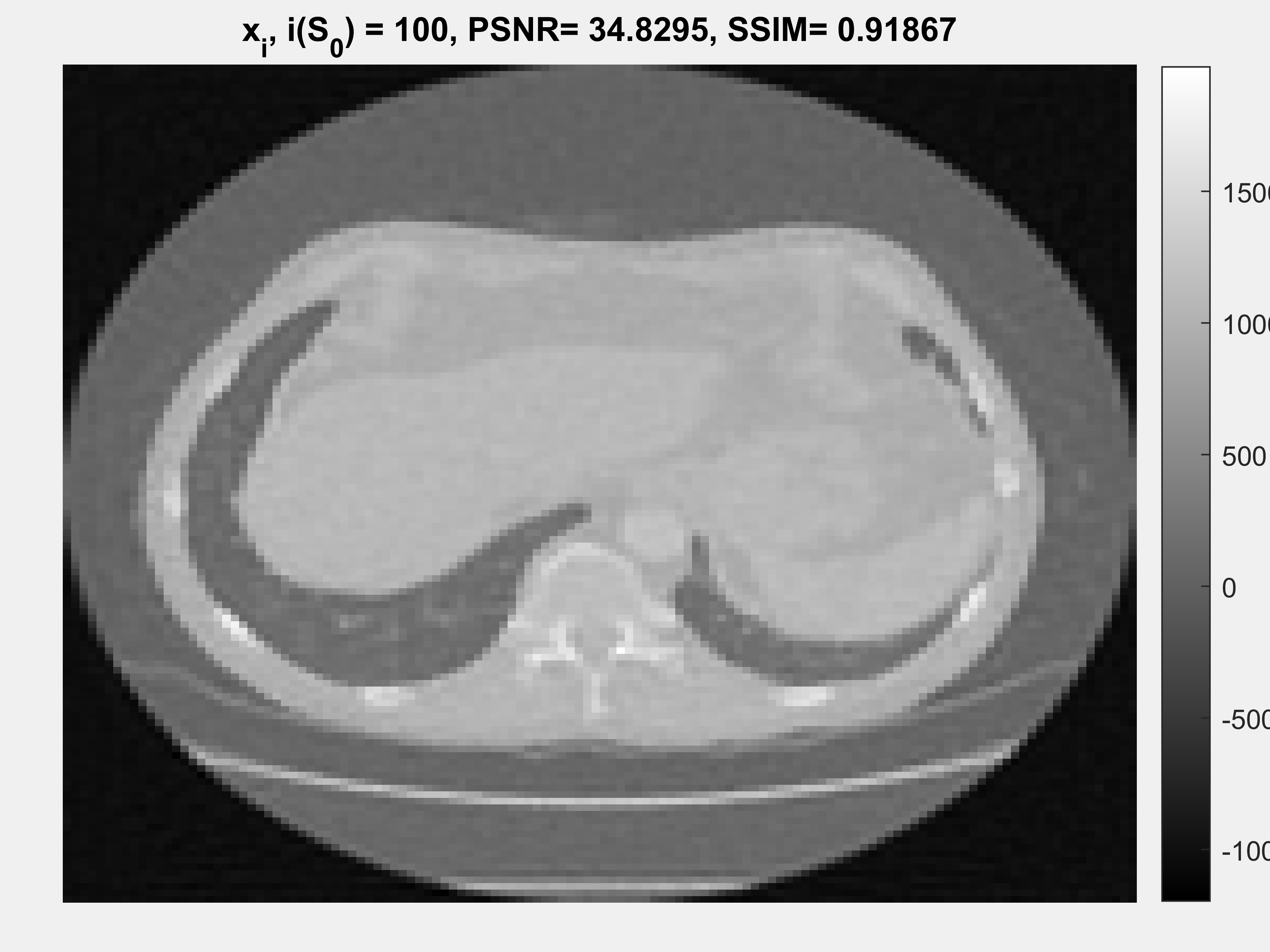}
        \caption{$x_{N}^{N_0}(\beta_i)$; $N=N_0=100, \beta_i = \beta_i(\mcal{S}_0,y_\delta)$}
        \label{Fig. Phantom 2 R}
    \end{subfigure}
    \begin{subfigure}{0.495\textwidth}
        \includegraphics[width=\textwidth]{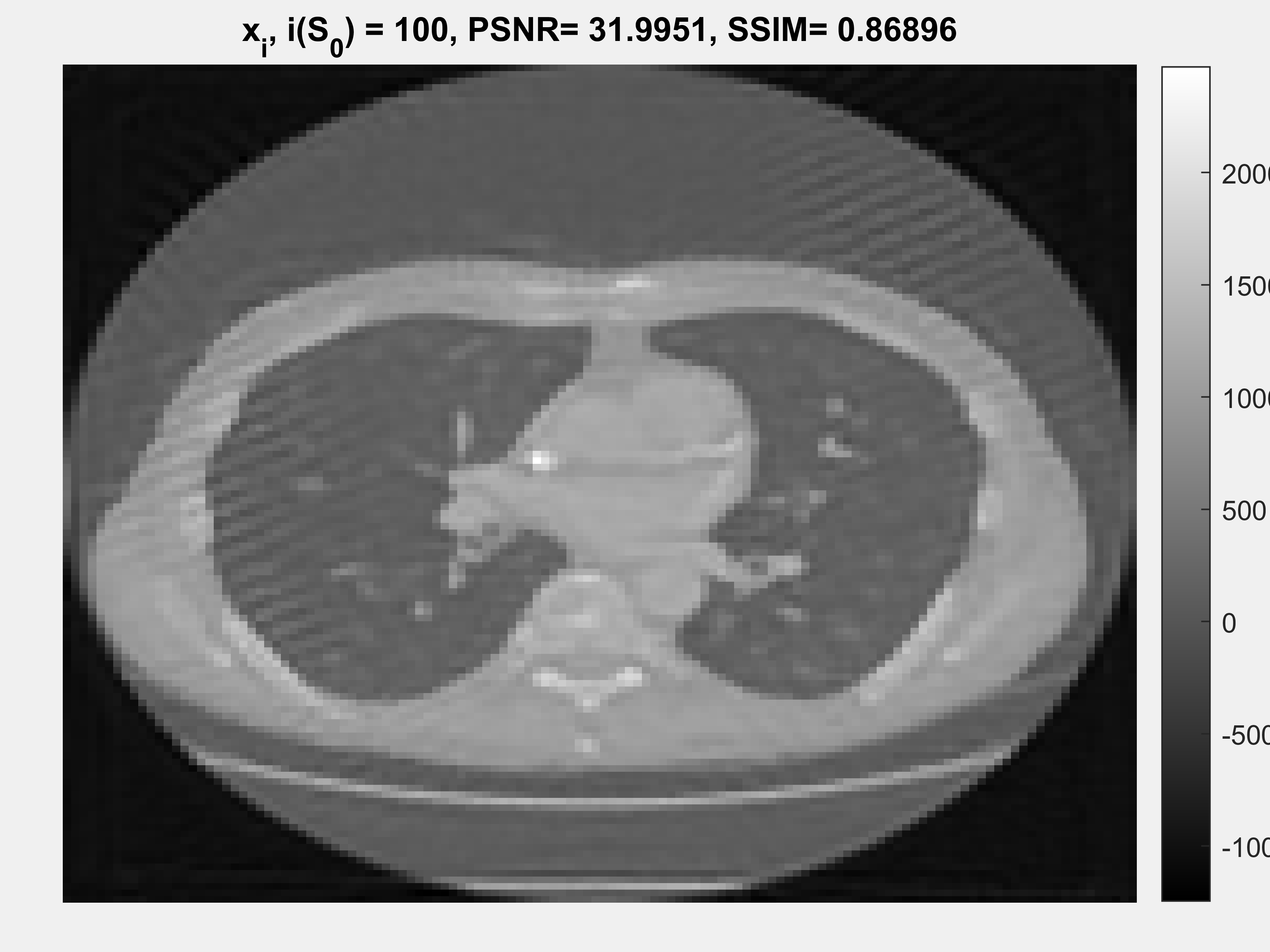}
        \caption{$x_{N}^{N_0}(\beta_i)$; $N=N_0=100, \beta_i = 1$}
        \label{Fig. Phantom 3 NR}
    \end{subfigure}
    \begin{subfigure}{0.495\textwidth}
        \includegraphics[width=\textwidth]{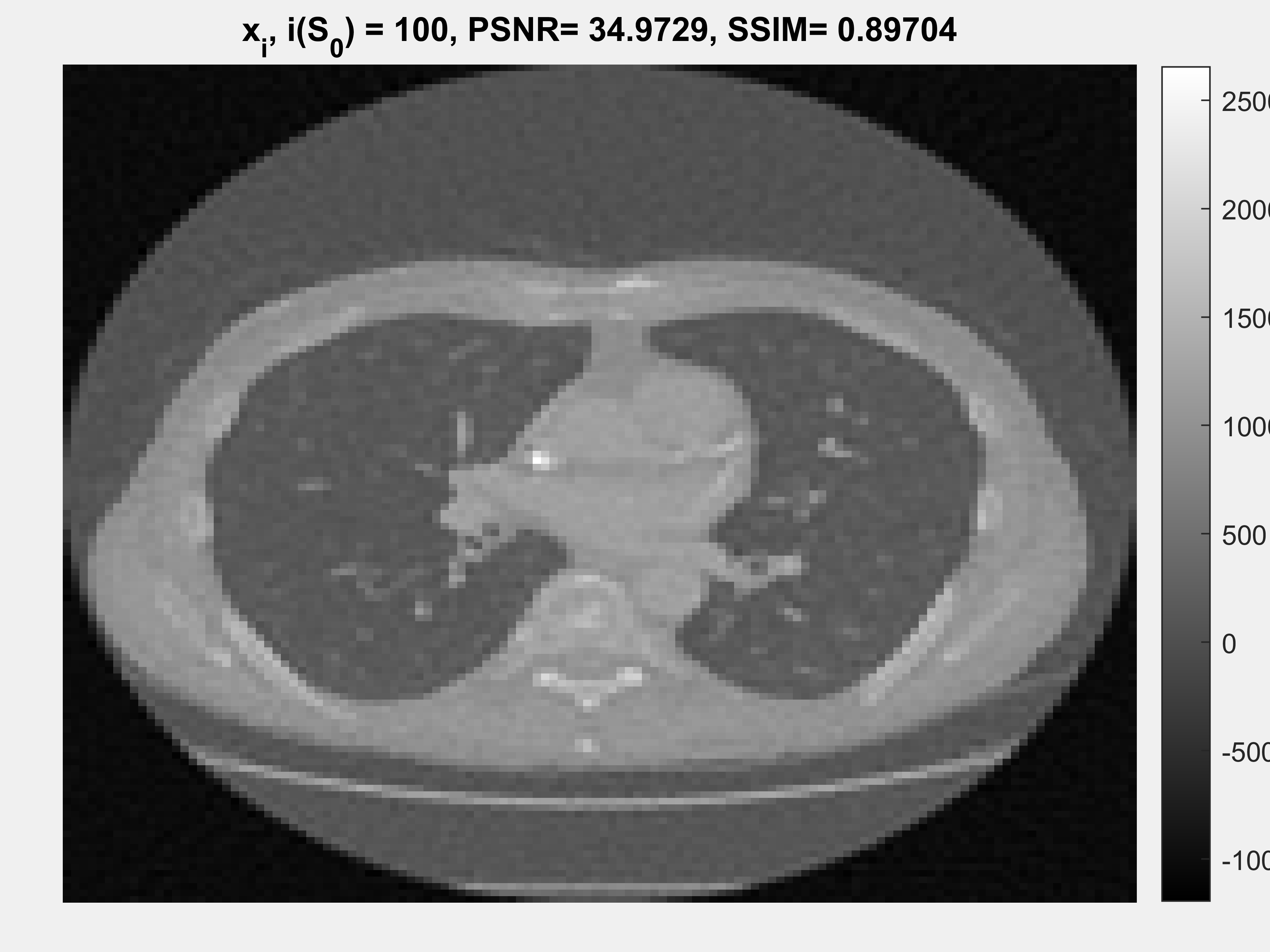}
        \caption{$x_{N}^{N_0}(\beta_i)$; $N=N_0=100, \beta_i = \beta_i(\mcal{S}_0,y_\delta)$}
        \label{Fig. Phantom 3 R}
    \end{subfigure}    
    \caption{Final solutions of an unrolled reconstruction algorithm, for regularized and not regularized $R_{\theta_i}s$.}
    \label{Fig. Regularized vs Unregularized}
\end{figure}

To further validate our regularization method, we experimented on sparser views, higher noise levels and few other CT phantoms from MATLAB's \textit{chestVolume} dataset. For the unrolled network architecture, we stuck with $N=100$ (total unrolled steps), $N_0=100$ and $\tau=10^{-5}$ for the total number of gradient steps and step-size in each $R_{\alpha_i}$, respectively, and the recovered solution as the final output of the unrolled network, i.e., $x_N^{N_0}$. 
\begin{enumerate}
\item Figures \ref{Fig. high noise NR} and \ref{Fig. high noise R} show the recovered solutions, for 90-views with $I_0 = 10^{5}$ (higher noise level), corresponding to the unregularized and the regularized $R_{\theta_i}s$, respectively. 

\item Figures \ref{Fig. sparser views NR} and \ref{Fig. sparser views R} show the recovered solutions, for 60-views (even sparser) with $I_0 = 10^{6}$ (noise level unchanged), corresponding to the unregularized and the regularized $R_{\theta_i}$, respectively. Note that, in this case, even with stronger classical regularization components ($R_{\alpha_i}s$), the final solution is unstable, unlike the previous examples. The reason might be, the ill-posedness of the problem for 60-views is stronger than that of 90-views.  

\item Figures \ref{Fig. Phantom 2 NR}, \ref{Fig. Phantom 2 R}, \ref{Fig. Phantom 3 NR} and \ref{Fig. Phantom 3 R} show the recovered solutions, for 90-views (unchanged) with $I_0 = 10^{6}$ (unchanged), corresponding to the unregularized and the regularized $R_{\theta_i}s$, respectively, for different phantom images. Note that, the behavior of all the recovery processes corresponding to 90-views and $I_0=10^6$, for different phantom images, are very similar, as expected.
\end{enumerate}

\end{example}

Unlike the previous examples, where there is a complete breakdown of the recovery process (severe instabilities), in the next example, we show that the recovered solution may contain certain (hallucinated) features/structures that are not present in the true phantom, even for a ``well recovered" solution (i.e., without severe instabilities). These recoveries are even more treacherous than the completely broken recoveries, as it can lead to an impression of a proper recovery with (hallucinated) features, that can be deceiving, and can lead to wrong interpretations/diagnoses. 

\begin{figure}[h!]
    \centering
    \begin{subfigure}{0.495\textwidth}
        \includegraphics[width=\textwidth]{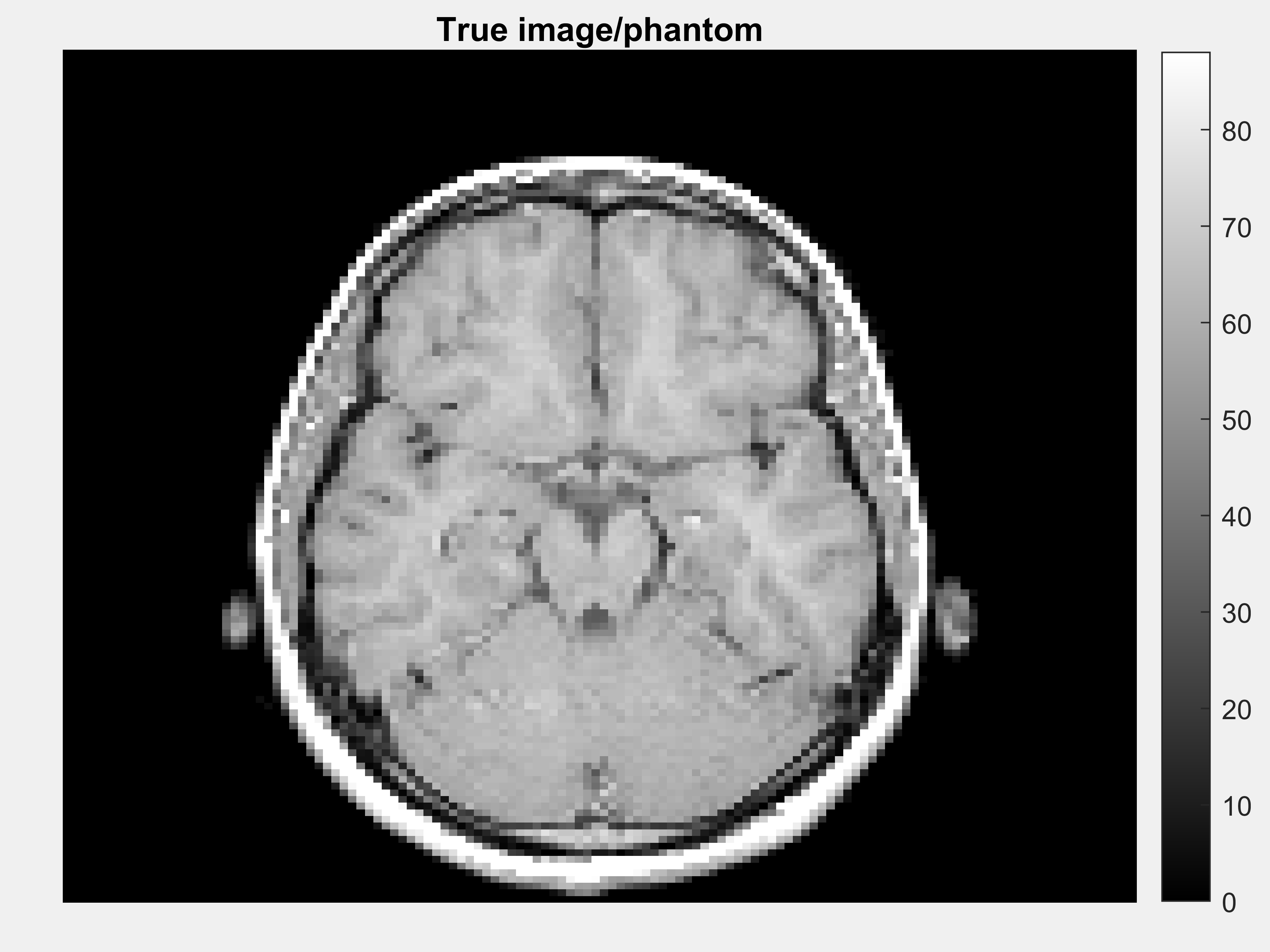}
        \caption{True image/phantom}
        \label{True image}
    \end{subfigure}       
    \begin{subfigure}{0.495\textwidth}
        \includegraphics[width=\textwidth]{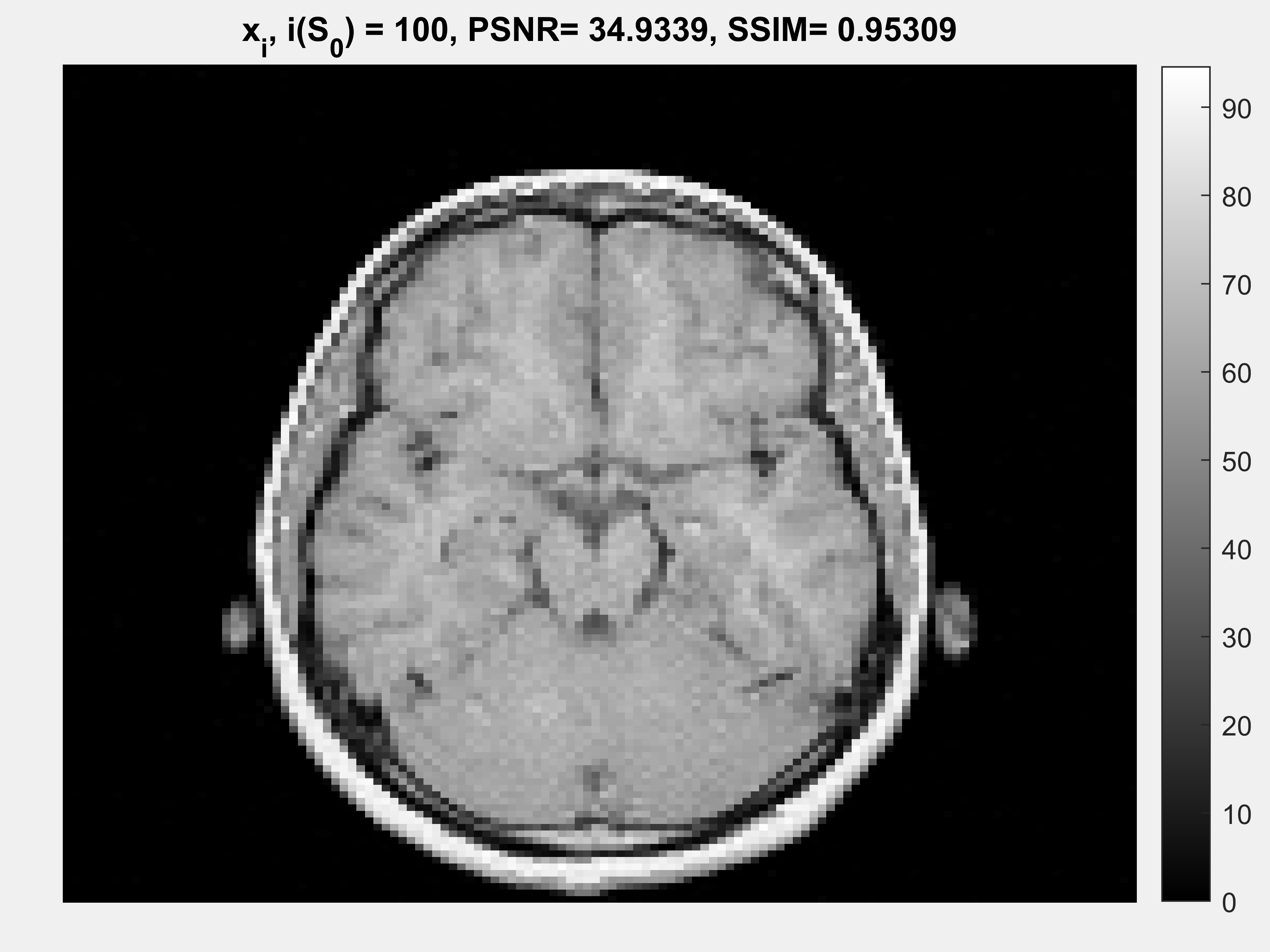}
        \caption{$x_{+N}^{\beta_i(\mcal{S}_0)}$, constrained and regularized}
        \label{brain_xN_C_R}
    \end{subfigure}       
    \begin{subfigure}{0.495\textwidth}
        \includegraphics[width=\textwidth]{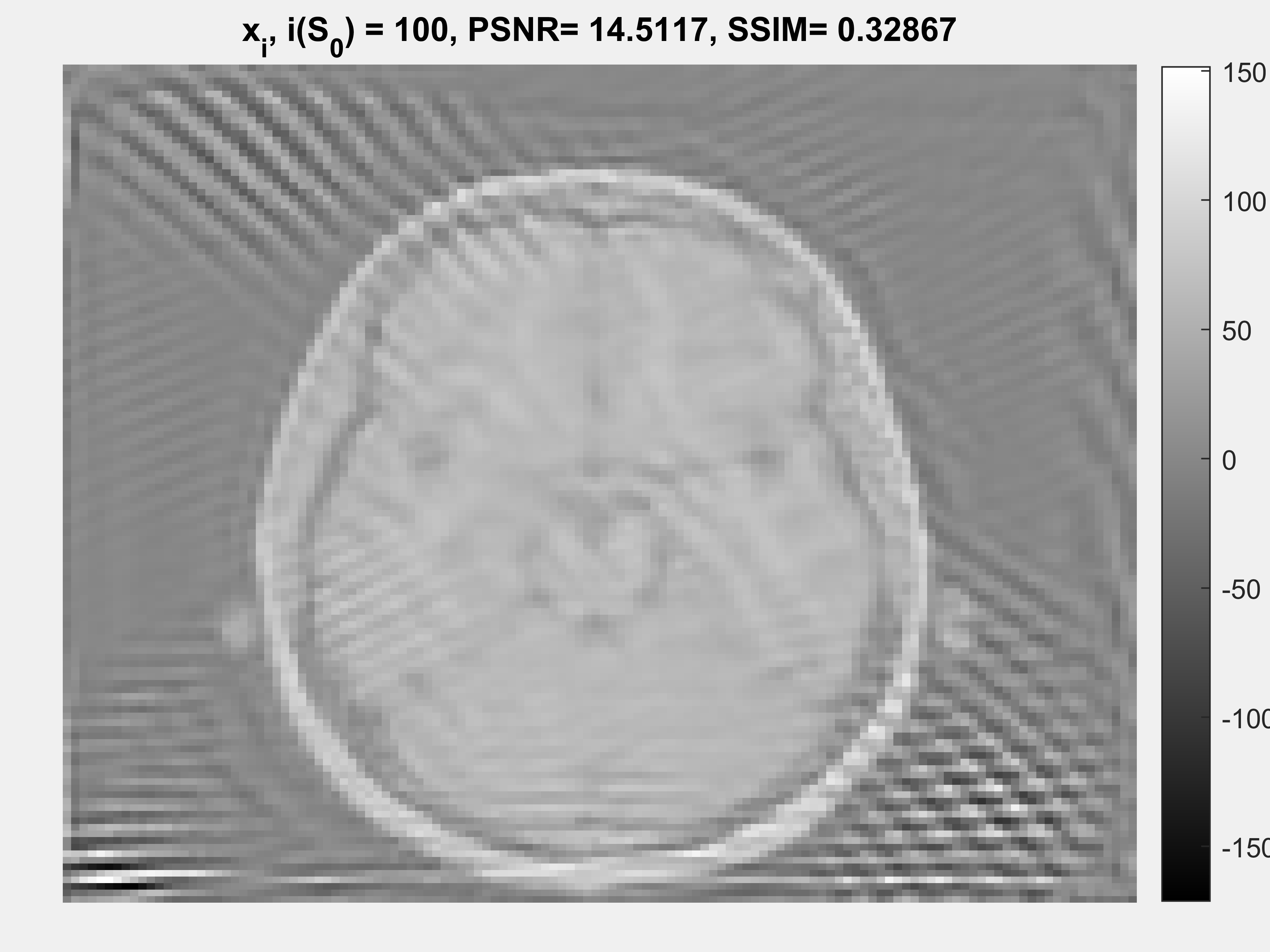}
        \caption{$x_{N}^{\beta_i=1}$, unconstrained and unregularized}
        \label{brain_xN_NC_NR}
    \end{subfigure} 
    \begin{subfigure}{0.495\textwidth}
        \includegraphics[width=\textwidth]{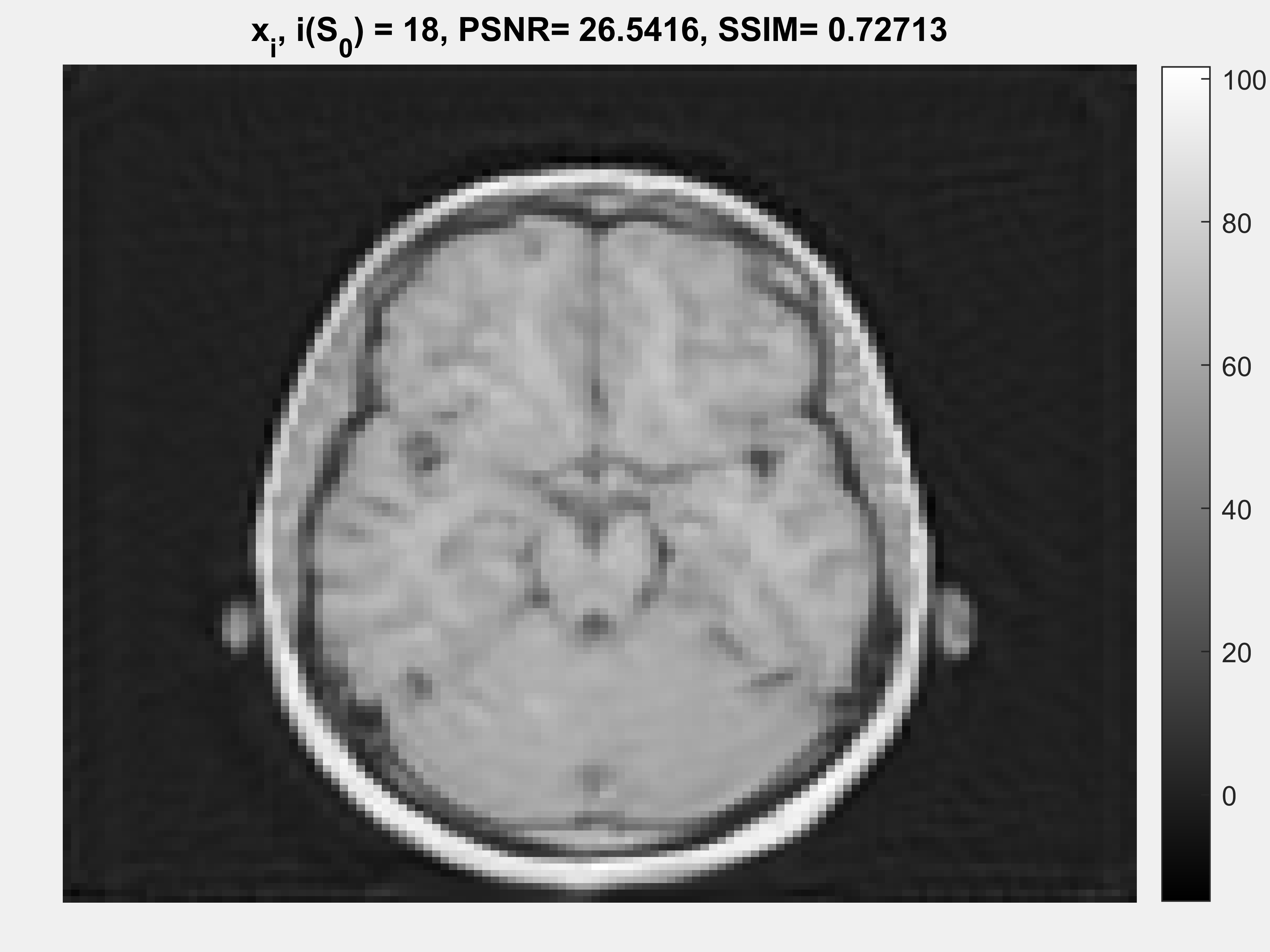}
        \caption{$x_{\mcal{S}_0}^{\beta_i=1}$, unconstrained and unregularized}
        \label{brain_xS0_NC_NR}
    \end{subfigure}    
    \begin{subfigure}{0.495\textwidth}
        \includegraphics[width=\textwidth]{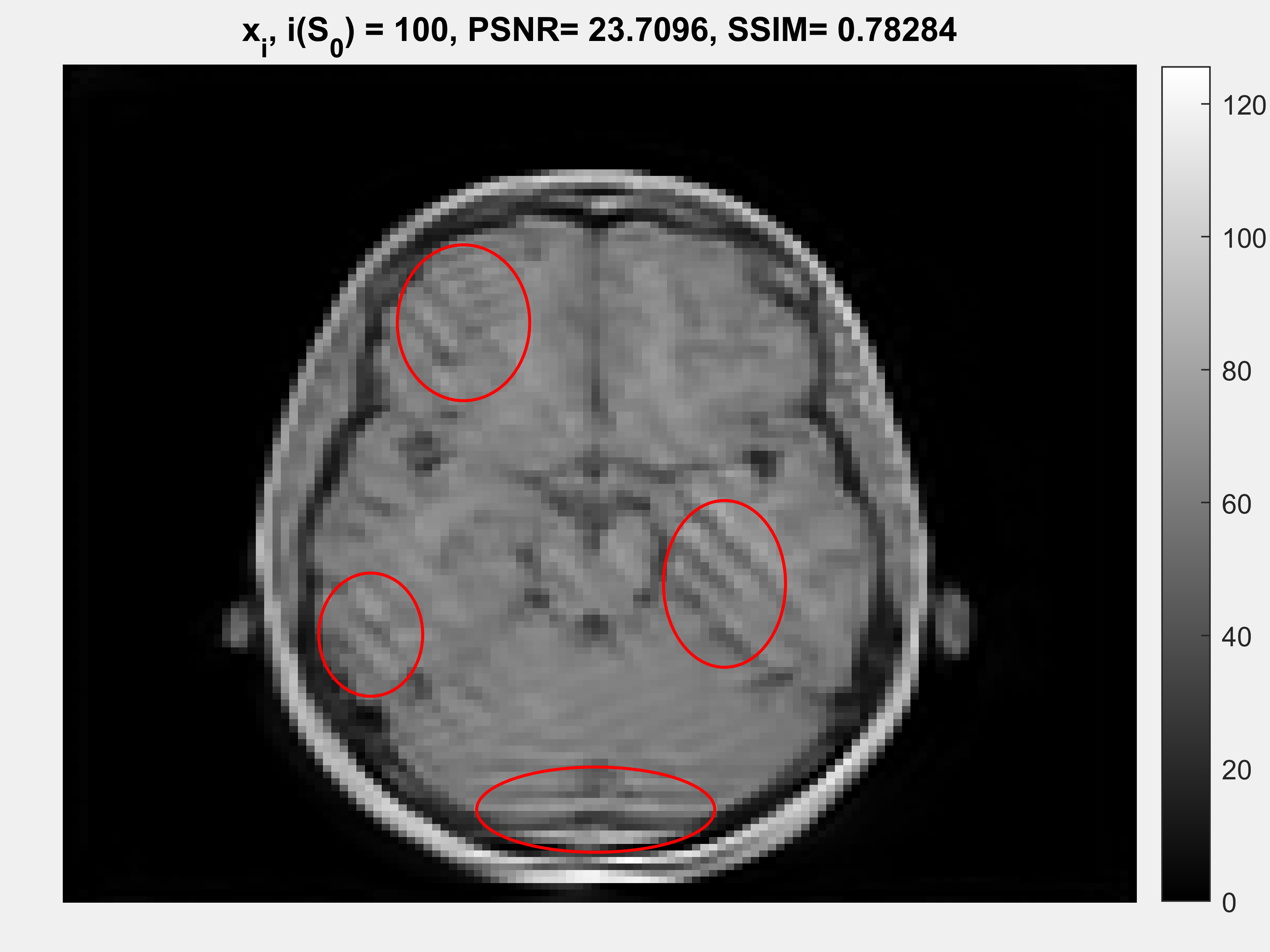}
        \caption{$x_{+N}^{\beta_i=1}$, constrained and unregularized}
        \label{brain_xN_C_NR}
    \end{subfigure}
    \begin{subfigure}{0.495\textwidth}
        \includegraphics[width=\textwidth]{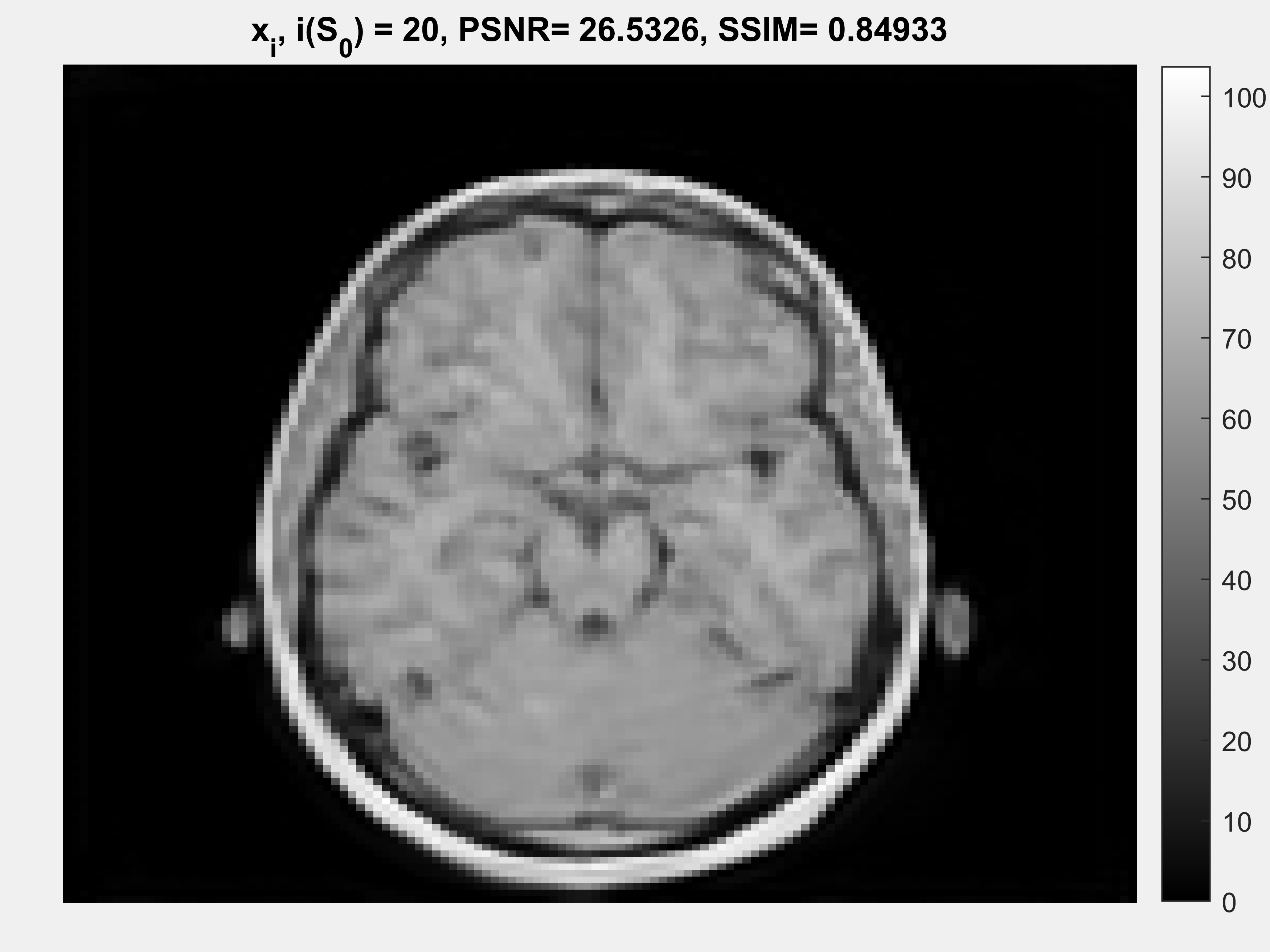}
        \caption{$x_{+\mcal{S}_0}^{\beta_i=1}$, constrained and unregularized}
        \label{brain_xS0_C_NR}
    \end{subfigure}
    \begin{subfigure}{0.495\textwidth}
        \includegraphics[width=\textwidth]{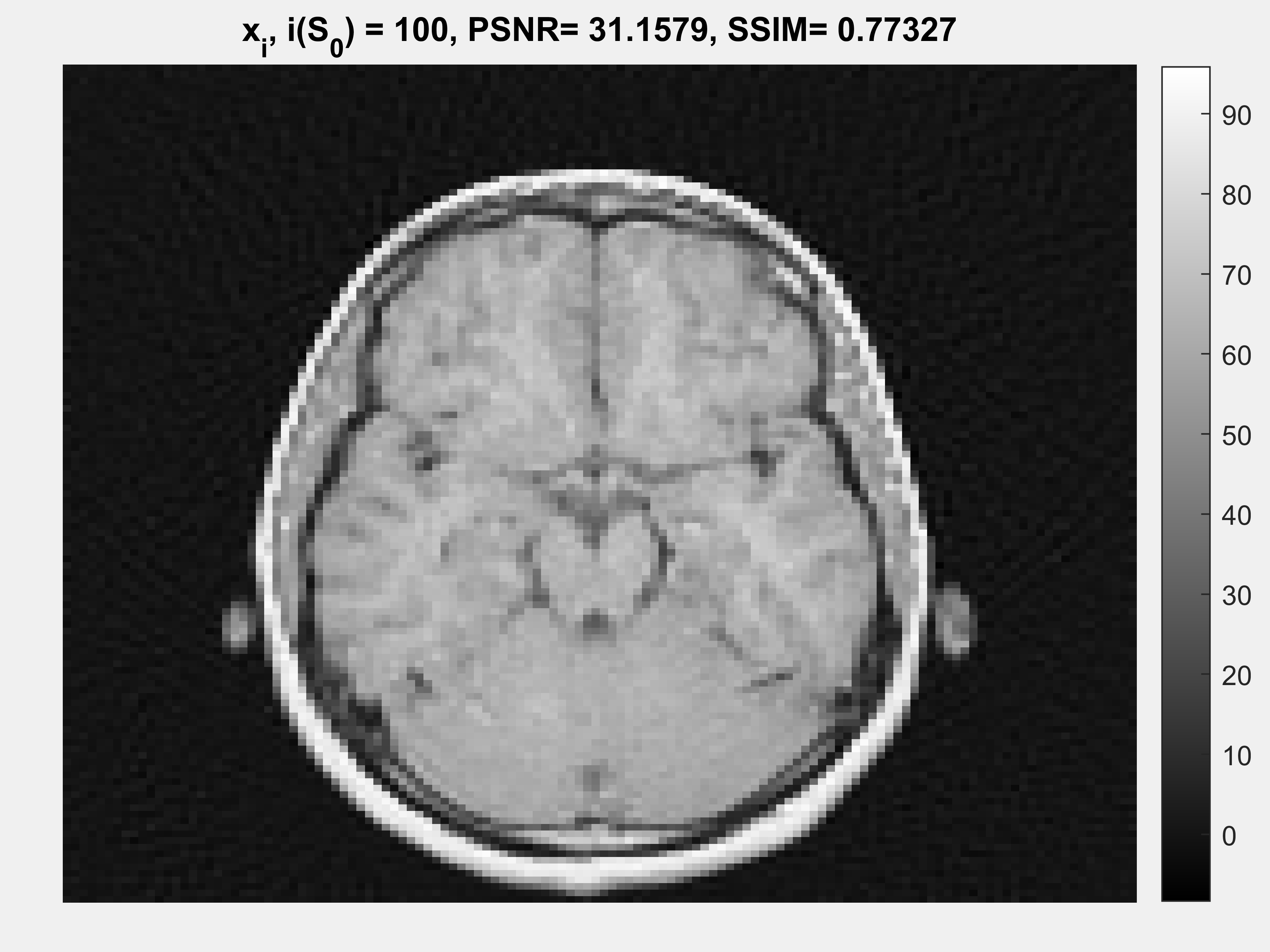}
        \caption{$x_{N}^{\beta_i(\mcal{S}_0)}$, unconstrained and regularized}
        \label{brain_xN_NC_R}
    \end{subfigure}
    \begin{subfigure}{0.495\textwidth}
        \includegraphics[width=\textwidth]{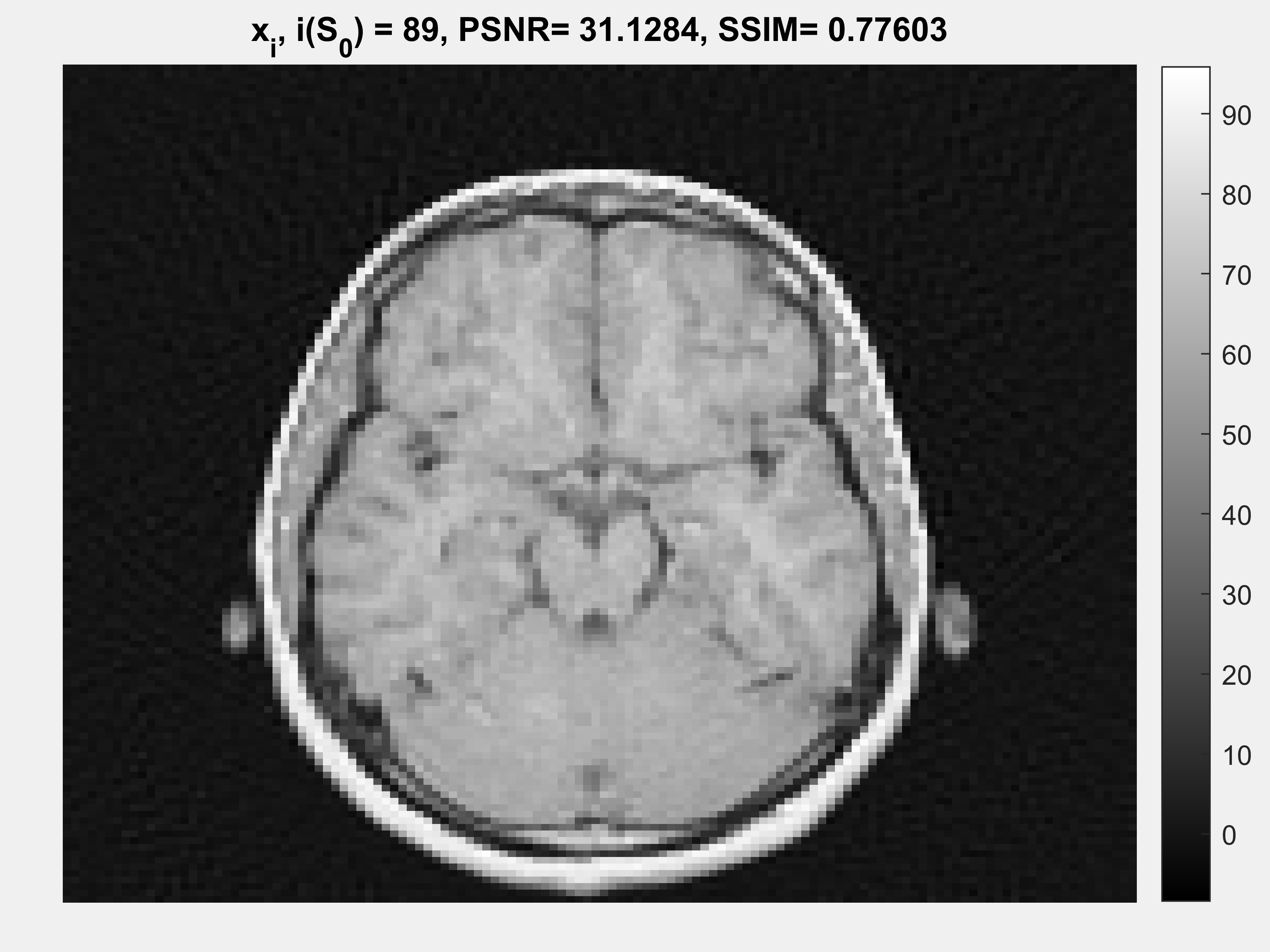}
        \caption{$x_{\mcal{S}_0}^{\beta_i(\mcal{S}_0)}$, unconstrained and regularized}
        \label{brain_xS0_NC_R}
    \end{subfigure}    
    \caption{Instabilities and hallucinations in the recovered solutions for constrained and uncontrained learned algorithms.}
    \label{Fig. Example hallucinations}
\end{figure}

\begin{example}\label{Example hallucination}
\textbf{[Hallucinated features]}

In this example, we consider a brain phantom from MATLAB's \textit{``mri"} dataset and generated the sinogram data, corresponding to 90-views and $I_0=10^6$, as discussed in the previous example. Here, we kept $N=100$ (\# unrolled steps), $\tau=10^{-5}$, but $N_0=30$ (\# inner iterations). In addition, here the intensities of true image is always non-negative ($\hat{x} \geq 0$), and hence, one can also restrict the iterates to be non-negative ($x_i \geq 0$, for all $i$), during the recovery process. Figure \ref{True image} shows the true phantom, Figure \ref{brain_xN_C_NR} shows the final solution $x_{+N}^{\beta_i=1}$ (with the constraint $x \geq 0$) and Figure \ref{brain_xN_NC_NR} shows the final solution $x_N^{\beta_i=1}$ (without the constraint), for the unrolled scheme without regularizing $R_{\theta_i}s$. Note that, although the constrained solution $x_{+N}^{\beta_i=1}$ doesn't completely break down like the unconstrained one ($x_N^{\beta_i=1}$), it does contain certain (hallucinated) features (which are circled red) that are not present in the true image $\hat{x}$. However, if we notice the solutions corresponding to the selection criterion ($\mcal{S}_0$), they are neither unstable nor contaminated by the hallucinated features, but they are of not great qualities, i.e., see Figures \ref{brain_xS0_C_NR} and \ref{brain_xS0_NC_NR} for $x_{+\mcal{S}_0}^{\beta_i=1}$ (constrained) and $x_{\mcal{S}_0}^{\beta_i=1}$ (unconstrained), respectively. Of course, the recoveries ($x_{+N}^{\beta_i(\mcal{S}_0)}$ and $x_{N}^{\beta_i(\mcal{S}_0)}$) significantly outperforms all other recoveries, irrespective of whether constrained or nor, and whether stopped early or not, see Figures \ref{brain_xN_C_R}, \ref{brain_xN_NC_R} and \ref{brain_xS0_NC_R}. This example validates that, even with additional constraints (like non-negativity etc.), the solutions of such learned reconstructive algorithms can not only be unstable, but can also contain (hallucinated) features/structures that are absent in the true solution, and by regularizing the learned components (together with the classical components), one can subdue these instabilities, and hence, can recover excellent stable solutions of the ill-posed inverse problems.

\end{example}

\section{Conclusion and Future Research}
In this paper, we showed that the recovery processes or reconstruction algorithms depending on a learned component can be very susceptible to strange instabilities, if not regularized properly. It's shown that such learned algorithms, which depend on pre-trained denoisers or pre-learned reconstruction components, can be inherently unstable and can exhibit instabilities in the recovered solutions, even without the adversarial noises, i.e., simple stochastic noises (in this case, Poisson noises) can either break down the recovery process completely or can induce hallucinated features/structures that are not present in the true solution. We also presented certain regularization methods to stabilize these instabilities, leading to much efficient and stable recoveries. We present the importance of continuity in a reconstruction algorithm, the failure of which, can lead to these wild instabilities. It is shown that, by parameterizing each step in an unrolled scheme one can stabilize/regularize an unstable recovery process. Furthermore, we also presented a technique to select the values of these (intermediate) regularization parameters, which is dependent on some selection criterion for the solution. Hence, the quality of the recovered solution, as well as, the stability of the reconstruction algorithm, is greatly dependent on the selection criterion, which is also the essence of a convergent regularization method.

In a future work we would to extend this idea of regularization to a proper (deep) unrolled reconstruction scheme, i.e., to train a deep neural network with classical regularization and learned regularization components and test the robustness of the algorithm by exposing it to  adversarial noises.

\begin{appendices}
\section{Proof of Deep-Learning Reconstruction Regularization}\label{Proof 1}
{From Definition} \ref{Definition Regularization},  {for a specific $y \in \mcal{D}(T^\dagger)$, a pair $(R_\alpha, \alpha)$ is called a (convergent) regularization method (for solving $Tx = y$) if the regularization operators $R_\alpha: \mcal{X} \mapsto \mcal{Y}$ are continuous and satisfy}
\begin{equation}
\limsup_{\delta \rightarrow 0} \; \{ ||R_{\alpha(\delta,y_\delta)}y_\delta - T^\dagger y||_{\mcal{X}} \; : \; y_\delta \in \mcal{Y}, \; ||y_\delta - y||_{\mcal{Y}} \leq \delta \; \} = 0, \tag{\ref{Regularization convergence}}
\end{equation}
 {and the regularization parameters $\alpha : \mbb{R}^+ \times \mcal{Y} \mapsto (0, \infty]$ are such that $\alpha = \alpha(\delta,y_\delta)$, for an existing parameter choice rule $\alpha(\delta,y_\delta)$ satisfying }
\begin{equation}
\limsup_{\delta \rightarrow 0} \; \{ \alpha(\delta,y_\delta) \; : \; y_\delta \in \mcal{Y}, ||y_\delta - y||_{\mcal{Y}} \leq \delta \} = 0. \tag{\ref{Parameter choice convergence}}
\end{equation}
 {Now, in} \eqref{Rtheta0 regularized 2}  {we have $R_\alpha$ continuous (since it's a classical convergent regularization method) and the product $\beta(\alpha,.)R_{\theta_0}$ satisfying condition} \eqref{betaRtheta0 continuous},  {i.e., being continuous, and hence, $R_{\alpha}^{\theta_0}$ is also continuous. In addition,}
\begin{align*}
||R_{\alpha(\delta,y_\delta)}^{\theta_0}(y_\delta) - T^\dagger y||_{\mcal{X}} &= ||R_{\alpha(\delta,y_\delta)}(y_\delta) - T^\dagger y + \beta(\alpha(\delta,y_\delta),y_\delta)R_{\theta_0}(y_\delta)||_{\mcal{X}}\\
& \leq ||R_{\alpha(\delta,y_\delta)}(y_\delta) - T^\dagger y||_{\mcal{X}} + \beta(\alpha(\delta,y_\delta),y_\delta)||R_{\theta_0}(y_\delta)||_{\mcal{X}},
\end{align*}
 {where $\limsup_{\delta \rightarrow 0}||R_{\alpha(\delta,y_\delta)}(y_\delta) - T^\dagger y||_{\mcal{X}} = 0$ and $\limsup_{\delta \rightarrow 0} \alpha(\delta, y_\delta) = 0$, since the parameter choice rule $\alpha(\delta,y_\delta)$ satisfies the condition} \eqref{Parameter choice convergence}.  {And if $\beta(\alpha,.)R_{\theta_0}$ satisfies the condition} \eqref{betaRtheta0 convergence to 0},  {we have $\limsup_{\delta \rightarrow 0} \beta(\alpha(\delta,y_\delta),y_\delta)||R_{\theta_0}(y_\delta)||_{\mcal{X}} = 0$. Therefore, we have}
\begin{equation}
\limsup_{\delta \rightarrow 0} \; \{ ||R_{\alpha(\delta,y_\delta)}^{\theta_0}y_\delta - T^\dagger y||_{\mcal{X}} \; : \; y_\delta \in \mcal{Y}, \; ||y_\delta - y||_{\mcal{Y}} \leq \delta \; \} = 0, \notag
\end{equation}
 {and hence, the pair $(R_\alpha^{\theta_0},\alpha)$ is also a convergent regularization method.}

\section{Connecting Plug-and-Play like methods to the Regularization operator}\label{PnP Regularization Connection}
 {The iterative (plug-and-play) algorithm used in the numerical results can be connected to} \eqref{Rtheta0 regularized 2}.  {Here, for a given noisy data $y_\delta$ with $\delta > 0$, the classical regularization method implemented is Landweber-iterations or semi-iterative regularization method, i.e., $R_\alpha(y_\delta; x_0^\delta) := x_{k(\delta,y_\delta)}^\delta$, for some parameter choice rule $k(\delta,y_\delta) < \infty$ or $||TR_\alpha(y_\delta) - y_\delta|| \geq \epsilon(\delta,y_\delta) > 0$, where $x_k^\delta = x_{k-1}^\delta - \tau_kT^*(Tx_{k-1}^\delta - y_\delta)$ and starting from $x_0^\delta$. And the regularized learned component $\beta(\alpha,y_\delta)R_{\theta_0}$ can be expressed as a sequence of nested regularized learned components}
\begin{align*}
\beta(\alpha,y_\delta)R_{\theta_0}(y_\delta) = R_{\theta_0^N}(\beta(\alpha_N,y_\delta),.) \circ R_{\theta_0^{N-1}}(\beta(\alpha_{N-1},y_\delta),.) \circ \cdots \circ R_{\theta_0^1}(\beta(\alpha_1,y_\delta),R_\alpha(y_\delta;x_0^\delta))
\end{align*}
 {where, $x_0^\delta \equiv 0$ and for $1 \leq i \leq N$}
\begin{align*}
R_{\theta_0^i}(\beta(\alpha_i,y_\delta),x) = R_\alpha(y_\delta; \; \beta(\alpha_i,y_\delta) R_{\theta_0}(x) + (1-\beta(\alpha_i,y_\delta))x),
\end{align*}
 {where $R_{\theta_0}$ is a pre-trained deep-learning based denoiser. The value of $N=100$ and $k(\delta,y_\delta)$-values are defined in the respective problem settings. Here, $\beta(\alpha_i,y_\delta)$ is also estimated according to a selection criterion (parameter choice rule), which is cross-validation in our examples.}
\end{appendices}

\bibliography{thesisref} 

\begin{thebibliography}{10}

\bibitem{Engl+Hanke+Neubauer}
H.~W. Engl, M.~Hanke, and A.~Neubauer, {\em Regularization of inverse
  problems}, vol.~375 of {\em Mathematics and its Applications}.
\newblock Kluwer Academic Publishers Group, Dordrecht, 1996.

\bibitem{Rudin_Osher_Fatemi}
L.~I. Rudin, S.~Osher, and E.~Fatemi, ``Nonlinear total variation based noise
  removal algorithms,'' {\em Physica D: Nonlinear Phenomena}, vol.~60, no.~1,
  pp.~259 -- 268, 1992.

\bibitem{Rudin_Osher_1994}
L.~I. {Rudin} and S.~{Osher}, ``Total variation based image restoration with
  free local constraints,'' in {\em Proceedings of 1st International Conference
  on Image Processing}, vol.~1, pp.~31--35 vol.1, 1994.

\bibitem{Donoho_Elad_Bruckstein_2009}
A.~M. Bruckstein, D.~L. Donoho, and M.~Elad, ``From sparse solutions of systems
  of equations to sparse modeling of signals and images,'' {\em SIAM Review},
  vol.~51, no.~1, pp.~34--81, 2009.

\bibitem{Daubechies_Defrise_DeMol_2004}
I.~Daubechies, M.~Defrise, and C.~De~Mol, ``An iterative thresholding algorithm
  for linear inverse problems with a sparsity constraint,'' {\em Communications
  on Pure and Applied Mathematics}, vol.~57, no.~11, pp.~1413--1457, 2004.

\bibitem{Candes_Romberg_Tao_2006}
E.~J. Candès, J.~K. Romberg, and T.~Tao, ``Stable signal recovery from
  incomplete and inaccurate measurements,'' {\em Communications on Pure and
  Applied Mathematics}, vol.~59, no.~8, pp.~1207--1223, 2006.

\bibitem{Tibshirani_2006}
R.~Tibshirani, ``Regression shrinkage and selection via the lasso,'' {\em
  Journal of the Royal Statistical Society: Series B (Methodological)},
  vol.~58, no.~1, pp.~267--288, 1996.

\bibitem{Abinash1}
A.~Nayak, ``A new regularization approach for numerical differentiation,'' {\em
  Inverse Problems in Science and Engineering}, vol.~0, no.~0, pp.~1--26, 2020.

\bibitem{Zhu_Liu_Cauley_Bruce_Matthew_2018}
B.~Zhu, J.~Z. Liu, S.~F. Cauley, B.~R. Rosen, and M.~S. Rosen, ``Image
  reconstruction by domain-transform manifold learning,'' {\em Nature},
  vol.~555, 2018.

\bibitem{Gregor_LeCun_2010}
K.~Gregor and Y.~LeCun, ``Learning fast approximations of sparse coding,''
  ICML'10, (Madison, WI, USA), p.~399–406, Omnipress, 2010.

\bibitem{Yang_Sun_Li_Xu_2016}
y.~yang, J.~Sun, H.~Li, and Z.~Xu, ``Deep admm-net for compressive sensing
  mri,'' in {\em Advances in Neural Information Processing Systems} (D.~Lee,
  M.~Sugiyama, U.~Luxburg, I.~Guyon, and R.~Garnett, eds.), vol.~29, Curran
  Associates, Inc., 2016.

\bibitem{Oktem_Adler_2017}
J.~Adler and O.~Öktem, ``Solving ill-posed inverse problems using iterative
  deep neural networks,'' {\em Inverse Problems}, vol.~33, p.~124007, nov 2017.

\bibitem{Unser_Gupta_Jin_Nguyen_McCann_2018}
H.~Gupta, K.~H. Jin, H.~Q. Nguyen, M.~T. McCann, and M.~Unser, ``Cnn-based
  projected gradient descent for consistent ct image reconstruction,'' {\em
  IEEE Transactions on Medical Imaging}, vol.~37, no.~6, pp.~1440--1453, 2018.

\bibitem{Jin_McCann_Froustey_Unser_2017}
K.~Jin, M.~McCann, E.~Froustey, and M.~Unser, ``Deep convolutional neural
  network for inverse problems in imaging,'' {\em {IEEE} Transactions on Image
  Processing}, vol.~26, no.~9, pp.~4509--4522, 2017.

\bibitem{Knoll_Pock_Hammernik_Klatzer_Kobler_Recht_Sodickson_2018}
K.~Hammernik, T.~Klatzer, E.~Kobler, M.~P. Recht, D.~K. Sodickson, T.~Pock, and
  F.~Knoll, ``Learning a variational network for reconstruction of accelerated
  mri data,'' {\em Magnetic Resonance in Medicine}, vol.~79, no.~6,
  pp.~3055--3071, 2018.

\bibitem{Bora_Dimakis_Jalal_Price_2017}
A.~Bora, A.~Jalal, E.~Price, and A.~G. Dimakis, ``Compressed sensing using
  generative models,'' in {\em Proceedings of the 34th International Conference
  on Machine Learning} (D.~Precup and Y.~W. Teh, eds.), vol.~70 of {\em
  Proceedings of Machine Learning Research}, pp.~537--546, PMLR, 06--11 Aug
  2017.

\bibitem{Wang_Chen_Yi_Weihua_Liao_Li_Zhou_2017}
H.~Chen, Y.~Zhang, W.~Zhang, P.~Liao, K.~Li, J.~Zhou, and G.~Wang, ``Low-dose
  ct via convolutional neural network,'' {\em Biomed. Opt. Express}, vol.~8,
  pp.~679--694, Feb 2017.

\bibitem{Genzel_Macdonald_Marz_arXiv2020}
M.~Genzel, J.~Macdonald, and M.~März, ``Solving inverse problems with deep
  neural networks -- robustness included?,'' 2020.

\bibitem{Carola_Oktem_Maass_arridge_2019}
S.~Arridge, P.~Maass, O.~Öktem, and C.-B. Schönlieb, ``Solving inverse
  problems using data-driven models,'' {\em Acta Numerica}, vol.~28,
  p.~1–174, 2019.

\bibitem{Willet_Dimakis_Ongie_Jalal_Metzler_Baraniuk_2020}
G.~Ongie, A.~Jalal, C.~A. Metzler, R.~Baraniuk, A.~G. Dimakis, and R.~M.
  Willett, ``Deep learning techniques for inverse problems in imaging,'' {\em
  IEEE Journal on Selected Areas in Information Theory}, vol.~1, pp.~39--56,
  2020.

\bibitem{Oktem_Andreas_Jonas_Arridge_2020}
A.~Hauptmann, J.~Adler, S.~Arridge, and k.~Ozan, ``Multi-scale learned
  iterative reconstruction,'' {\em IEEE Transactions on Computational Imaging},
  vol.~PP, pp.~1--1, 04 2020.

\bibitem{Antun_Renna_Poon_Adcock_Hansen}
V.~Antun, F.~Renna, C.~Poon, B.~Adcock, and A.~C. Hansen, ``On instabilities of
  deep learning in image reconstruction and the potential costs of ai,'' {\em
  Proceedings of the National Academy of Sciences}, vol.~117, no.~48,
  pp.~30088--30095, 2020.

\bibitem{Antun_Hansen_Adcock_Gottschling_arXiv2020}
N.~M. Gottschling, V.~Antun, B.~Adcock, and A.~C. Hansen, ``The troublesome
  kernel: why deep learning for inverse problems is typically unstable,'' 2020.

\bibitem{Huang_Maier_Tobias_Katharina_Ling_Gunther_2018}
Y.~Huang, T.~W{\"u}rfl, K.~Breininger, L.~Liu, G.~Lauritsch, and A.~Maier,
  ``Some investigations on robustness of deep learning in limited angle
  tomography,'' in {\em Medical Image Computing and Computer Assisted
  Intervention -- MICCAI 2018}, pp.~145--153, Springer International
  Publishing, 2018.

\bibitem{Schwab_Antholzer_Haltmeier_2019}
J.~Schwab, S.~Antholzer, and M.~Haltmeier, ``Deep null space learning for
  inverse problems: convergence analysis and rates,'' {\em Inverse Problems},
  vol.~35, p.~025008, jan 2019.

\bibitem{nayak2021PnPInstabilities}
A.~Nayak,
  ``\href{https://18a354c4-35f1-4d10-b309-d7d10965a85c.filesusr.com/ugd/a2137e_6e46cdeeb97d499497a9de82e2166b4b.pdf}{Instabilities
  in Plug-and-Play (PnP) algorithms from a learned denoiser},'' 2021.

\bibitem{Hansen_IRtools}
S.~Gazzola, P.~Hansen, and J.~Nagy, ``Ir tools - a matlab package of iterative
  regularization methods and large-scale test problems,'' {\em Numerical
  Algorithms}, 2018.

\bibitem{nayak2021PnP}
A.~Nayak, ``\href{https://arxiv.org/abs/2106.07795}{Interpretation of
  Plug-and-Play (PnP) algorithms from a different angle},'' 2021.

\end{thebibliography}
\bibliographystyle{ieeetr}

\end{document}